\documentclass{article} % For LaTeX2e
\usepackage{iclr2027_conference,times}

\usepackage{amsmath,amsfonts,bm}

\def\eqref#1{equation~\ref{#1}}
\def\1{\bm{1}}

\DeclareMathAlphabet{\mathsfit}{\encodingdefault}{\sfdefault}{m}{sl}
\SetMathAlphabet{\mathsfit}{bold}{\encodingdefault}{\sfdefault}{bx}{n}

\usepackage{hyperref}
\usepackage{url}

\usepackage[utf8]{inputenc} % allow utf-8 input
\usepackage[T1]{fontenc}    % use 8-bit T1 fonts
\usepackage{hyperref}       % hyperlinks
\usepackage{url}            % simple URL typesetting
\usepackage{booktabs}       % professional-quality tables
\usepackage{amsfonts}       % blackboard math symbols
\usepackage{nicefrac}       % compact symbols for 1/2, etc.
\usepackage{microtype}      % microtypography
\usepackage{multirow}

\usepackage{xpatch}
\usepackage[table]{xcolor}
\usepackage{graphicx} % 必须引入此宏包用于 resizebox
\usepackage{subcaption}
\usepackage{float}
\usepackage[most]{tcolorbox}
\tcbuselibrary{raster,listings,skins,breakable}
\usepackage{placeins}
\usepackage{upquote}
\usepackage[most]{tcolorbox}
\lstdefinestyle{promptstyle}{
    basicstyle=\ttfamily\scriptsize,
    breaklines=true,
    breakatwhitespace=false,
    columns=fullflexible,
    keepspaces=true,
    showstringspaces=false,
    upquote=true,
    mathescape=false,
    texcl=false,
    language={}
}

\usepackage{needspace}
\newcommand{\promptfilebox}[3][]{%
\par\medskip
\Needspace{0.35\textheight}
\noindent
\tcbinputlisting{
    enhanced,
    listing only,
    listing file={#3},
    title={#2},
    colback=white,
    colframe=black,
    fonttitle=\bfseries,
    left=1mm,
    right=1mm,
    top=1mm,
    bottom=1mm,
    boxsep=1mm,
    before skip=4pt,
    after skip=4pt,
    listing options={style=promptstyle},
    #1
}
\par\medskip
}
\makeatletter
\xapptocmd{\NAT@bibsetnum}{\setlength{\leftmargin}{0pt}\setlength{\itemindent}{\labelwidth}\addtolength{\itemindent}{\labelsep}}{}{}
\makeatother

\newcounter{diag}[section]

\usepackage{enumitem}
\usepackage{tabularx}
\usepackage{pifont}
\usepackage{wrapfig}

\definecolor{gtback}{HTML}{F3F8FF}      % very light blue background
\definecolor{gtframe}{HTML}{5B8FD9}     % medium blue border
\definecolor{gttitle}{HTML}{DCEBFF}     % light blue title background

\definecolor{baseback}{HTML}{FFF8EF}      % very light warm orange background
\definecolor{baseframe}{HTML}{D98A2B}     % moderate amber-orange border
\definecolor{basetitle}{HTML}{C56A00}     % darker orange title bar

\newcommand{\agent}[1]{\noindent\textcolor{blue!70!black}{\textbf{Assistant: }}#1\par\vspace{1mm}}
\newcommand{\userA}[1]{\noindent\textcolor{red!70!black}{\textbf{User: }#1}\par\vspace{1mm}}

\title{TRACER: Trajectory-Aligned Learning for Multi-Turn User Simulation}

\author{%
\makebox[\dimexpr\textwidth-2\tabcolsep\relax][c]{%
\begin{tabular}{c}
\textbf{Geng Chen}\textsuperscript{1,$\dagger$},
\textbf{Ruotong Pan}\textsuperscript{2,$\dagger$},
\textbf{Zhirui Yang}\textsuperscript{2},
\textbf{Qiqi He}\textsuperscript{3}
\\
\textbf{Jiawei Chen}\textsuperscript{4},
\textbf{Zhang Yunfei}\textsuperscript{2},
\textbf{Chongyuan Chen}\textsuperscript{2},
\textbf{Minxuan Lv}\textsuperscript{5}
\\
\textbf{Zheng Yang}\textsuperscript{6},
\textbf{Win-Bin Huang}\textsuperscript{1},
\textbf{Xiangyu Wu}\textsuperscript{2,*},
\textbf{Wenwu Ou}\textsuperscript{2}
\\[3pt]
{\normalfont
\textsuperscript{1}Peking University
\qquad
\textsuperscript{2}Kuaishou Technology
\qquad
\textsuperscript{3}Xi'an Jiaotong University
}
\\
{\normalfont
\textsuperscript{4}Institute of Software, Chinese Academy of Sciences
\qquad
\textsuperscript{5}Institute of Information Engineering
}
\\
{\normalfont
\textsuperscript{6}Beijing University of Posts and Telecommunications
}
\end{tabular}}%
}

\usepackage{amsmath}  % 提供基础数学公式支持（如多行公式、特定环境等）
\usepackage{amssymb}  % 提供 \mathbb 等额外数学符号和字体
\usepackage{wrapfig}

\newcommand{\OurMethod}{TRACER}

\iclrfinalcopy % Uncomment for camera-ready version, but NOT for submission.【加上作者】
\begin{document}

\maketitle
% \fancyhead{}\renewcommand{\headrulewidth}{0pt} % 用来隐去页眉的
\fancyhead[L]{Under review as a conference paper at ICLR 2027} %把页眉变回under review
\renewcommand{\thefootnote}{\fnsymbol{footnote}}

\footnotetext[2]{These authors contributed equally to this work.}

\footnotetext[1]{Corresponding author. E-mail:
\href{mailto:wuxiangyu06@kuaishou.com}
{\texttt{wuxiangyu06@kuaishou.com}}.}

\renewcommand{\thefootnote}{\arabic{footnote}}

\begin{abstract}
Faithful user simulation is fundamental to building, evaluating, and improving interactive AI at scale. Yet current simulators often produce plausible individual responses without reproducing the intent evolution and outcomes observed in real interactions. We propose TRACER, a multi-turn user simulator that models evolving user intent and aligns simulated trajectories with real ones. TRACER is trained in two stages: supervised
fine-tuning on real user dialogues, followed by multi-turn reinforcement learning. The RL stage combines hierarchical outcome- and trajectory-level rewards with deviation-aware advantage modulation, jointly addressing reward sparsity and credit assignment challenges in long dialogues. On real customer-service sessions organized into reference cohorts, TRACER-7B surpasses the strongest baseline by 11.4 conversion F1 points, while outperforming all baselines on group-level conversion-rate error and semantic trajectory distance and generalizing to out-of-distribution scenarios.
In human Turing tests, annotators identified TRACER conversations at near-chance accuracy. Building on this simulator, we further introduce the Dynamic Marketing Benchmark, which jointly evaluates persuasion and response quality via simulated interactions, revealing that higher response quality does not necessarily correspond to higher conversion rates.

%Our code and benchmark are available at \url{https://anonymous.4open.science/r/TRACER-F0E8}.
\end{abstract}

\section{Introduction}
\label{src:intro}

% 随着大语言模型在对话系统中的广泛应用，可扩展的用户反馈对模型评测与改进愈发重要。
% 用户模拟器通过与客服代理交互，为离线评测和强化学习提供反馈。
% 然而，这些反馈的价值不仅取决于回复是否自然，还取决于模拟器能否复现用户在交互中的响应与决策。
% 因此，多轮用户模拟需要同时考虑用户如何表达，以及其行为如何在完整交互中展开。
With the widespread application of large language models (LLMs) in dialogue systems, scalable user feedback has become increasingly important for model evaluation and improvement. User simulators provide such feedback through interactions with dialogue agents~\citep{hu2023unlocking,abbasiantaeb2024let,qiu2024interactive,dou-etal-2025-simulatorarena}, supporting offline evaluation and reinforcement learning~\citep{wu2025collabllm,wang2025rlver,qian2025userrl}. Their usefulness, however, depends not only on linguistically natural responses but also on whether simulated users make decisions consistent with real human behavior across the interaction. Multi-turn simulation must therefore account for how users express themselves and how their behavior unfolds over complete interactions.

% 然而，现有用户模拟器仍存在两个根本局限。
% L1：对齐局限于语言风格，而未涉及决策过程。主流方法以静态用户画像为条件，引导大语言模型生成符合人设的回复，使模拟器仅在表层语言上与真实用户对齐，而底层决策过程未得到对齐；但真实的多轮行为由交互历史、情境和决策状态共同驱动。
% L2：优化针对单轮回复，而非完整轨迹。近期方法采用面向逐轮回复的优化范式，因此在多轮交互中难以处理奖励稀疏与信用分配，也无法捕捉跨轮次的状态变化和决策连贯性。
% 因此，模拟反馈往往表现出行为不稳定和决策不一致，限制了其在高保真评测与训练中的用途。
However, existing user simulators still suffer from two fundamental limitations. \emph{(L1) Alignment is limited to linguistic style, not decision-making.} Mainstream methods condition LLMs on static user profiles to elicit persona-consistent responses~\citep{wang2025know,shi2025you,ye2025cpoaddressingrewardambiguity,kim2025few}, aligning simulators with real users only at the surface linguistic level while leaving the underlying decision-making process unaligned. \emph{(L2) Optimization is single-turn, not trajectory-level.} Recent methods adopt optimization paradigms tailored to turn-level responses~\citep{wang2025raiden,liao2025moa,zhang2026userlm,wu2026humanlm,dou-etal-2025-simulatorarena}, and thus inherently struggle with reward sparsity and credit assignment in multi-turn interactions, failing to capture state evolution and decision coherence. Consequently, simulated feedback often exhibits behavioral instability and inconsistent decision-making, limiting its utility for high-fidelity evaluation and training.

% 为共同应对这两个局限，我们提出面向用户模拟的轨迹对齐信用分配框架 \OurMethod{}。
% 该多轮用户模拟框架基于真实用户对话训练，关注两个互补维度：作为必要基础的语言风格一致性，以及作为核心目标的会话层面行为一致性。
% 为刻画后者，我们在模拟过程中引入意图状态，以描述用户潜在的决策倾向及其在多轮交互中的变化。
% \OurMethod{} 采用两阶段训练，逐步从表层语言对齐推进到决策层面的行为对齐。
To address these limitations, we propose \textbf{TR}ajectory-\textbf{A}ligned \textbf{C}redit Assignment for Us\textbf{ER} Simulation (\textbf{TRACER}), a multi-turn user simulator trained on real dialogues. TRACER combines linguistic grounding with behavioral alignment, using intent states inferred from observed dialogue as proxies for users' decision tendencies. In the first stage, supervised fine-tuning learns user-like responses and initializes structured intent prediction. In the second stage, the simulator interacts with a fixed assistant and learns from rewards measuring terminal agreement and trajectory alignment against paired human logs. 
Trajectory-level rewards are further decomposed to credit individual turns, using deviation magnitude and persistence to calibrate per-turn updates without overriding the session-level signal.
To evaluate behavioral fidelity, we organize 3,866 real customer-service sessions into 762 reference groups based on similar user conditions. Each group provides multiple observed trajectories, allowing evaluation to account for variation among comparable users and reducing dependence on a single reference interaction. We compare simulated and real behavior within these groups along three dimensions: session-level behavior, dialogue structure, and semantic consistency.
%This evaluation complements the paired-reference training objective by examining behavior across multiple real sessions.
We also assess generalization to unseen scenarios and perceived human-likeness. In a Turing study, annotators identified TRACER conversations at near-chance accuracy, suggesting high perceived naturalness.

Building on TRACER, we further introduce the Dynamic Marketing Benchmark (DM-Bench), which evaluates LLMs as conversational marketing agents through interactions with simulated users. DM-Bench jointly measures response quality and conversion outcomes, providing a behavioral perspective on agent performance. Within this simulated environment, models with higher response quality do not necessarily achieve higher conversion rates, motivating evaluation of both dimensions.

We summarize our contributions as follows:
% 1）提出 TRACER，通过结合时间轨迹对齐与终态结果一致性，对完整交互进行行为监督。
1) We propose \OurMethod{}, a two-stage user simulator that combines outcome and trajectory rewards with deviation-aware turn-level advantage modulation, which uses the magnitude and persistence of trajectory deviations to adjust per-turn updates while preserving session-level advantage direction.
% 2）提出轨迹对齐优势调制，利用共享对齐中偏差的大小与持续性，在保留会话级优势方向的同时调整逐轮更新。
2) We construct a multi-reference evaluation set with 762 groups and 3,866 real sessions, supporting assessment of session behavior, dialogue structure, and semantic consistency.
%fine-grained
% 3）在真实对话及未见领域上评测行为保真度，并通过 DM-Bench 将学得的行为反馈用于客服比较，同时提供人工策略响应评测证据。
3) We develop DM-Bench to jointly evaluate response quality and conversion outcomes in simulated marketing interactions, revealing a  misalignment between the two.

\section{Related Work}

\paragraph{LLM-based User Simulator.} Recent LLM-based user simulators have evolved from prompt-based role-playing toward deeper cognitive modeling. Early works often rely on explicit personas or static intent predictions, capturing surface linguistic patterns or individual responses but struggling to model dynamic intent evolution across multi-turn interactions \citep{park2024generative, wang2025know, kim2025few, kolluri2025finetuning}. 
Recent advances incorporate explicit reasoning and state modeling. HUMANLM \citep{wu2026humanlm} uses reinforcement learning to align latent states and generated responses with observed responses, while \citet{lu2025can} use reasoning-augmented supervised fine-tuning to predict the next user action from interaction histories. UserLM-R1 \citep{zhang2026userlm} improves reasoning and strategic behavior through multi-reward reinforcement learning over generated rationales and responses. Although these methods leverage multi-turn context, their optimization focuses on individual responses or next-step actions rather than aligning complete rollouts with human reference trajectories.
Overall, existing simulators fall short in capturing how continuous behavioral evolution shapes final dialogue outcomes.

%\paragraph{Applications of user simulator.}
%User simulators have been widely adopted in dialogue system research, primarily serving two purposes: training and evaluation.

%\textbf{Training.} User simulators provide scalable interactive environments for dialogue systems, alleviating the reliance on costly human annotation. One line of work leverages simulators to generate training data. For instance, CAMEL \citep{li2023camel} exploit LLM-based self-play to produce large-scale dialogue corpora for augmenting downstream systems. Another line of work embeds user simulators directly into RL loops. RLVER \citep{wang2025rlver} employs an emotion-aware user simulator to deliver multi-turn dynamic feedback for empathetic dialogue systems, while UserRL \citep{qian2025userrl}, DuCA\citep{yang2026harmonizing}, and CollabLLM\citep{wu2025collabllm} use simulated users to provide online reward signals for multi-turn RL training of assistant models.

%\textbf{Evaluation.} Beyond training, user simulators are increasingly used to overcome the limitations of static benchmark evaluation. SAGE \citep{zhang2025sentient} and UserBench\citep{qian2025userbench} simulate diverse user behaviors to assess dialogue systems under more realistic and varied interaction conditions. Although such approaches substantially improve the efficiency and coverage of evaluation, the reliability of simulated users remains a central bottleneck that constrains the credibility of simulator-based evaluation \citep{naous2025flipping}.

\paragraph{Applications of user simulator.}
User simulators are heavily utilized in dialogue system research for both training and evaluation, alleviating the reliance on costly human annotation. \textbf{Training:} Simulators provide scalable interactive environments, either by generating large-scale dialogue corpora \citep{li2023camel} or by supplying multi-turn dynamic feedback and reward signals in RL loops \citep{wang2025rlver, qian2025userrl, yang2026harmonizing, wu2025collabllm}. \textbf{Evaluation:} Simulators overcome the limits of static benchmarks by enabling diverse and realistic interaction assessments \citep{zhang2025sentient, qian2025userbench}. However, the reliability of these simulated users remains a central bottleneck constraining the credibility of simulator-based evaluation \citep{naous2025flipping}.
\section{Problem Formulation}
\label{problem_formulation}

\textbf{Interactive user simulation.} We formulate multi-turn user simulation as conditional sequential generation through interaction with a conversational assistant. Let \(u\) denote the user profile and task background available before simulation, and let \(h_t\) denote the interaction context at turn \(t\), including the dialogue history and the assistant’s latest utterance. The simulator policy generates

$$
(\hat{\ell}_t,\hat{y}_t)
\sim \pi_\theta(\cdot\mid u,h_t),
\qquad
\hat{\ell}_t=(\hat r_t,\hat z_t),
$$

% 其中，(\hat r_t) 表示自然语言形式的推理依据（rationale），(\hat z_t) 表示有序的意图分数（ordinal intent score），(\hat y_t) 表示用户回复。助手遵循策略 (\pi_A)，并根据不断演化的对话内容生成回复，从而形成下一轮用户交互的上下文。一次模拟会话在用户完成转化（conversion）、退出（dropout），或达到最大交互轮数时结束。
where \(\hat r_t\) is a natural-language rationale, \(\hat z_t\) is an ordinal intent score, and \(\hat y_t\) is the user’s response. The assistant follows a policy \(\pi_A\) and responds to the evolving dialogue, producing the context for the next user turn. A simulated session ends upon conversion, dropout, or reaching the maximum number of turns.

\textbf{Behavioral trajectory representation.} For each observed user response \(y_t\), we associate an annotation \(\ell_t=(r_t,z_t)\) consisting of a rationale and an intent score inferred from the dialogue. These annotations provide an interpretable proxy for the user’s expressed engagement and commitment. The intent score can increase, decrease, or remain unchanged as the interaction unfolds. We represent an annotated human session and a simulated session as

$$
\tau=\{(z_t,y_t)\}_{t=1}^{T},
\qquad
\hat\tau=\{(\hat z_t,\hat y_t)\}_{t=1}^{T^*},
$$

where \(T\) and \(T^*\) are their respective lengths, and \(z_T\) and \(\hat z_{T^*}\) denote their final intent scores. Terminal outcomes characterize the observable result of an interaction, such as whether the user provides contact information. This representation supports analysis of intent evolution, response content, and interaction length alongside terminal behavior.

\textbf{Behavioral modeling objective.} Let \(p_H(\tau\mid u,h_1,\pi_A)\) denote the conditional distribution of annotated human trajectories, and let \(p_\theta(\hat\tau\mid u,h_1,\pi_A)\) denote the trajectory distribution induced by the simulator interacting with the assistant. Our goal is to learn \(\pi_\theta\) that captures human behavioral dynamics under these conditions, including the distribution of terminal outcomes, the evolution of expressed intent, and the linguistic and temporal characteristics of the interaction. Observed human sessions provide empirical reference trajectories for learning; Section \ref{method} describes the resulting supervision and policy optimization. For evaluation, we aggregate human and simulated trajectories within groups of users with comparable pre-interaction contexts, with each rollout initialized from its own user’s context. The grouping procedure and behavioral metrics are specified in Section \ref{sec:experimental}.

%\input{latex/4_method}
% Replacement for latex/4_method.tex.
% Preamble dependencies: amsmath, amssymb, graphicx, natbib.
% The optional appendix algorithm below additionally requires float.
% Retains the original figure path and core method labels.
% See trail_section4_editorial_notes.md for implementation checks.
\section{TRACER}
\label{method}

Faithful multi-turn user simulation requires both natural responses and coherent behavior across an interaction. Relevant behavioral patterns include the final intent state, the sequence of intermediate intent transitions, and the pace at which the conversation progresses. Fitting individual responses provides no explicit trajectory-level constraint, while terminal supervision alone leaves intermediate behavior underdetermined. This motivates two design questions: how to evaluate behavioral consistency across a session, and how to use that feedback to guide updates at individual turns.

\begin{figure}[t]
    \centering
    \includegraphics[width=0.96\linewidth]{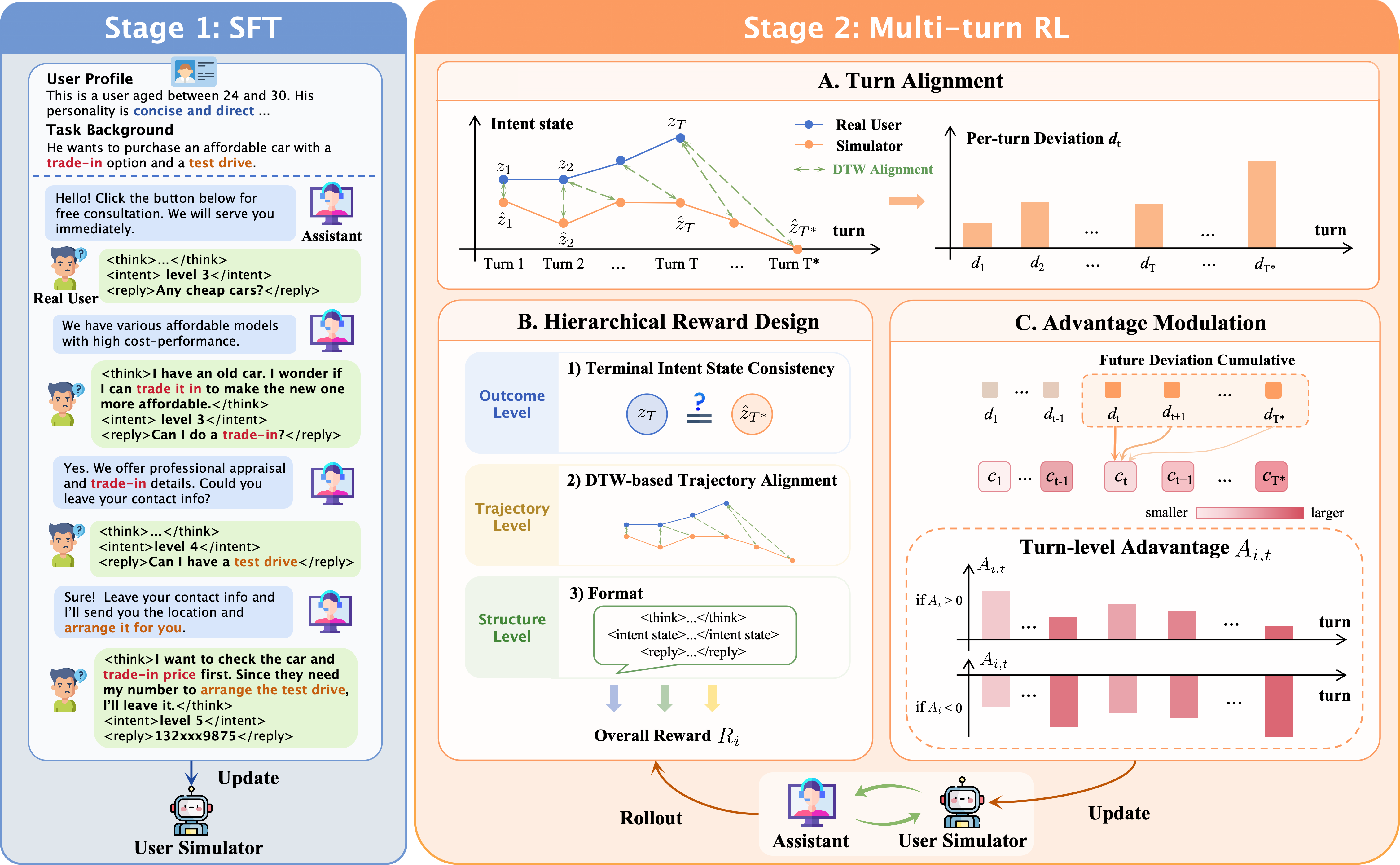}
    \caption{\OurMethod{} first learns from annotated human dialogues through supervised fine-tuning, then performs multi-turn reinforcement learning. A shared DTW alignment provides the trajectory cost for session-level rewards and local deviations for turn-level advantage modulation.}
    \label{fig:main}
\end{figure}

\OurMethod{} addresses these questions through two-stage training (Figure~\ref{fig:main}). Supervised fine-tuning initializes user-like language generation and structured intent prediction. Interactive reinforcement learning then combines outcome and trajectory rewards with deviation-aware advantage modulation. A shared trajectory alignment connects evaluation and optimization: its overall cost scores a completed interaction, while its local discrepancies inform turn-level updates.

\subsection{Simulator Initialization and Interactive Rollout}
\label{sec:simulator_rollout}

Following Section~\ref{problem_formulation}, the simulator generates a rationale $r_t$, intent score $z_t$, and user reply $y_t$ conditioned on user information $u$ and context $h_t$. We initialize these outputs by supervised fine-tuning on real dialogues and their derived annotations with a standard autoregressive likelihood objective.

To train under contexts induced by its own responses, the simulator then interacts with a fixed assistant policy $\pi_A$. Each rollout starts from the corresponding user information and initial assistant utterance, and continues until conversion, dropout, or the turn limit. The generated user reply advances the interaction with the assistant. For each input, we sample $G$ trajectories $\{\hat\tau_i\}_{i=1}^{G}$ of lengths $T_i^*$, scored against the paired human reference $\tau=\{(z_t,y_t)\}_{t=1}^{T}$.

\subsection{Session-Level Rewards}
\label{sec:hierarchical_reward}

Behavioral consistency has complementary outcome and process dimensions: a generated session should reflect the reference outcome as well as the intent evolution leading to it. We therefore combine a terminal-intent reward with a trajectory-alignment reward. Together, they provide supervision beyond local response similarity, while a format reward encourages valid structured outputs.

\paragraph{Outcome consistency.}
The final annotated intent state summarizes the reference session's outcome, providing a direct behavioral target beyond the wording of individual responses. We measure agreement with this state through the terminal-intent reward:
\begin{equation}
    R_{\mathrm{final},i}
    =\mathbb{I}\!\left(z_T=\hat z_{i,T_i^*}\right).
    \label{eq:trail_terminal_reward}
\end{equation}
This provides an empirical terminal-consistency signal. Terminal agreement alone does not constrain how the interaction unfolds: sessions with the same final intent may exhibit substantially different intent trajectories.

\paragraph{Trajectory alignment.}
\label{method:dtw_based_reward}
A direct turn-by-turn comparison can penalize comparable intent transitions simply because they occur at different conversational paces. We therefore use dynamic time warping (DTW)~\citep{dtw1,dtw2} to align the human and simulated intent sequences while preserving their temporal order. With local cost $|\hat z_{i,t}-z_s|$, let $\mathcal{P}_i$ denote the set of warping paths satisfying the standard boundary, monotonicity, and step constraints. The optimal path and its cumulative cost are
\begin{equation}
    P_i^*\in\arg\min_{P\in\mathcal{P}_i}
        \sum_{(t,s)\in P}|\hat z_{i,t}-z_s|,
    \qquad
    D_i=\sum_{(t,s)\in P_i^*}|\hat z_{i,t}-z_s|.
    \label{eq:trail_shared_alignment}
\end{equation}
Flexible alignment accommodates differences in pacing, but does not by itself require similar session lengths. We therefore combine the alignment cost with an explicit length penalty and a coefficient that discourages nearly constant intent sequences:
\begin{equation}
    R_{\mathrm{traj},i}
    =\exp\!\left[-\alpha\,
        \frac{D_i+\beta|T_i^*-T|}{T}\right]\eta_i.
    \label{eq:dtw_trajectory_reward}
\end{equation}
Here, $\alpha>0$ controls reward sensitivity, $\beta\geq0$ weights the length mismatch, and normalization by $T$ scales the cost relative to the reference length. The coefficient $\eta_i\in(0,1]$ discounts degenerate, nearly constant intent trajectories. Appendix~\ref{appendix:dtw_alignment} gives the alignment details.

Outcome and trajectory agreement do not ensure that individual outputs follow the required structure. We therefore include a format reward to check the prescribed output schema and basic field consistency. If $r^{\mathrm{fmt}}_{i,t}\in\{0,1\}$ is the turn-level format score, the combined session reward is
\begin{equation}
    \begin{aligned}
    R_{\mathrm{fmt},i}&=\frac{1}{T_i^*}
        \sum_{t=1}^{T_i^*}r^{\mathrm{fmt}}_{i,t}, &
    R_i&=\lambda_1R_{\mathrm{fmt},i}
        +\lambda_2R_{\mathrm{final},i}
        +\lambda_3R_{\mathrm{traj},i}.
    \end{aligned}
    \label{eq:overall_reward}
\end{equation}
The resulting scalar incorporates output validity, terminal agreement, and intermediate trajectory information. Reward hyperparameters are reported in Appendix~\ref{appendix:exper_details}.

\subsection{Deviation-Aware Policy Optimization}
\label{method:advantage_modulation}
\label{sec:trail_policy_optimization}

% The rewards above evaluate both outcomes and processes, but still reduce each session to a single scalar. Following GRPO~\citep{shao2024deepseekmath}, we standardize these rewards within the $G$ rollouts generated for the same input to obtain a trajectory-level advantage $A_i$. Broadcasting this advantage across the session does not distinguish turns with different alignment deviations. We therefore retain $A_i$ as the global reinforcement signal and modulate it using turn-specific discrepancies from the shared DTW alignment.
Following GRPO~\citep{shao2024deepseekmath}, we standardize rewards within each rollout group to obtain the trajectory advantage $A_i$. Broadcasting this scalar across turns does not reflect their different alignment deviations. We therefore retain its global reinforcement direction while adapting update strength using the shared DTW alignment.
\paragraph{Cumulative alignment deviation.}
We reuse the optimal DTW path $P_i^*$ from Eq.~\eqref{eq:trail_shared_alignment} to derive a local deviation for each generated turn. With $\mathcal{A}_{i,t}=\{s:(t,s)\in P_i^*\}$ denoting the reference turns aligned to generated turn $t$, we define
\begin{equation}
    d_{i,t}=\frac{1}{|\mathcal{A}_{i,t}|}
    \sum_{s\in\mathcal{A}_{i,t}}|\hat z_{i,t}-z_s|.
    \label{eq:trail_local_deviation}
\end{equation}
This averages the discrepancies over all reference turns aligned to the same generated turn. A local score describes agreement at one position, but does not indicate whether the subsequent interaction remains aligned. To incorporate this continuation information, we aggregate deviations over a discounted suffix of the sampled trajectory:
\begin{equation}
    c_{i,t}=\sum_{k=0}^{T_i^*-t}\rho^k d_{i,t+k},
    \qquad \rho\in[0,1].
    \label{eq:future_deviation_score}
\end{equation}
The discount $\rho$ controls how strongly later discrepancies affect the score: smaller values emphasize nearby turns, while larger values retain more information about the continuation. Setting $\rho=0$ recovers the local signal $d_{i,t}$.

\paragraph{Advantage modulation.}
A relatively high-reward session can still contain poorly aligned segments, while a low-reward session may exhibit varying degrees of deviation across turns. We use $c_{i,t}$ to adjust the strength of reinforcement or suppression within each session, with a bounded modulation that preserves the sign of the trajectory advantage:
\begin{equation}
    \widetilde A_{i,t}=w_{i,t}A_i,
    \qquad
    w_{i,t}=1-\lambda_{\mathrm{adv}}\operatorname{sign}(A_i)
        \tanh(\gamma c_{i,t}),
    \label{eq:advantage_modulation_weight}
\end{equation}
where $\lambda_{\mathrm{adv}}\in[0,1]$ bounds the modulation magnitude and $\gamma>0$ controls sensitivity. Equivalently, $\widetilde A_{i,t}=A_i-\lambda_{\mathrm{adv}}|A_i|\tanh(\gamma c_{i,t})$. Larger cumulative deviations reduce reinforcement when $A_i>0$ and increase the magnitude of suppression when $A_i<0$. The modulation prevents sign reversal, while $\lambda_{\mathrm{adv}}=0$ recovers the unmodulated trajectory advantage.

We use $\widetilde A_{i,t}$ in place of $A_i$ for all simulator tokens selected for optimization at turn $t$, excluding assistant and padding tokens. Alignment and advantage values remain fixed during each update. This preserves the GRPO clipped surrogate structure while changing the turn-level optimization signal.

\newcommand{\modelicon}[2]{\raisebox{-0.2\height}{\includegraphics[height=1em]{#1}}~#2}
\colorlet{HeatColor}{red}
\definecolor{MorandiRed}{HTML}{d19a8e} % 对应图中 TRACER
\definecolor{TableHeat}{HTML}{92a8d1} % 对应图中常用的淡蓝色，作为热力图基色
\definecolor{MorandiRed}{HTML}{C9B6F2}   % TRACER 主色
\definecolor{MorandiBlue}{HTML}{92A8D1}  % 基准模型主色

\section{Experiments}
\subsection{Experimental Setup \label{sec:experimental}}

\paragraph{Datasets.}
We construct CustomerService-Dialogue from real-world dialogues
between users and merchant-side customer service agents on a mobile
application, spanning industries such as healthcare and automotive. Each session contains
a 3--15-turn dialogue, a user profile (demographics, personality,
and initial intent), and merchant information. Conversion is defined
by whether the user provides contact information. Claude-4.5-Sonnet
provides weak turn-level supervision $\ell_t=(r_t,z_t)$, generating
a rationale and an intent score conditioned on the full dialogue
and user profile. Human assessment on a stratified sample shows
strong annotation agreement (Table~\ref{tab:agreement}).
The dataset contains 8,040 training sessions and 3,866 test
sessions organized into 762 reference groups. Each group contains
sessions from the same merchant, with similar personality traits
and initial user needs selected using semantic similarity thresholds.
Grouping is used only for evaluation; training retains paired-session
supervision. Construction and grouping details appear in
Appendix~\ref{appendix:dataset_construction}.
%Each instance contains: \textbf{(1) a multi-turn dialogue with 3 to 15 turns}, \textbf{(2) user profile information}, and \textbf{(3) merchant information}. The user profile includes demographic attributes, personality traits, and the initial intent. Sessions are labeled as conversion if contact information is provided, or dropout otherwise. 
%To obtain turn-level supervision for the variables $\ell_t = (r_t, z_t)$, we use Claude-4.5-Sonnet as a weak annotator that, conditioned on the full session context and user profile, generates for each user turn both a natural-language rationale and an intent score. 
%This weak-supervision pipeline yields turn-level annotations across the entire corpus at scale, eliminating the need for costly manual labeling while preserving consistency with the surrounding dialogue. 
% Human evaluation on a stratified sample shows strong agreement with LLM annotations (Table \ref{tab:agreement}).
% The dataset is split into 8,040 training and 1,000 test samples, with the test set balanced 1:1 for conversion vs. dropout. 
%The test set is balanced with a 1:1 ratio of converted and non-converted sessions to support stable and comparable evaluation. 
% Details are provided in Appendix~\ref{appendix:dataset_construction}.
\paragraph{Evaluation Protocol.} For each test instance, the simulated user model receives the user profile and the assistant's first utterance. It role-plays as the user and conducts a multi-turn interaction with the assistant until the dialogue terminates via conversion, dropout or reaching a maximum turns. The assistant is the same model deployed online, conditioned on the merchant information and user query, and follows a standardized SOP-driven policy that yields low-variance, reproducible responses. Metrics are computed over the simulated trajectories against the ground-truth dialogues.
% \paragraph{Assistant Model.}
% We use the same deployed customer service model as online, ensuring realistic interactions. The assistant follows a standardized SOP-driven policy, producing low-variance and reproducible responses. This controlled setup isolates user-side behavioral differences.

\paragraph{Metrics.}
We evaluate three dimensions.
\textbf{Session-Level Behavior}: PCR reports the overall conversion rate; ACC and conversion-class F1 measure outcome agreement with the original paired sessions.
\textbf{Group-Level Fidelity}: Group-$\Delta$PCR measures the absolute difference between simulated and real conversion rates within each group; W1-Turns measures the Wasserstein-1 distance between their dialogue-length distributions.
\textbf{Semantic Content}: OT-DTW measures optimal transport distance between simulated and real sessions within each group, using DTW-based semantic trajectory costs.
Group-based metrics are averaged across the 762 groups.

\paragraph{Baselines.}
%We consider representative baselines from three perspectives, covering mainstream paradigms for user simulation in both academia and industry.
We compare three categories of user simulators.
\textbf{General LLM baselines:}
We evaluate several general LLMs as zero-shot user simulators, including the Qwen2.5-7B-Instruct\citep{qwen2025qwen25technicalreport}, Qwen3-4B-Instruct\citep{yang2025qwen3}, DeepSeek-V3.2\citep{liu2025deepseek}, Gemini-3.1-Pro, Doubao-Seed-2.0-Pro and Claude-4.5-Sonnet. 
%These models rely solely on system prompts without task-specific training.
\textbf{Role-playing baselines:}
We include models explicitly designed for role-playing, including Doubao-Seed-Character, Qwen-Character, and CharGLM-4, which emphasize strong role consistency and conversational naturalness.
\textbf{Training-based baseline:}
We adopt HUMANLM~\citep{wu2026humanlm}, which introduces latent natural-language states aligned with responses via RL. We train it on our dataset using SFT and RL for fair comparison.

%We use the same customer service model as the one deployed online to ensure that simulated users interact with a standardized assistant under realistic conditions. The core behaviors of the assistant include answering user inquiries, asking targeted follow-up questions, and guiding users toward conversion-related actions such as leaving contact information.
%Notably, the assistant follows a standardized SOP-driven response policy, leading to low-variance and highly reproducible outputs across similar contexts. This controlled setup reduces confounding factors and ensures that performance differences are mainly driven by the user-side behaviors.

\paragraph{Implementation details.}
We instantiate \OurMethod{} using Qwen2.5-7B-Instruct and Qwen3-4B-Instruct. In the Stage 1, we train the models for 2 epochs with a learning rate of $10^{-5}$. In Stage 2, we train for 6 epochs, sample 8 rollout trajectories per prompt. All models use an inference temperature of 0.7. Full configurations, evaluation details, and stability analysis
are provided in Appendix~\ref{appendix:exper_details}.

\begin{table}[tbp]
%PCR target = 50.
\centering
\caption{Overall comparison of user simulators across three evaluation dimensions: Session-Level Behavior (PCR, ACC, F1), Group-Level Fidelity (Group-$\Delta$PCR, W1-Turns), and Semantic Content (OT-DTW). Groups are balanced 1:1; the session-level real PCR is 51.24\%. Best results are in bold, second-best are underlined.}
\label{tab:experimental_results}
\renewcommand{\arraystretch}{1.15}
\resizebox{0.90\textwidth}{!}{%
\begin{tabular}{l ccc ccc}
\toprule
\multirow{2}{*}{\textbf{Method}} & \multicolumn{3}{c}{\textbf{Session-Level Behavior}} & \multicolumn{2}{c}{\textbf{Group-Level Fidelity}} & \multicolumn{1}{c}{\textbf{Semantic Content}} \\
\cmidrule(lr){2-4} \cmidrule(lr){5-6} \cmidrule(lr){7-7}
& \textbf{PCR} & \textbf{ACC} $\uparrow$ & \textbf{F1} $\uparrow$ & \textbf{Group-$\Delta$PCR} $\downarrow$ & \textbf{W1-Turns} $\downarrow$ & \textbf{OT-DTW} $\downarrow$ \\
\midrule
\rowcolor{gray!15} \multicolumn{7}{c}{\textit{\textbf{General Baselines}}} \\
\modelicon{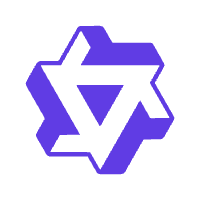}{Qwen2.5-7B-Instruct} & 36.7 & 53.7 & 47.4 & 0.456 & 5.383 & 0.500 \\
\modelicon{icons/qwen.png}{Qwen3-4B-Instruct} & 32.6 & 57.8 & 49.6 & 0.419 & 5.307 & 0.505 \\
\modelicon{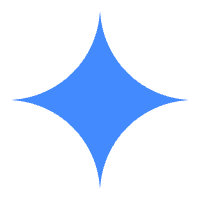}{Gemini-3.1-Pro} & 40.8 & 56.4 & 52.6 & 0.425 & 1.588 & 0.480 \\
\modelicon{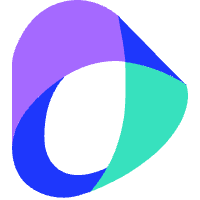}{Doubao-Seed-2.0-Pro} & 79.7 & 58.4 & 68.2 & 0.425 & 2.633 & 0.466 \\
\modelicon{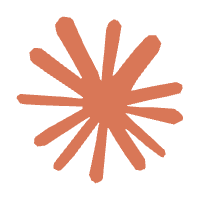}{Claude-4.5-Sonnet} & 5.8 & 50.7 & 13.7 & 0.473 & 2.278 & 0.480 \\
\modelicon{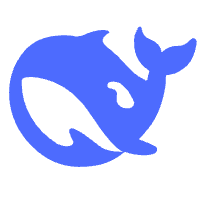}{Deepseek-V3.2} & 33.7 & 55.3 & 47.4 & 0.426 & 2.689 & 0.486 \\
\midrule
\rowcolor{gray!15} \multicolumn{7}{c}{\textit{\textbf{Role-playing Baselines}}} \\
\modelicon{icons/doubao.png}{Doubao-Seed-Character} & 26.2 & 51.3 & 37.0 & 0.471 & 7.063 & 0.531 \\
\modelicon{icons/qwen.png}{Qwen-Character} & 9.9 & 53.5 & 23.9 & 0.453 & 8.618 & 0.521 \\
\modelicon{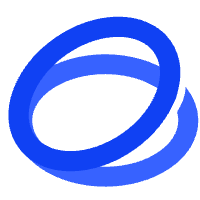}{CharGLM-4} & 16.6 & 51.5 & 28.5 & 0.470 & 7.549 & 0.529 \\
\midrule
\rowcolor{gray!15} \multicolumn{7}{c}{\textit{\textbf{Trained on Our Dataset}}} \\
HUMANLM-8B & 13.2 & 57.4 & 33.9 & 0.390 & \textbf{0.944} & 0.443 \\
\OurMethod-7B & 50.5 & \underline{79.2} & \underline{79.6} & \underline{0.165} & \underline{0.977} & \textbf{0.358} \\
\OurMethod-4B & 51.6 & \textbf{79.8} & \textbf{80.3} & \textbf{0.161} & 1.43 & \underline{0.366} \\
\bottomrule
\end{tabular}
}
\end{table}

\subsection{Main Results}
%Session-Level Behavior (PCR, ACC, F1), Dialogue Structure (MAE-Turns, $\Delta$ Tokens/Turn), and Semantic Content (Emb Sim, MAUVE)
Table \ref{tab:experimental_results} reports the performance of all methods along the three evaluation dimensions. We summarize our main findings as follows.

\textbf{General LLMs exhibit insufficient behavioral alignment,
with opposing tendencies toward over-compliance and over-refusal.}
Their PCR ranges from 5.8\% to 79.7\%
, compared with the real conversion rate
of 51.24\%, while ACC remains within 50--59\% and F1 reaches
at most 68.2\%. These results suggest that general LLMs tend
to be either overly willing or overly reluctant to convert,
failing to match the conversion behavior observed among real users.

\textbf{Role-playing baselines remain behaviorally misaligned
despite their focus on persona consistency.}
Doubao-Seed-Character, Qwen-Character, and CharGLM-4 exhibit
a pronounced under-conversion tendency, with PCRs of
9.9--26.2\% and F1 scores of 23.9--37.0\%. Their high
W1-Turns and OT-DTW further
reveal mismatches in dialogue length and semantic progression.
These results suggest that persona-oriented modeling alone
is insufficient to reproduce real users' conversion tendencies
and multi-turn interaction patterns.

\textbf{TRACER jointly achieves linguistic-style consistency and trajectory-level behavioral consistency.} 
\OurMethod{}-7B achieves a PCR of 50.5\% and outperforms
all baselines on ACC, F1, Group-$\Delta$PCR, and OT-DTW,
exceeding the strongest baseline by 11.4 F1 points.
Its W1-Turns performance ranks second, behind HUMANLM.
Among all evaluated models, the 4B variant achieves the
best performance on ACC, F1, and Group-$\Delta$PCR,
while the 7B variant performs best on OT-DTW.
These results support the framework's effectiveness
across both backbones. Compared with HUMANLM trained
on the same data using SFT and RL, both variants achieve
better outcome agreement and semantic trajectory
alignment, highlighting the value of jointly aligning
terminal outcomes and the course of interaction.
% \OurMethod{}-7B achieves a PCR of 50.5\%, together with
% the best ACC and F1, exceeding the
% strongest baseline by 11.4 F1 points.
% It also achieves the lowest Group-$\Delta$PCR and OT-DTW,
% while remaining competitive on W1-Turns. The 4B variant
% shows similar improvements, supporting the framework's
% effectiveness across both backbones. Compared with HUMANLM
% trained on the same data using SFT and RL, TRACER better
% captures real users' conversion tendencies and semantic
% trajectories. These results highlight the value of jointly
% aligning terminal outcomes and the course of interaction.
% TRACER-7B attains PCRs of 48.7, closely matching the 50\% ground-truth rate.
% \OurMethod-7B achieves an F1 of 85.6, surpassing the strongest baseline, Doubao-Seed-2.0-Pro, by more than 16 points.
% In parallel, our models also achieve leading  performance on MAE-Turns, token variation per turn, embedding similarity and MAUVE. A more controlled comparison against HUMANLM, trained on the same data with both SFT and RL, further isolates the advantage of our framework: despite sharing the linguistic substrate, HUMANLM still exhibits a severe calibration bias at the session-level behavior and a larger deviation in semantic similarity.
%showing that decision-level gains do not compromise conversational pacing or semantic fidelity.

%与在相同真实用户数据上进行sft和rl训练的HUMANLM进行更严格的对比，进一步揭示了我们轨迹级对齐的优势：HUMANLM在Session-level Behavior仍然存在严重的校准偏差（PCR 37.1，F1 76.0），且语义相似度偏差较大。

\subsection{Detailed Analysis}
\subsubsection{Human Evaluations}
To assess simulation fidelity, we conduct a Turing-style evaluation testing whether AI-simulated users are distinguishable from real humans. 
Annotators are given two dialogues with the same user profile information, one produced by a real user and the other by a simulator, and are asked to identify the human-generated dialogue.
We randomly sample 500 sessions from the test set, covering diverse personas and intents. Details about the annotators are provided in the Appendix \ref{app:annotators}.
%For each case, annotators are shown two dialogues sharing the \emph{same persona and intent}—one from a real user and one from a simulator—and asked to identify the human.
%To rigorously assess the simulation fidelity of the proposed model, we conduct a comprehensive human-centered Turing-style evaluation. The objective is to determine whether AI-simulated users can achieve indistinguishability from real human users under identical constraints.
%We random select 500 dialogue sessions from the test set, covering diverse personas and interaction intents. For each test case, we present human annotators with two parallel dialogues: one featuring a real human user and the other generated by a simulator, both sharing the \emph{same persona and underlying intent}. Annotators are tasked with identifying the real human user based on the consistency and naturalness of the interaction. This side-by-side setup eliminates the influence of task difficulty and focuses purely on the nuances of user behavior.
\paragraph{TRACER-7B achieves human-level indistinguishability.} 
\begin{wrapfigure}{r}{0.45\textwidth}
    \centering
    \vspace{-20pt}
    \includegraphics[width=0.42\textwidth]{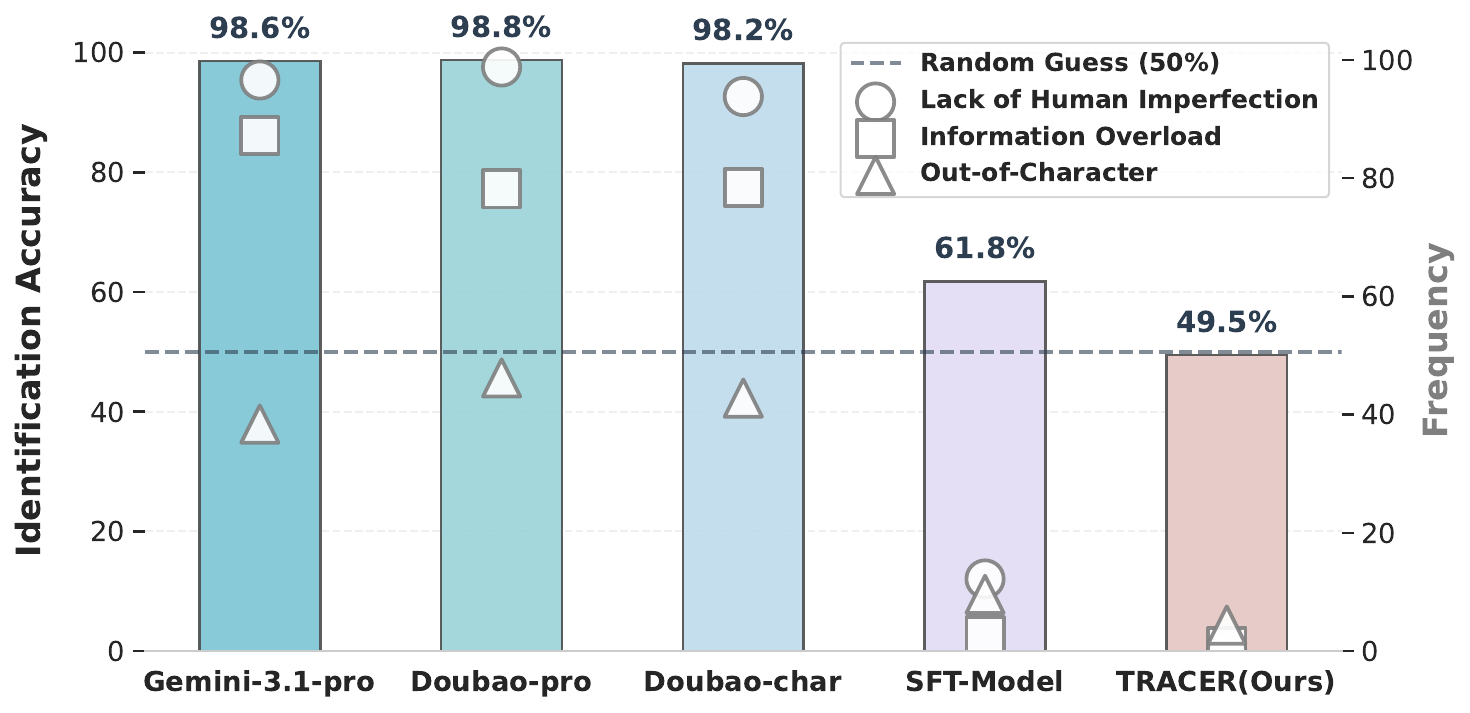}
    \caption{Human-indistinguishability evaluation. Accuracy is shown as bars, while the presence of non-human characteristics is indicated with markers. The dashed line at 50\% represents the random guess threshold.}
    \vspace{-20pt}
    \label{fig:turing_test_accuracy}
\end{wrapfigure}
Figure~\ref{fig:turing_test_accuracy} compares TRACER-7B with selected general-purpose and role-playing baselines from Table~\ref{tab:experimental_results}, as well as its SFT-only variant. General LLMs are correctly identified in over 98\% of cases, revealing persistent machine-like patterns. The SFT variant reduces identification accuracy to 61.80\% but still falls short of indistinguishability. In contrast, TRACER-7B reaches 49.50\%, converging to the random-guess threshold and indicating that it captures subtle human behavioral nuances missed by prior simulators.
%相比之下，TRACER-7B 达到了49.50%，接近随机猜测的阈值（50%），这表明它能够从语言风格以及行为选择上模拟真实人类。 

\paragraph{Why are baselines recognized?} Table~\ref{tab:turing_reason_full} groups detection cues into three dimensions. (1) \emph{Linguistic over-regularity}: baselines almost entirely lack human imperfection 
%(above 95\%)
and frequently show overly neat structure. (2) \emph{Mis-calibrated interaction rhythm}: they are consistently information-overloading and excessively cooperative,
%(both over 80\%) 
with Doubao variants further exhibiting mechanical patterns. (3) \emph{Out-of-character} behavior appears in roughly 40\% of cases, while higher-order defects such as emotional discontinuity remain rare. %TRACER-7B suppresses the dominant cues, cutting \emph{lack of human imperfection} to \textbf{3.0\%} and \emph{excessive cooperativeness} to \textbf{27.2\%}.
Details are in Appendix~\ref{app:turing_analysis}.
%To further investigate this performance gap, we provide a fine-grained diagnostic analysis in Table~\ref{tab:turing_reason_full}. As shown, the high detection rates of baselines are primarily attributed to a \textit{Lack of Human Imperfection} (>$93.8\%$) and \textit{Excessive Cooperativeness} (>$81.0\%$). TRACER-7B effectively mitigates these issues, reducing the frequency of such non-human traits to \textbf{3.0\%} and \textbf{27.2\%} respectively, thereby bridging the gap between simulated and real-world user interactions.
%TRACER-7B achieves an accuracy of 49.50%, which is closest to the random guess threshold (50%), demonstrating its superior capability in simulating indistinguishable human-like interactions compared to Gemini and Doubao (>$98\%$).

\subsubsection{Ablation Study}
\label{sec:ablation}
\newcommand{\std}[1]{{\scriptstyle \,\pm\, #1}}

\begin{table}[htbp] 
\vspace{-5pt}
\centering
\caption{Ablation study of TRACER-7B on session-level behavior, dialogue structure, and semantic content. Results are reported as mean $\pm$ standard deviation over five evaluation seeds.}
\label{tab:ablation_7B_mean_std}

\renewcommand{\arraystretch}{1.20}
\resizebox{\textwidth}{!}{
\begin{tabular}{l ccc cc c}
  \toprule

  \multirow{2}{*}{\textbf{Method}}
  & \multicolumn{3}{c}{\textbf{Session-Level Behavior}}
  & \multicolumn{2}{c}{\textbf{Group-Level Fidelity}}
  & \multicolumn{1}{c}{\textbf{Semantic Content}} \\
  \cmidrule(lr){2-4} \cmidrule(lr){5-6} \cmidrule(lr){7-7}
  & \textbf{PCR}
  & \textbf{ACC} $\uparrow$
  & \textbf{F1} $\uparrow$
  & \textbf{Group-$\Delta$PCR} $\downarrow$
  & \textbf{W1-Turns} $\downarrow$
  & \textbf{OT-DTW} $\downarrow$ \\

  \midrule

  \textbf{TRACER-7B}
  & $50.3 \pm 0.2$
  & $\mathbf{79.0} \pm 0.2$
  & $\mathbf{79.4} \pm 0.2$
  & $\mathbf{0.166} \pm 0.003$
  & $\mathbf{0.972} \pm 0.006$
  & $\mathbf{0.357} \pm 0.001$
  \\

  \quad $-$ w/o advantage modulation
  & $49.5 \pm 0.4$
  & $\underline{78.7} \pm 0.1$
  & $\underline{78.8} \pm 0.2$
  & $\underline{0.181} \pm 0.004$
  & $\underline{1.030} \pm 0.014$
  & $0.369 \pm 0.002$ \\

  \quad $-$ w/o $R_{traj}$
  & $52.8 \pm 0.3$
  & $77.4 \pm 0.3$
  & $78.3 \pm 0.3$
  & $0.187 \pm 0.005$
  & $1.036 \pm 0.020$
  & $0.373 \pm 0.001$ \\

  \quad $-$ w/o DTW alignment
  & $48.3 \pm 0.3$
  & $77.7 \pm 0.3$
  & $77.6 \pm 0.4$
  & $0.186 \pm 0.004$
  & $1.064 \pm 0.014$
  & $0.367 \pm 0.002$ 
   \\

  \quad $-$ w/o multi-turn RL
  & $52.8 \pm 0.4$
  & $74.3 \pm 0.3$
  & $75.3 \pm 0.2$
  & $0.218 \pm 0.004$
  & $1.081 \pm 0.007$
  & $\underline{0.365} \pm 0.001$ \\

  \quad $-$ w/o RL
  & $12.2 \pm 0.5$
  & $57.6 \pm 0.1$
  & $33.2 \pm 0.4$
  & $0.398 \pm 0.003$
  & $1.962 \pm 0.035$
  & $0.380 \pm 0.001$ \\

  \bottomrule
  \end{tabular}
}
\end{table}

Table \ref{tab:ablation_7B_mean_std} summarizes the ablation
results for \OurMethod{}. The full model achieves the best
mean performance across all five direction-specified metrics.
Removing advantage modulation or the trajectory-level reward
$R_{traj}$ consistently degrades behavioral accuracy,
group-level fidelity, and semantic alignment, supporting
the contribution of each component. Replacing DTW with
strict turn-by-turn matching also worsens all five metrics,
supporting the value of accommodating differences in
conversational pacing. Disabling multi-turn RL substantially
reduces behavioral accuracy and worsens Group-$\Delta$PCR,
highlighting the value of multi-turn optimization. Removing
RL entirely causes the largest overall degradation,
underscoring its importance beyond supervised fine-tuning.

\subsubsection{Out-of-Domain Generalization}
\begin{figure}[htbp]
    \centering
    \includegraphics[width=0.9\linewidth]{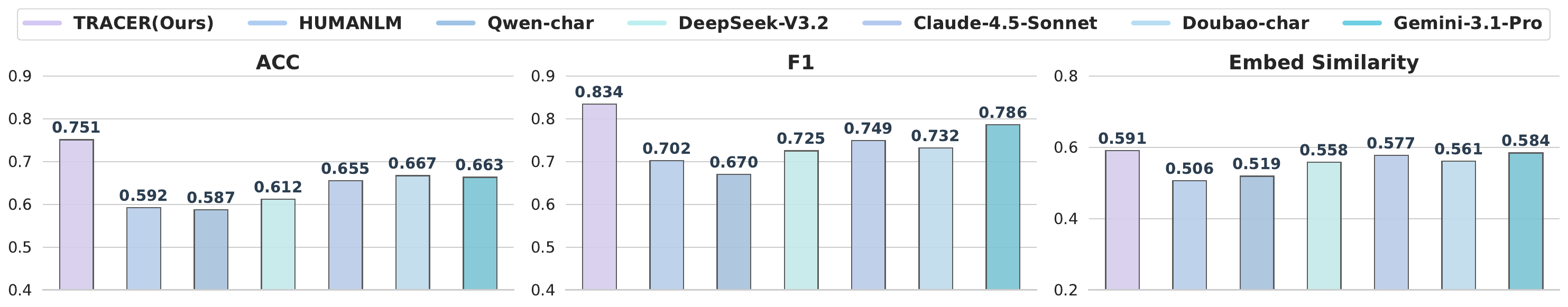}
    \caption{Performance Comparison of Models on the Out-of-Domain Dataset.}
    \label{fig:ood}
\end{figure}
%To evaluate the generalizability of our framework beyond the training distribution, we construct an out-of-domain (OOD) test set comprising dialogues from domains not seen during training (e.g., finance, real estate). The OOD set contains [X] sessions following the same annotation protocol as the in-domain test set, with a balanced 1:1 conversion ratio.
To assess TRACER's generalization beyond its training distribution, we evaluate it on CSC-Conv~\citep{zhu2026evaluating}, an open-source dataset of real user–agent conversations in financial customer service unseen during training.
For each dialogue, we extract the user's initial intent and a binary resolution label defining the session-level terminal state. We randomly sample 1{,}000 dialogues and compare TRACER with the strongest baselines from Table~\ref{tab:experimental_results} using terminal-state ACC, F1, and dialogue embedding similarity.
As shown in Figure~\ref{fig:ood}, TRACER performs best on all three metrics, outperforming the strongest baseline by +8.4, +4.8, and +0.7 points, respectively, despite no exposure to this domain. These outcome- and trajectory-level gains suggest TRACER learns transferable user-simulation behaviors rather than domain-specific patterns. Detailed settings are provided in Appendix~\ref{appendix:ood_details}.

% 动态营销基准
\section{Dynamic Marketing Benchmark}
\label{sec:dm-bench}

% TRACER 将模拟用户与人类交互轨迹对齐，同时关注中间意图变化与终态决策。
% 这两个层面的行为也为交互式客服评测提供支持：用户的中间反应影响对话如何展开，终态决策则提供结果反馈。
% 因此，我们使用训练后的模拟器，通过完整的多轮交互评测客服，并将这一评测设置具体构建为动态营销基准（DM-Bench）。
% 在共同的用户与商家条件下，DM-Bench 考察不同客服如何与模拟用户交互，并比较其回复质量与模拟转化结果。
% 我们进一步通过与人工策略响应预期的一致性，检验这一比较所依赖的行为反馈。
\OurMethod{} aligns simulated users with human trajectories, capturing intent changes and terminal decisions. This enables multi-turn assistant evaluation, where user responses shape interactions and decisions provide outcome feedback. We instantiate this as the Dynamic Marketing Benchmark, comparing response quality and simulated conversion under shared user and merchant conditions.
%We further examine whether the simulator's local intent changes agree with human expectations about assistant strategies.

\begin{figure}[htbp]
    \centering
    \begin{minipage}[b]{0.58\textwidth}
        \begin{center}
            \begin{subfigure}[b]{0.46\linewidth}
                \centering
                \includegraphics[width=0.95\linewidth]{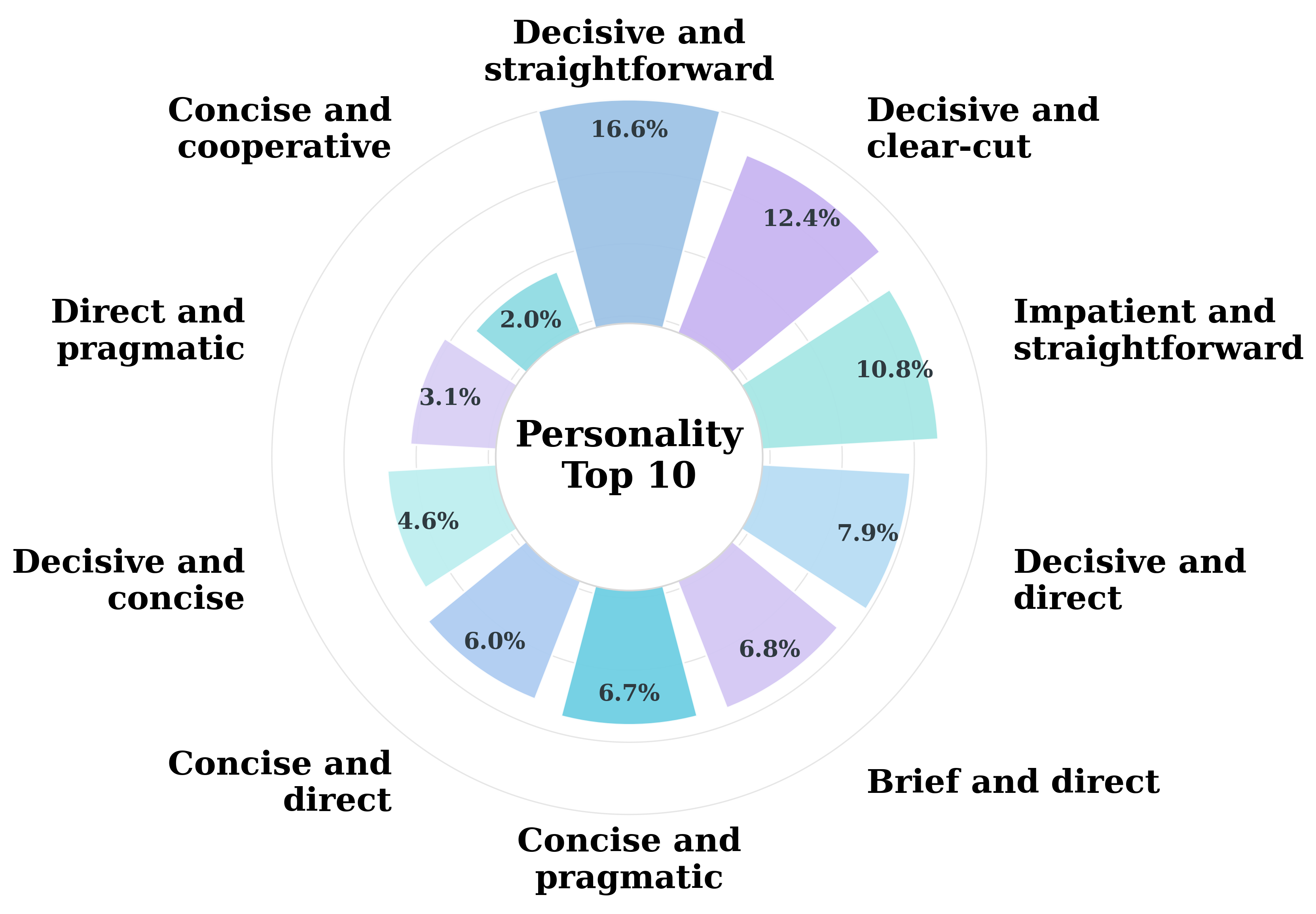}
                \captionsetup{font=footnotesize}
                % 性格特征
                \caption{Personality}
                \label{fig:c}
            \end{subfigure}
            \hfill
            \begin{subfigure}[b]{0.49\linewidth}
                \centering
                \includegraphics[width=0.98\linewidth]{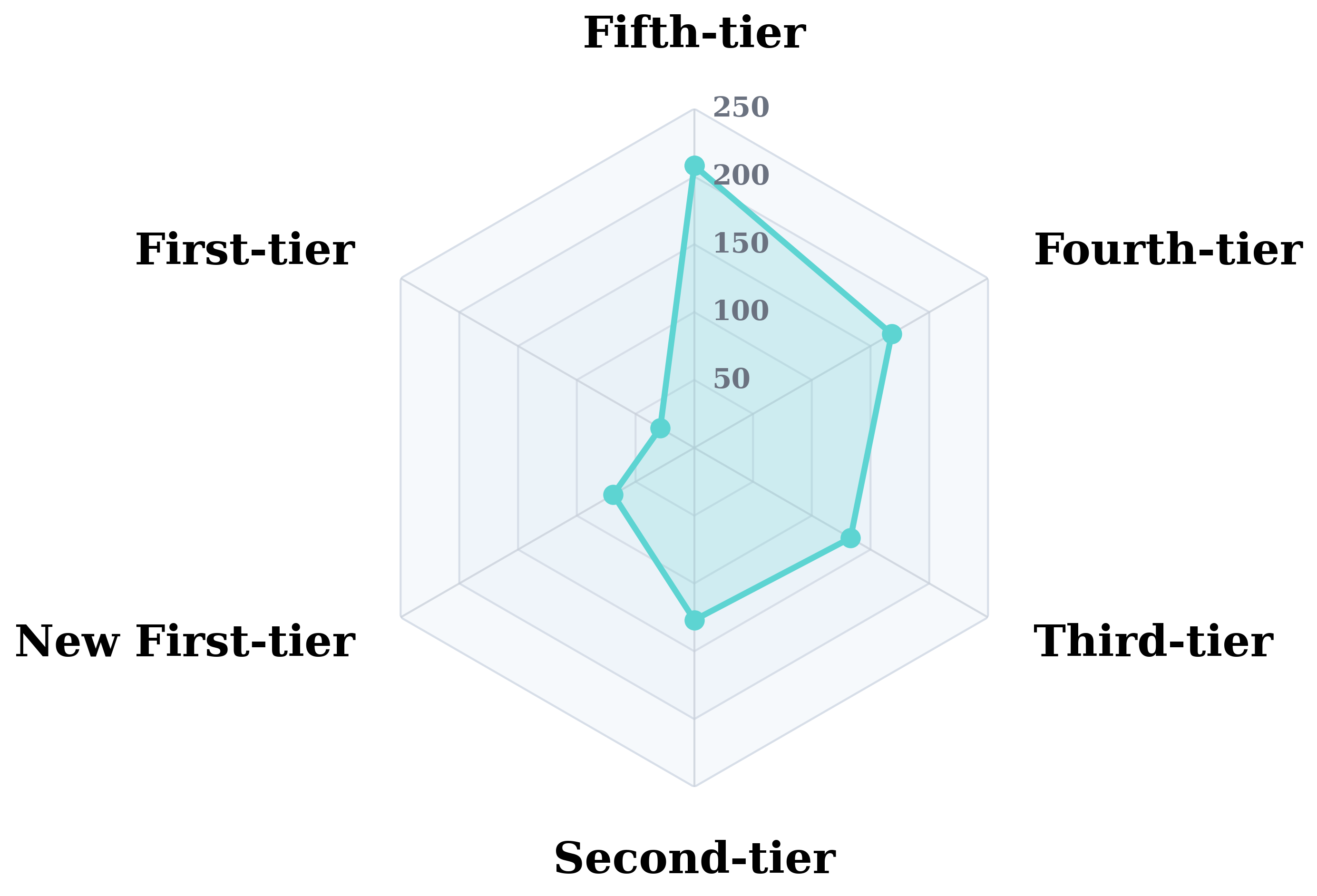}
                \captionsetup{font=footnotesize}
                % 城市分布
                \caption{City Profile}
                \label{fig:b}
            \end{subfigure}
        \end{center}
        \begin{center}
            \begin{subfigure}[b]{0.46\linewidth}
                \centering
                \includegraphics[width=0.98\linewidth, height=2cm]{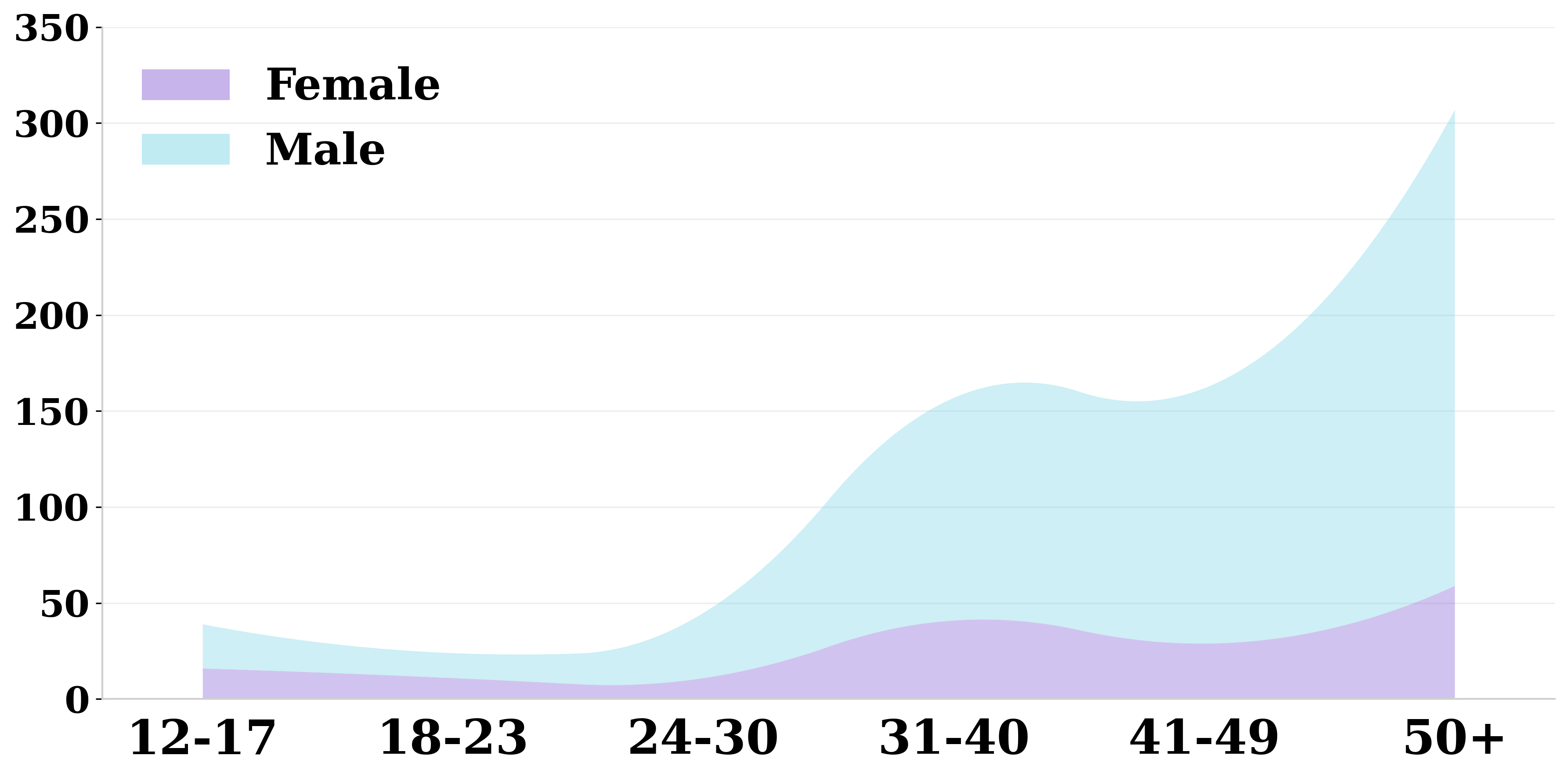}
                \captionsetup{font=footnotesize}
                % 人口统计特征
                \caption{Demographic Profile}
                \label{fig:a}
            \end{subfigure}
            \hspace{0.2em}
            \begin{subfigure}[b]{0.48\linewidth}
                \centering
                \includegraphics[width=0.85\linewidth]{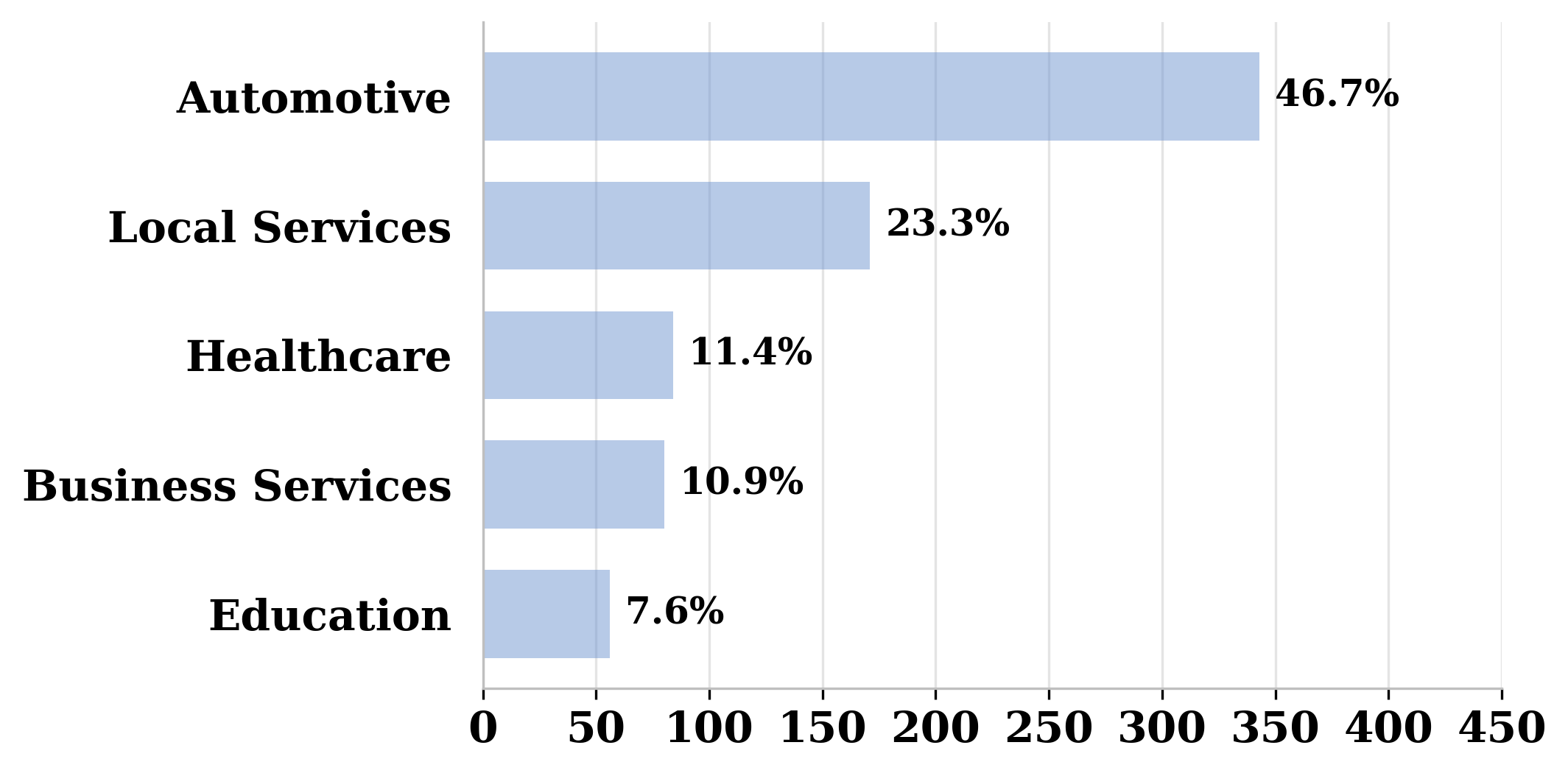}
                \captionsetup{font=footnotesize}
                % 商家分布
                \caption{Merchant Distribution}
                \label{fig:d}
            \end{subfigure}
        \end{center}
    \end{minipage}
    \hspace{0.1em}
    \begin{subfigure}[b]{0.4\textwidth}
        \centering
        \includegraphics[width=\linewidth, height=5.5cm]{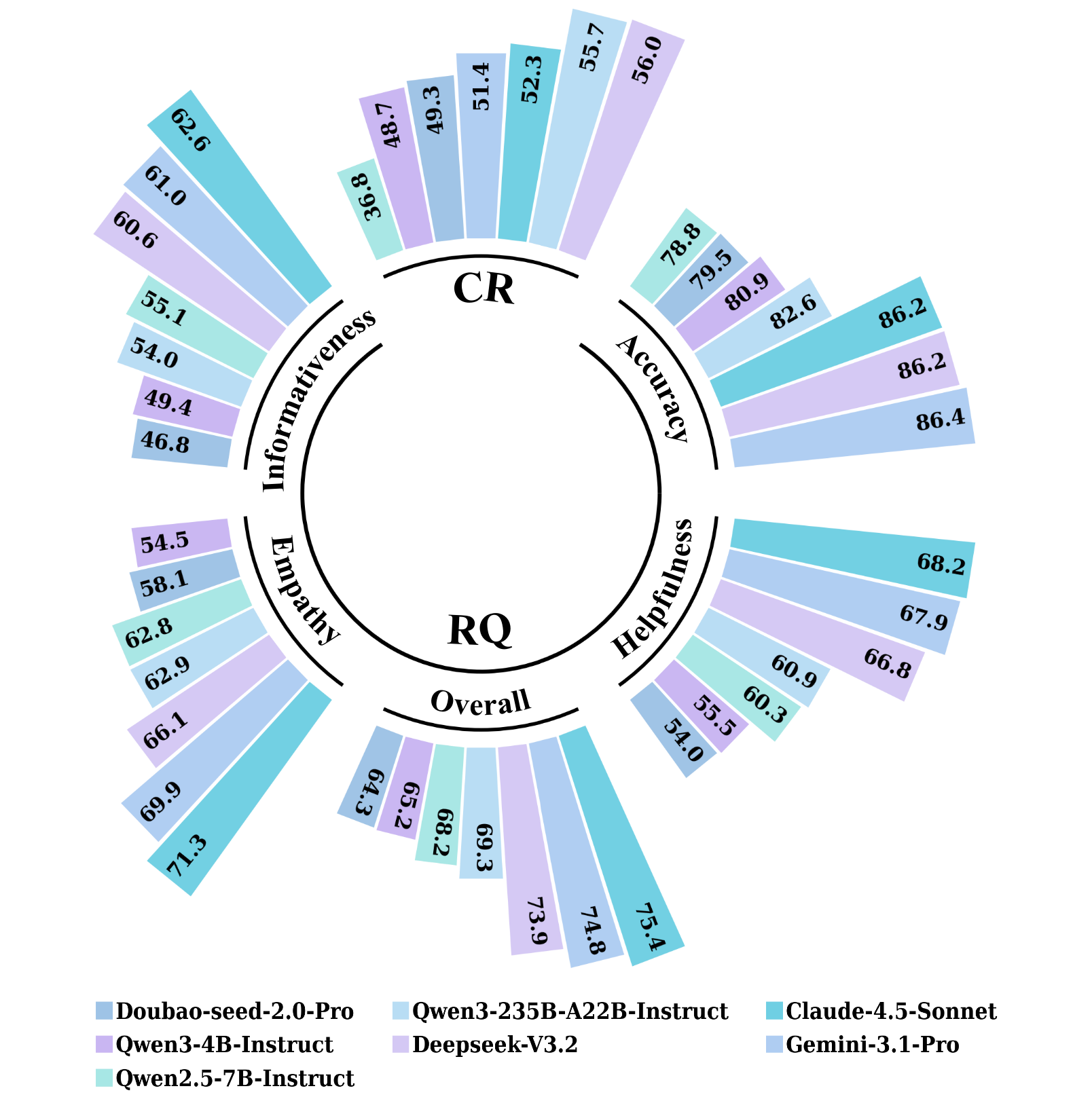}
        \captionsetup{font=footnotesize}
        % 大语言模型作为客服代理的表现
        \caption{LLM Performance as Customer Service Agents}
        \label{fig:e}
    \end{subfigure}
    % DM-Bench 概览：用户画像、商家分布与模型表现。
    \caption{Overview of DM-Bench: User Profiles, Merchant Distribution, and Model Performance.}
    \label{fig:total}
    \vspace{-0.9pt}
\end{figure}

% 基准构建
\subsection{Benchmark Construction}
\label{sec:dm-setup}

% DM-Bench 包含从 CustomerService-Dialogue 中抽取的 734 个实例，并保留该数据的经验分布。
% 每个实例包含人口统计属性、初始意图等用户画像信息，以及商家信息。
% 七个客服模型在同一组初始条件下与 TRACER 交互，后续对话则可以以不同方式展开。
% 图 \ref{fig:total} 概述用户与商家的分布。
% 数据来源和构建流程见第 \ref{sec:experimental} 节。
DM-Bench comprises 734 instances sampled from CustomerService-Dialogue. Each instance contains a user profile, including demographic attributes and initial intent, together with merchant information. Seven assistant models interact with TRACER under the same set of initial conditions, with user responses and intent evolving as each conversation unfolds. Figure~\ref{fig:total} summarizes the user and merchant distributions. The data source and construction protocol are described in Section~\ref{sec:experimental}.

We report two complementary metrics. \textbf{Simulated Conversion Rate (CR)} measures the proportion of sessions ending in conversion, as determined by the user simulator. \textbf{Response Quality (RQ)} follows the LLM-as-a-judge protocol of \citet{zhu2026evaluating}, assessing multi-turn responses in terms of accuracy, helpfulness, informativeness, and empathy. Together, these metrics capture response quality and terminal outcomes within the simulated environment.

% 我们报告两个互补指标。
% 模拟转化率（CR）衡量由用户模拟器判定为转化的会话比例，汇总模拟环境中的终态结果。
% 回复质量（RQ）遵循 Zhu 等（2026）的 LLM 评审协议，从准确性、有帮助程度、信息充分性和共情能力评价多轮回复。
% 两项指标共同考察客服回复及完整交互所产生的模拟用户结果。

% 行为反馈验证
% \subsection{Validation of Behavioral Feedback}
% \label{sec:dm-strategy-responses}
% 为检验 DM-Bench 所使用的行为反馈，我们评估模拟意图变化是否符合人类对客服策略的预期。
% 我们从不同客服模型中抽取 200 段对话，覆盖六类策略：共情、紧迫感、利益／折扣表述、信息帮助、联系方式请求，以及追问／澄清。
% 三名标注员独立标注客服发言后意图的预期变化方向，即下降、不变或上升。
% 参考标签由多数投票确定，标注员两两一致率超过 90%。
% To examine the behavioral feedback used by DM-Bench, we evaluate whether simulated intent changes agree with human expectations about assistant strategies. We sample 200 conversations across assistant models, covering six strategy categories: empathy, urgency, benefit/discount framing, informative help, contact-information requests, and probing/clarification. Three annotators independently label the expected direction of intent change following assistant turns as decrease, unchanged, or increase. Reference labels are determined by majority vote, with pairwise inter-annotator agreement above 90\%.
\noindent\textbf{Behavioral validation.}
We validate feedback through turn-level agreement with human expectations and assistant-ranking agreement with human role-play.
For intent changes, three annotators label 449 instances from 200 conversations across six strategy types (pairwise agreement $>90\%$).
Against majority-vote labels, TRACER achieves recalls of 88.7\%, 74.2\%, and 81.0\% for decrease, unchanged, and increase, respectively (Appendix~\ref{app:dm-strategy-evaluation}).
Human role-play on 100 contexts covers seven assistants (Table~\ref{tab:human-cr-validation}).
Human and TRACER conversion-rate rankings show strong agreement (Spearman's $\rho=0.955$; Kendall's $\tau_b=0.878$).
These results support the plausibility of TRACER's local intent changes and its agreement with human role-play in ranking assistants.

\subsection{Results and Findings}
We report the results in Figure \ref{fig:total}, Table~\ref{tab:dm_bench_full_results} and Table~\ref{tab:cr_rq_spearman}, with the following conclusions: 
% Figure~\ref{fig:total} and Tables~\ref{tab:dm_bench_full_results} and~\ref{tab:cr_rq_spearman} show that \textbf{response quality and simulated conversion yield different model rankings}. Across the seven models, RQ and CR exhibit a positive but statistically inconclusive rank association (Spearman $\rho = 0.57$, $p = 0.18$, $n = 7$). Claude-4.5-Sonnet achieves the highest RQ, whereas DeepSeek-V3.2 and Qwen3-235B achieve higher simulated CR. These ranking differences motivate reporting both metrics: DM-Bench complements response-quality evaluation with terminal outcomes from complete multi-turn interactions.
\textbf{(1) Response quality is not a reliable proxy for conversion effectiveness.} RQ and CR show only moderate rank correlation across seven models (Spearman $\rho=0.57$, $p=0.18$, $n=7$). For instance, Claude-4.5-Sonnet ranks highest on RQ but is surpassed in CR by DeepSeek-V3.2 and Qwen3-235B, indicating that response quality does not fully capture persuasion effectiveness.
\textbf{(2) Even frontier LLMs face a substantial conversion ceiling.} Despite high RQ scores, top models still fail to convert many users, suggesting limited adaptation to user intent and decision dynamics across turns.

These findings show the value of DM-Bench as a user-centered, outcome-driven evaluation protocol: it reveals performance gaps that are largely invisible under quality-only benchmarks, especially the gap between generating good-sounding responses and generating decision-moving ones.
\section{Conclusion}

In this work, we present TRACER, a multi-turn user simulator aligned with real users at both linguistic and decision-making levels, trained via a two-stage paradigm of SFT and trajectory-aligned RL.
%with a hierarchical reward and trajectory-aligned advantage modulation
Building on TRACER, we further construct the Dynamic Marketing Benchmark, which jointly measures persuasion effectiveness and response quality. Experiments show that TRACER largely outperforms existing simulators and achieves near-human fidelity in Turing tests, while revealing a misalignment between response quality and persuasion effectiveness in mainstream LLMs. In future work, we plan to extend TRACER to broader interactive scenarios and explore its use as a training environment for dialogue agents.

\section*{AI use statement}

% (This section is \textbf{required} and does not count toward the page limit.)

% In this work, we used generative AI tools for [tasks with required disclosure].
% We have not used generative AI tools for [other tasks with required disclosure],
% and [the rest of the required disclosure tasks] are not applicable to this work.
% Additionally, we used generative AI tools for [tasks with recommended
% disclosure]. We have reviewed all AI-assisted work. [Elaborate. For example, “we
% checked LLM-generated research ideas for potential plagiarism through a manual
% literature survey”, “LLM-generated code was verified and tested for correctness
% by 2 authors”, etc.]. We take responsibility for the final content of this work,
% including text, claims or artifacts produced with the aid of generative AI.

% See the ICLR 2027 AI Policy for Authors for more details. This statement should
% not be more than 1 page.
\label{llm_use_statement}
During the preparation of this study, LLMs were utilized solely as auxiliary tools. The application of these models was strictly limited to linguistic refinement, including grammar and spelling corrections, as well as technical support for drafting and debugging code during the experimental phase. All suggestions provided by the models underwent rigorous auditing, revision, and verification by the authors. It must be explicitly stated that LLMs played no role in the conceptualization of the research problem, the development of the theoretical framework, the design of experimental protocols, or the scientific interpretation of the results. All core academic contributions such as the research logic, methodological design, and final conclusions were independently completed by the whole authors.

\section*{Ethics statement}

% (This section is \textbf{recommended} and does not count toward the page limit.)

% If authors feel that their paper submission raises questions regarding the Code
% of Ethics, they are encouraged to include a paragraph of Ethics Statement (at
% the end of the main text before references) to address potential concerns where
% appropriate. Topics include, but are not limited to, studies that involve human
% subjects, practices to data set releases, potentially harmful insights,
% methodologies and applications, potential conflicts of interest and sponsorship,
% discrimination/bias/fairness concerns, privacy and security issues, legal
% compliance, and research integrity issues (e.g., IRB, documentation, research
% ethics). This statement should not be more than 1 page.
All user behavior data involved in this study were sampled and processed in strict accordance with relevant laws, regulations, platform guidelines, and privacy protection requirements. The scope of this research is limited exclusively to data resources authorized for scientific analysis and benchmark construction. Prior to the release of the public benchmark, the raw data underwent multiple stages of quality control, representative sampling, and de-identification procedures. Sensitive information possessing identifying characteristics, such as names, contact details, and specific addresses, has been either deleted or replaced with semantic placeholders. Furthermore, this study implemented filters for inappropriate or potentially harmful content and conducted manual audits to ensure that the processed data meet ethical standards. All analyses and experiments were performed using the desensitized and cleaned dataset. The authors assume full legal and academic responsibility for ensuring that the construction and utilization of the benchmark data comply with all relevant data governance and privacy protection norms.

\section*{Reproducibility statement}

To support reproducibility of our experiments, we provide the \OurMethod{} code at~\url{https://github.com/ChenGeng0102/TRACER} and document the experimental procedures in the paper and appendices. Section~\ref{method} specifies the training objectives and optimization procedure, while Appendix~\ref{appendix:exper_details} reports the hardware, training and inference configurations, and hyperparameter settings. Appendix~\ref{appendix:dataset_construction} describes data construction and preprocessing, and Appendix~\ref{appendix:prompt} provides the prompts used for annotation and simulation. The evaluation protocol is presented in Section~\ref{sec:experimental}, with additional details on out-of-domain evaluation, DM-Bench, and human annotation in Appendices~\ref{appendix:ood_details}–~    \ref{sec:human_annotation}.

% To support reproducibility of our experiments, we provide the \OurMethod{} code at~\url{https://anonymous.4open.science/r/TRACER_code-0A27} and document the experimental procedures in the paper and appendices. Section~\ref{method} specifies the training objectives and optimization procedure, while Appendix~\ref{appendix:exper_details} reports the hardware, training and inference configurations, and hyperparameter settings. Appendix~\ref{appendix:dataset_construction} describes data construction and preprocessing, and Appendix~\ref{appendix:prompt} provides the prompts used for annotation and simulation. The evaluation protocol is presented in Section~\ref{sec:experimental}, with additional details on out-of-domain evaluation, DM-Bench, and human annotation in Appendices~\ref{appendix:ood_details}–~    \ref{sec:human_annotation}.



\bibliography{iclr2027_conference}
\bibliographystyle{iclr2027_conference}

\newpage
\appendix
\section*{Appendix}

\section{Experimental Details}
\label{appendix:exper_details}
\raggedbottom
\subsection{Implementation Details}
\label{appendix:implementation_details}

\paragraph{Training configuration.}
% 我们所有训练实验均在 8 × NVIDIA  H200 GPUs 的硬件环境上进行。模型方面，本研究主要采用 Qwen2.5-7B-Instruct 【参考文献】和 Qwen3-4B-Instruct 【参考文献】作为基础模型，并分别进行监督微调与多轮强化学习训练。
All training experiments were conducted on a hardware platform equipped with 
$8 \times$ NVIDIA H200 GPUs. For the backbone models, we mainly adopted 
Qwen2.5-7B-Instruct~\citep{qwen2025qwen25technicalreport} and Qwen3-4B-Instruct~\citep{yang2025qwen3} as the base models, and trained them through supervised fine-tuning and multi-turn reinforcement learning, respectively.

% 在监督微调阶段，我们基于 LLaMA-Factory 【参考文献】实现 SFT 训练流程，并使用 DeepSpeed ZeRO-2 进行分布式训练。最大上下文长度设置为 4096 tokens，每张 GPU 的 batch size 设置为 4，梯度累积步数为 2。模型训练轮数为 2 个 epoch，学习率设置为 1.0e-5，并采用 cosine 学习率调度器，warmup ratio 设置为 0.1。所有 SFT 实验均使用 bf16 精度训练，以提升训练效率并降低显存开销。
During the supervised fine-tuning stage, we implemented the SFT training pipeline based on LLaMA-Factory~\citep{zheng2024llamafactory} and used DeepSpeed ZeRO-2 for distributed training. The maximum context length was set to 4096 tokens. The per-GPU batch size was set to 4, with a gradient accumulation step of 2. The models were trained for 2 epochs with a learning rate of $1.0 \times 10^{-5}$. We adopted a cosine learning rate scheduler with a warmup ratio of 0.1. All SFT experiments were conducted using bf16 precision to improve training efficiency and reduce GPU memory consumption.

% 在多轮强化学习阶段，我们基于 verl 框架【参考文献】进行修改，构建了多轮rl交互训练框架，以适配用户模拟器的多轮对话学习任务。全局训练 batch size 设置为 32，mini-batch size 设置为 32，每张 GPU 的 micro-batch size 设置为 4。每个 prompt 采样 8 条 rollout 轨迹，训练总 epoch 数设置为 6，采样温度设置为 1.0。输入侧最大 prompt 长度设置为 6144，最大 response 长度设置为 2048，并限制rollout 阶段的最大对话轮数设置为 15。在上述配置下，模型多轮rl训练约需12小时。
During the multi-turn reinforcement learning stage, we modified the verl framework~\citep{sheng2024hybridflow} to construct a multi-turn RL interaction training framework tailored to the user simulator dialogue learning task. The global training batch size was set to 32, the mini-batch size was set to 32, and the per-GPU micro-batch size was set to 4. For each prompt, we sampled 8 rollout trajectories. The total number of training epochs was set to 6, and the sampling temperature was set to 1.0. On the input side, the maximum prompt length was set to 6144 tokens, while the maximum response length was set to 2048 tokens. We further limited the maximum number of dialogue turns during rollout to 15. Under the above configuration, multi-turn RL training took approximately 12 hours.

\paragraph{Inference configuration.}
During the inference stage, the temperature was set to 0.7 for both the baseline models and our model to ensure a consistent decoding configuration across all compared methods. Following the dataset setting, the maximum number of interaction turns was capped at 15 during inference, and sessions that did not reach conversion within this limit were treated as dropout.

\paragraph{Evaluation details.}
Across all evaluations, we employ a unified computational setup to ensure comparability among different methods. All token-related metrics are computed using the Qwen2.5-7B-Instruct tokenizer. 

Group-$\Delta$PCR is the mean absolute difference between
simulated and real conversion rates across reference groups:
$\mathrm{Group}\text{-}\Delta\mathrm{PCR}
= G^{-1}\sum_{g=1}^{G}|\hat{p}_g-p_g|$,
where $\hat{p}_g$ and $p_g$ denote the simulated and real
session-level conversion rates in group $g$.
Groups receive equal weight, and the metric is reported
as a proportion rather than a percentage.

W1-Turns averages the Wasserstein-1 distance between the
empirical distributions of real and generated user-turn
counts within each group, giving all groups equal weight.
It is measured in user turns. Real turns are counted from
non-empty user messages after merging consecutive messages
from the same speaker; generated turns are counted from
the interaction log, including valid empty exit replies.

For OT-DTW, non-empty user replies are encoded using
bge-base-zh-v1.5 and L2-normalized, excluding
reasoning text and assistant replies. The local cost is
$\max(0,1-\cos(\mathbf{e}_i,\mathbf{e}_j))$.
For each generated--real trajectory pair, we find the
minimum cumulative-cost DTW path and divide its cost by
its length, rather than directly minimizing average cost.
The cost is 1 if exactly one trajectory is empty and 0
if both are empty. Within each group, the Hungarian
algorithm finds a minimum-cost one-to-one matching over
the pairwise trajectory costs. Matched costs are averaged
within groups, followed by an equally weighted average
across groups. Lower values indicate better agreement
for all three metrics.

For Embedding Similarity in the out-of-domain evaluation, we utilize bge-base-zh-v1.5 as the encoding model to map both simulated and ground-truth user replies into normalized embeddings, measuring their semantic consistency via cosine similarity.
% For MAUVE, we also use bge-base-zh-v1.5 to encode replies. Except for the input features and computation device, we keep the random seed, number of clusters, and other internal hyperparameters of MAUVE at their default settings.

\paragraph{Reward and advantage hyperparameters.}
Unless otherwise specified, we use the same reward and advantage modulation configurations across both model scales. All trajectory-level rewards are computed at the session level and propagated to valid user turns via the proposed advantage modulation strategy. For the DTW-based trajectory reward in Eq.~\eqref{eq:dtw_trajectory_reward}, we set the path-deviation sensitivity coefficient to $\alpha=1.0$ and the length-penalty coefficient to $\beta=1.0$. The flattening coefficient is set to $\eta=0.2$ when the predicted intent trajectory degenerates into an overly flat sequence, and $\eta=1.0$ otherwise. For the hierarchical reward in Eq.~\eqref{eq:overall_reward}, we set $(\lambda_1,\lambda_2,\lambda_3)=(0.2,0.5,0.3)$ for the format reward, terminal intent consistency reward, and trajectory alignment reward, respectively. For the cumulative future deviation score in Eq.~\eqref{eq:future_deviation_score}, we set the future-deviation discount factor to $\rho=0.4$. For the turn-level advantage modulation in Eq.~\eqref{eq:advantage_modulation_weight}, we set the modulation strength to $\lambda_{\mathrm{adv}}=0.2$, the deviation sensitivity coefficient to $\gamma=0.5$, and the numerical stability term to $\epsilon=10^{-6}$. All coefficients were selected on a held-out validation split.

\subsection{Sensitivity Analysis of Hyperparameters in Multi-Turn Advantage Modulation}
\label{appendix:hyperparameters_analysis}

\begin{table*}[t]
\centering
\caption{
Hyperparameter sensitivity of TRACER-7B.
Each configuration is evaluated using its corresponding fixed checkpoint
with five evaluation seeds (42--46); hyperparameter variants correspond
to separately trained checkpoints.
Results are mean $\pm$ standard deviation across evaluation seeds,
rather than independent training runs.
PCR, ACC, and F1 are reported in percent.
The reference PCR is 51.24\%.
Bold indicates the best mean for each directional metric.
}
\label{tab:hyper_stability}
\setlength{\tabcolsep}{3.5pt}
\renewcommand{\arraystretch}{1.15}
\resizebox{\textwidth}{!}{
\begin{tabular}{lccc ccc ccc}
\toprule
\textbf{Configuration}
& $\lambda_{\mathrm{adv}}$
& $\gamma$
& $\rho$
& \textbf{PCR}
& \textbf{ACC} $\uparrow$
& \textbf{F1} $\uparrow$
& \textbf{Group-$\Delta$PCR} $\downarrow$
& \textbf{W1-Turns} $\downarrow$
& \textbf{OT-DTW} $\downarrow$ \\
\midrule

Without advantage modulation
& -- & -- & --
& $49.54\pm0.44$
& $78.68\pm0.13$
& $78.85\pm0.19$
& $0.181\pm0.004$
& $1.030\pm0.014$
& $0.369\pm0.002$ \\

\textbf{Default TRACER-7B}
& 0.2 & 0.5 & 0.4
& $50.29\pm0.20$
& $79.04\pm0.19$
& $79.36\pm0.21$
& $0.166\pm0.003$
& $\textbf{0.972}\pm0.006$
& $\mathbf{0.357}\pm0.001$ \\

\midrule
$\rho=0.0$
& 0.2 & 0.5 & 0.0
& $50.52\pm0.32$
& $78.36\pm0.20$
& $78.74\pm0.15$
& $0.176\pm0.002$
& $0.994\pm0.012$
& $0.373\pm0.001$ \\

$\rho=0.1$
& 0.2 & 0.5 & 0.1
& $47.21\pm0.22$
& $79.06\pm0.24$
& $78.73\pm0.26$
& $0.164\pm0.003$
& $0.975\pm0.007$
& $0.369\pm0.001$ \\

$\rho=0.3$
& 0.2 & 0.5 & 0.3
& $48.66\pm0.19$
& $78.55\pm0.29$
& $78.53\pm0.28$
& $0.170\pm0.004$
& $0.977\pm0.017$
& $0.365\pm0.001$ \\

$\rho=0.5$
& 0.2 & 0.5 & 0.5
& $53.86\pm0.16$
& $78.42\pm0.24$
& $79.46\pm0.23$
& $0.177\pm0.003$
& $0.987\pm0.009$
& $0.365\pm0.002$ \\

\midrule
$\lambda_{\mathrm{adv}}=0.1$
& 0.1 & 0.5 & 0.4
& $47.78\pm0.32$
& $\mathbf{79.80}\pm0.20$
& $\mathbf{79.60}\pm0.25$
& $\mathbf{0.160}\pm0.002$
& $1.059\pm0.012$
& $0.382\pm0.002$ \\

$\lambda_{\mathrm{adv}}=0.3$
& 0.3 & 0.5 & 0.4
& $52.16\pm0.25$
& $78.47\pm0.29$
& $79.18\pm0.25$
& $0.172\pm0.005$
& $1.101\pm0.024$
& $0.371\pm0.001$ \\

\midrule
$\gamma=0.4$
& 0.2 & 0.4 & 0.4
& $50.82\pm0.18$
& $78.67\pm0.11$
& $79.10\pm0.07$
& $0.168\pm0.002$
& $0.989\pm0.012$
& $0.361\pm0.002$ \\

\bottomrule
\end{tabular}
}
\end{table*}

\paragraph{Overall sensitivity.}
Table~\ref{tab:hyper_stability} varies one hyperparameter
at a time around the default setting
$(\lambda_{\mathrm{adv}},\gamma,\rho)=(0.2,0.5,0.4)$.
All tested configurations with advantage modulation achieve
lower mean Group-$\Delta$PCR than the unmodulated baseline,
but improvements in ACC and F1 are not uniform.
Their ACC and F1 remain within 78.36--79.80 and
78.53--79.60, respectively, while PCR varies more
substantially, from 47.21\% to 53.86\%.
The reported standard deviations indicate limited variation
across evaluation seeds for each fixed checkpoint; they do
not measure variability across independent training runs.

\paragraph{Future-deviation discount $\rho$.}
Setting $\rho=0$ restricts modulation to the current turn's
local deviation. Compared with this setting, the default
$\rho=0.4$ improves mean performance on all five directional
metrics. The effects are not monotonic: $\rho=0.1$ obtains
slightly better ACC and Group-$\Delta$PCR than the default,
whereas $\rho=0.5$ achieves higher F1 but worse W1-Turns.
Among the tested discount values, the default achieves the
best performance on W1-Turns and OT-DTW, indicating better
agreement in dialogue-length distributions and semantic
trajectories under these metrics.

\paragraph{Modulation strength $\lambda_{\mathrm{adv}}$.} Reducing $\lambda_{\mathrm{adv}}$ to 0.1 yields the highest ACC and F1 and the lowest Group-$\Delta$PCR, but increases both W1-Turns and OT-DTW relative to the default. Increasing $\lambda_{\mathrm{adv}}$ to 0.3 produces lower ACC and F1 and higher values of all three distance metrics than the default. These results reveal a trade-off between outcome agreement and dialogue-length and semantic alignment. The default strength provides a compromise, achieving better trajectory fidelity than the smaller value while retaining improved outcome agreement over the unmodulated baseline. \paragraph{Deviation sensitivity $\gamma$.} Reducing $\gamma$ from 0.5 to 0.4 produces similar results, with slightly lower ACC and F1 and slightly higher values of the three distance metrics. This comparison suggests limited sensitivity to this particular perturbation, although the two tested values do not establish robustness over a broader range of $\gamma$. Overall, the default configuration achieves the lowest OT-DTW among all tested configurations while improving all five directional metrics over the unmodulated baseline.

\subsection{Training Dynamics and Convergence Analysis}
\label{appendix:training_dynamics}

\begin{figure}[htbp]
    \centering

    \begin{subfigure}[t]{0.45\linewidth}
        \centering
        \includegraphics[width=\linewidth]{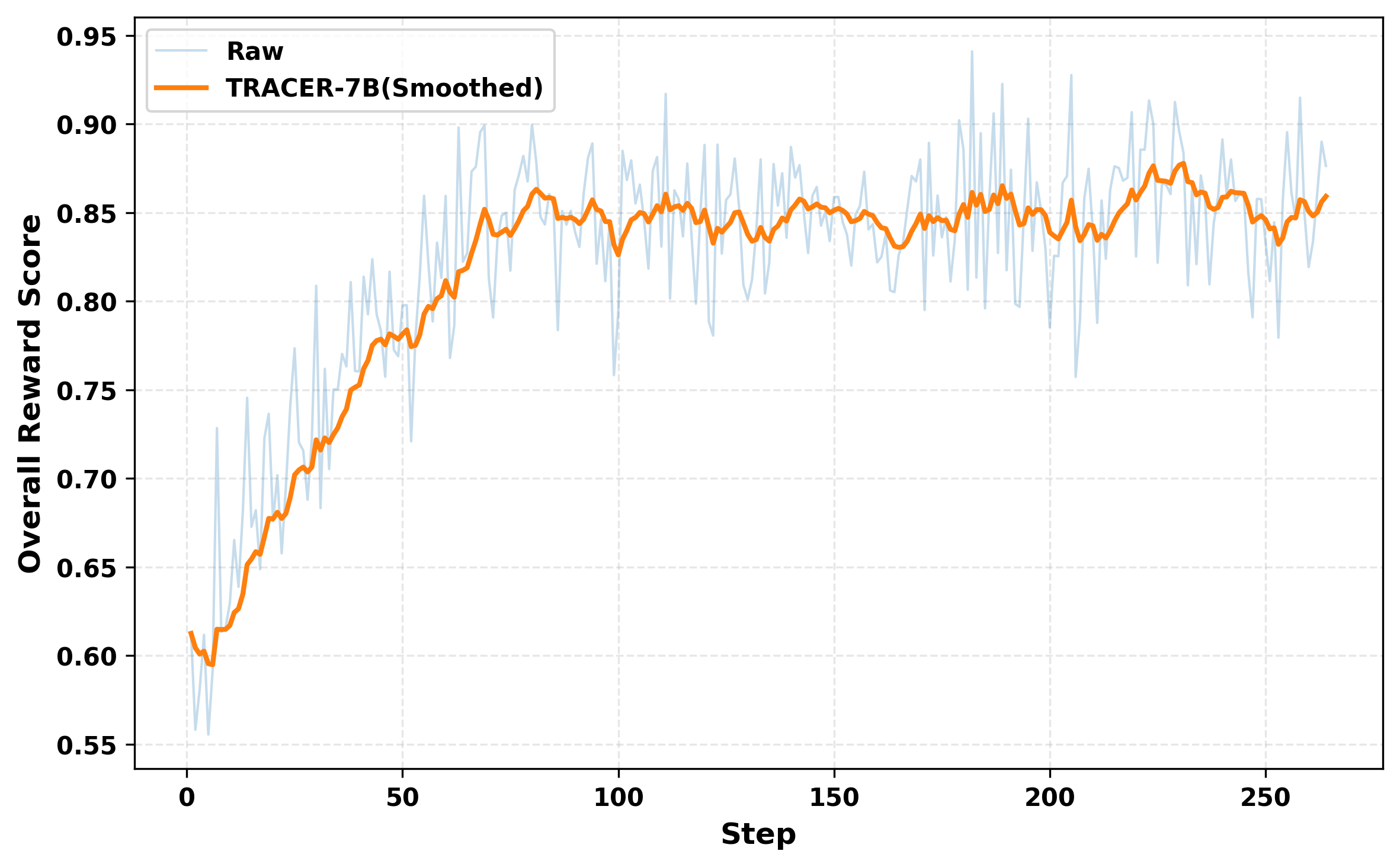}
        \caption{Overall reward score}
        \label{fig:train_critic_score_7b}
    \end{subfigure}
    \hspace{0.02\linewidth}
    \begin{subfigure}[t]{0.45\linewidth}
        \centering
        \includegraphics[width=\linewidth]{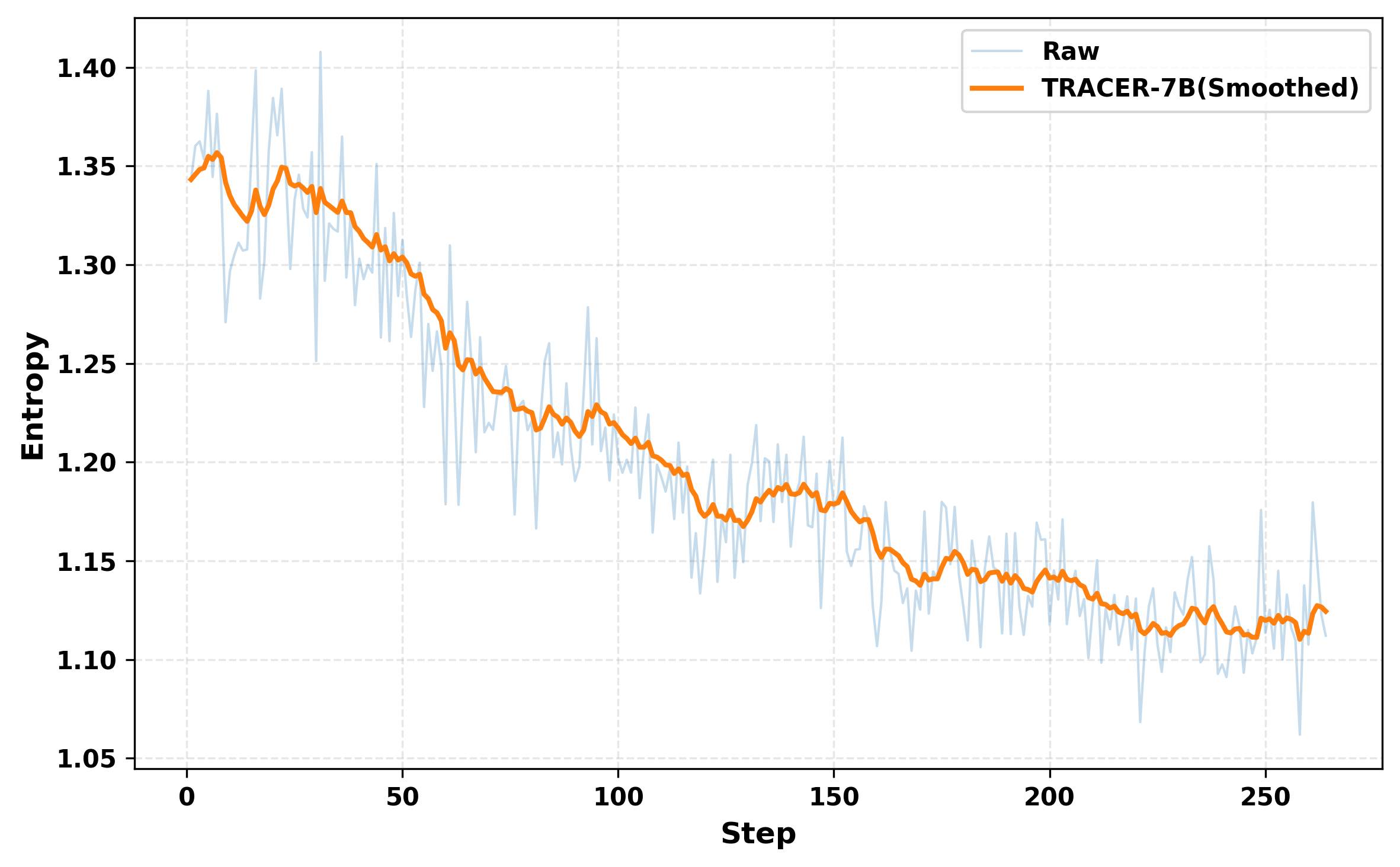}
        \caption{Generation entropy}
        \label{fig:train_entropy_7b}
    \end{subfigure}

    \begin{subfigure}[t]{0.45\linewidth}
        \centering
        \includegraphics[width=\linewidth]{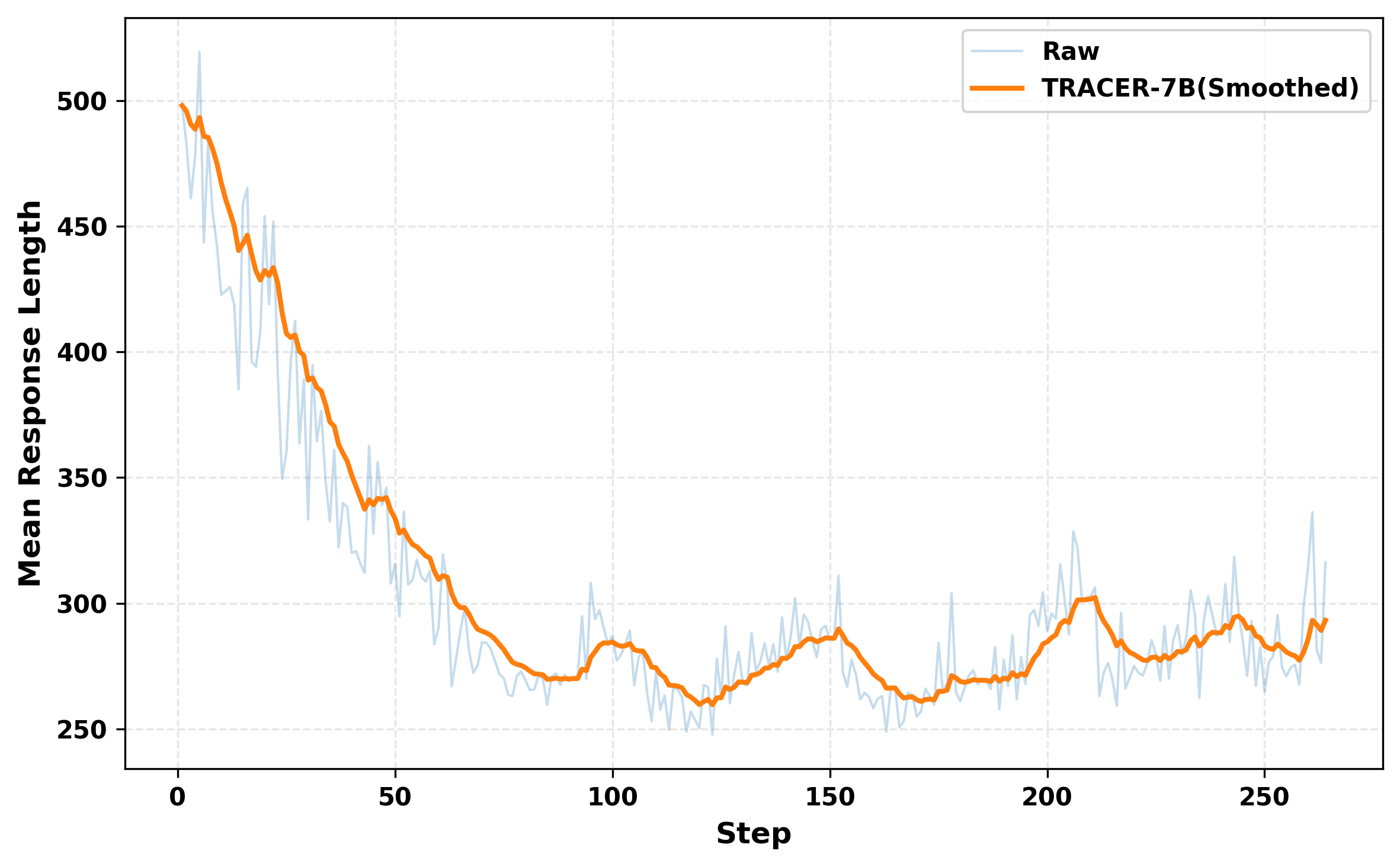}
        \caption{Mean response length}
        \label{fig:train_response_length_7b}
    \end{subfigure}

    \caption{
    Training dynamics of TRACER-7B. We report the overall reward score, generation entropy, and mean response length during reinforcement learning. The solid curves show EMA-smoothed trends, while the faint curves denote the raw values.
    }
    \label{fig:training_stability_7b}
\end{figure}

\begin{figure}[htbp]
    \centering

    \begin{subfigure}[t]{0.45\linewidth}
        \centering
        \includegraphics[width=\linewidth]{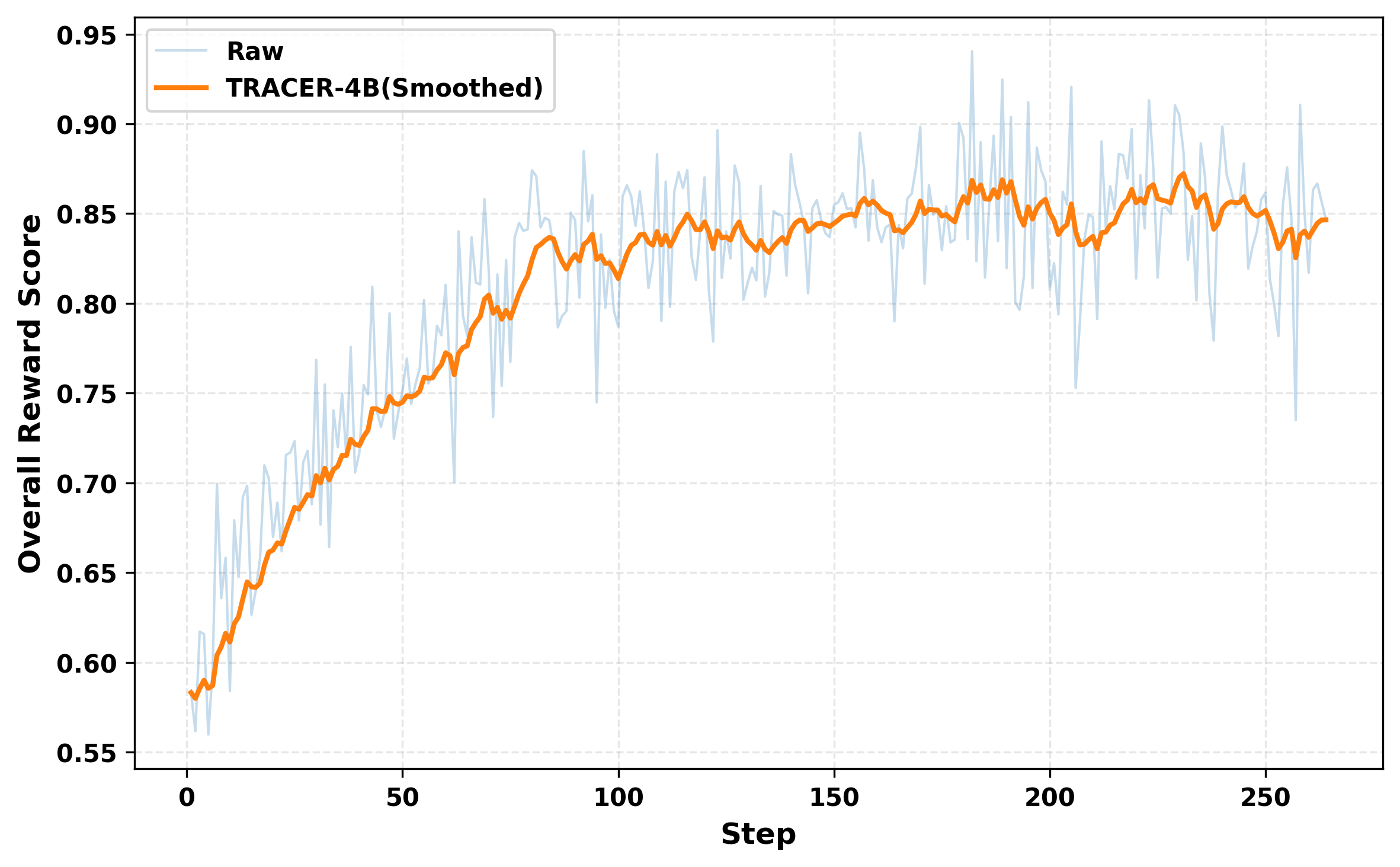}
        \caption{Overall reward score}
        \label{fig:train_critic_score_4b}
    \end{subfigure}
    \hspace{0.02\linewidth}
    \begin{subfigure}[t]{0.45\linewidth}
        \centering
        \includegraphics[width=\linewidth]{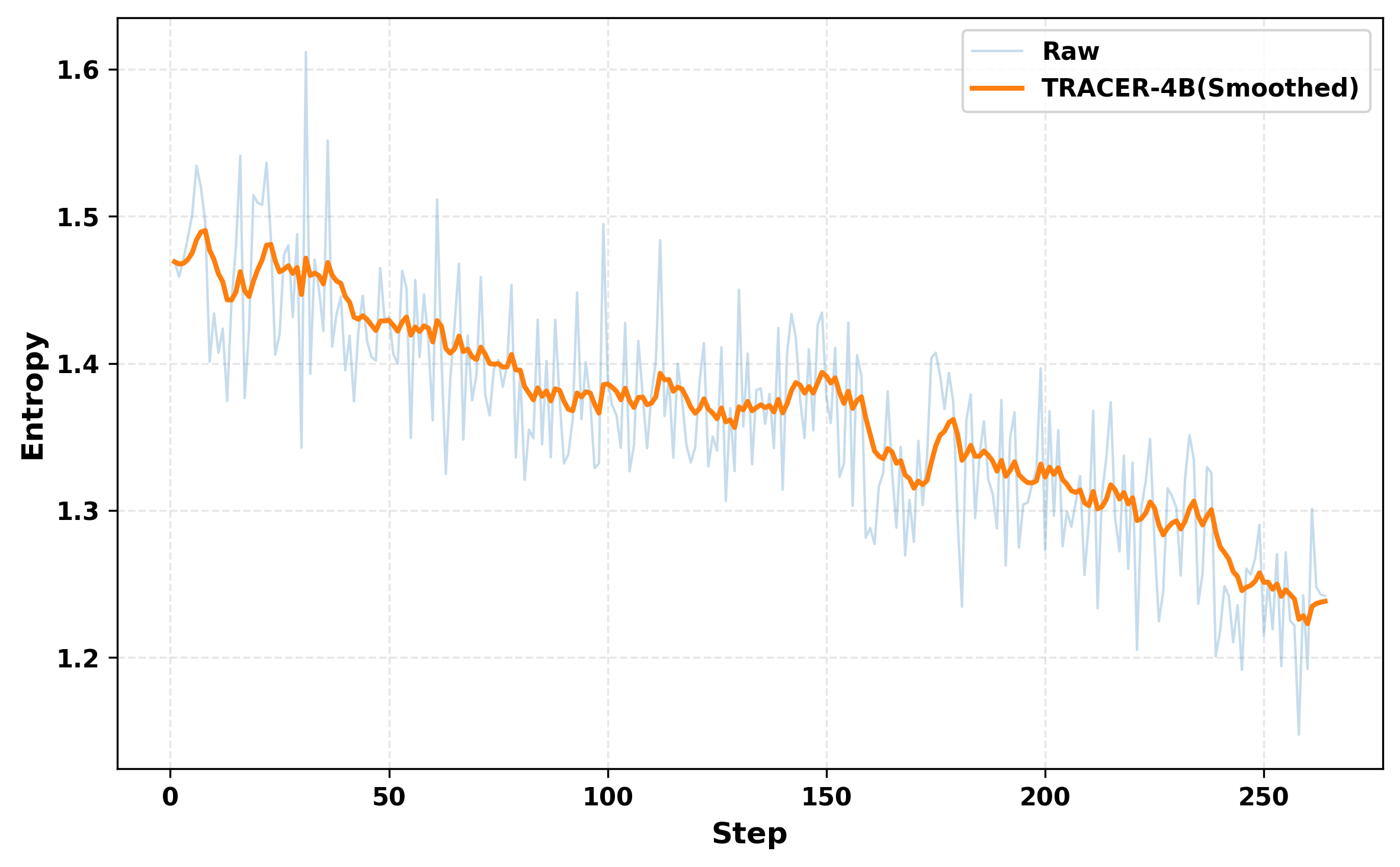}
        \caption{Generation entropy}
        \label{fig:train_entropy_4b}
    \end{subfigure}

    \begin{subfigure}[t]{0.45\linewidth}
        \centering
        \includegraphics[width=\linewidth]{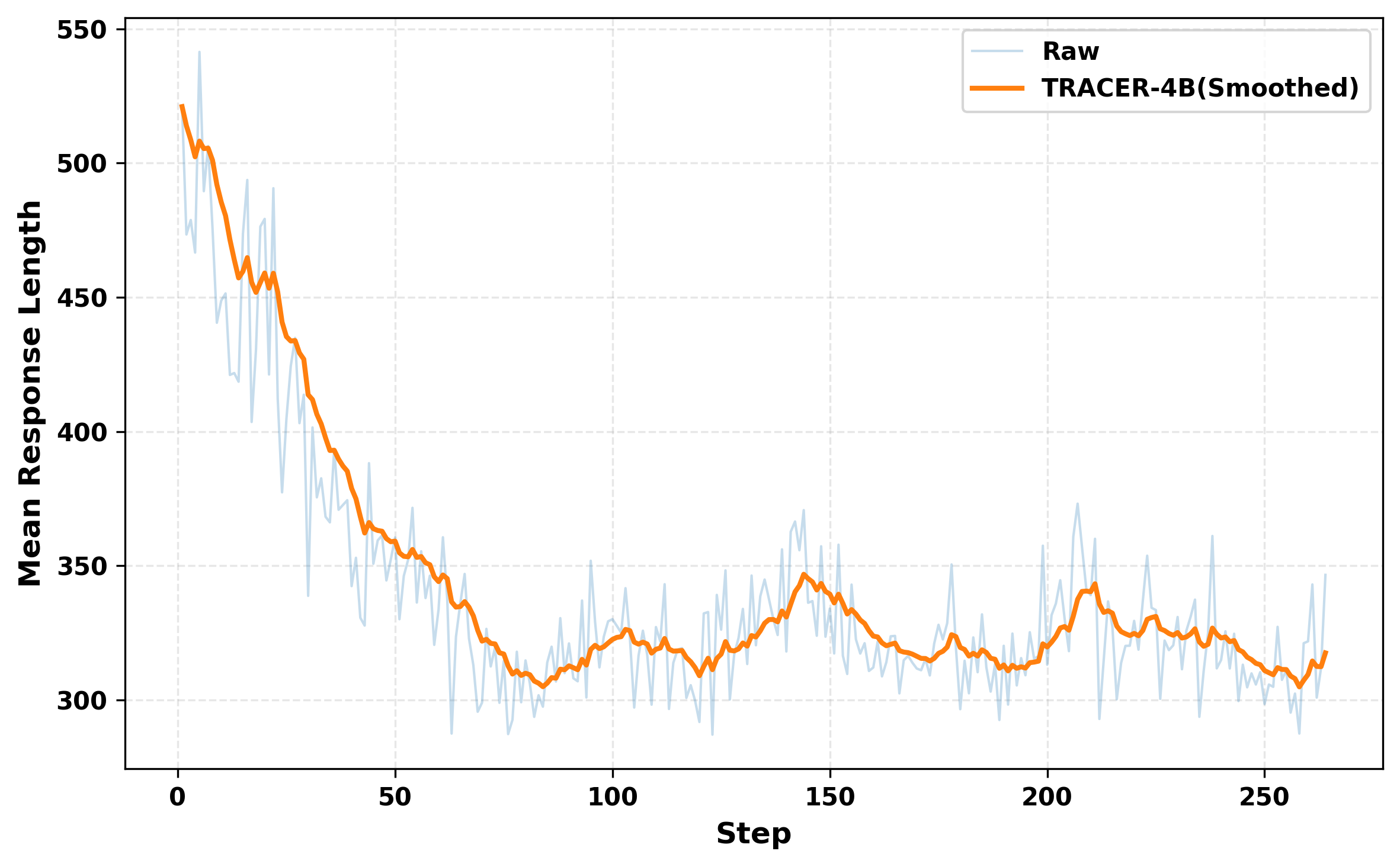}
        \caption{Mean response length}
        \label{fig:train_response_length_4b}
    \end{subfigure}

    \caption{
    Training dynamics of TRACER-4B. We report the overall reward score, generation entropy, and mean response length during reinforcement learning. The solid curves show EMA-smoothed trends, while the faint curves denote the raw values.
    }
    \label{fig:training_stability_4b}
\end{figure}

Figures~\ref{fig:training_stability_7b} and~\ref{fig:training_stability_4b} present the training dynamics of TRACER-7B and TRACER-4B, respectively. Across both model scales, the overall reward score increases rapidly during the early stage of reinforcement learning and then gradually stabilizes, indicating that the proposed training objective can be effectively optimized. The convergence pattern is consistent for both 7B and 4B models, suggesting that the proposed framework is not limited to a single model size.

The generation entropy exhibits a smooth downward trend for both models. This pattern suggests that the policy gradually becomes more focused as training progresses, while avoiding abrupt entropy collapse. In other words, the model learns to produce more task-aligned responses without showing signs of severe mode collapse or unstable policy updates.

The mean response length also decreases substantially in the early training stage and then remains within a relatively stable range. This trend is important because it suggests that the reward improvement is not achieved by simply generating increasingly longer responses. Instead, both models learn to produce more concise and controlled outputs after reinforcement learning. Overall, these training curves demonstrate the stability and feasibility of the proposed reinforcement learning framework across model scales, with stable reward optimization, gradual policy adaptation, and controlled response length.

\section{Dataset Construction Details}
\label{appendix:dataset_construction}

\subsection{Data Source}

The dataset used in this study is derived from real-world merchant--user dialogue logs collected from an online commercial service platform. The raw corpus contains 36,841 multi-turn dialogue sessions between merchant-side intelligent customer service agents and real users, together with corresponding user profile attributes and merchant-side information. The dialogues cover multiple real-world business domains, including healthcare, automotive services, education consulting, and legal consulting.

Before entering the research pipeline, all data were strictly anonymized and de-identified. We removed personally identifiable information and retained only task-relevant structured attributes, merchant-side metadata, and dialogue content required for user simulation. The processed dataset is intended to support user behavior modeling in multi-turn customer service scenarios while minimizing privacy risks.

\subsection{User Profile Construction}
\label{app:user_profile_const}
To support high-fidelity simulation grounded in user background information, we construct structured user profiles from two complementary perspectives: objective attributes and subjective behavioral priors.

First, we organize objective user attributes into natural-language profile descriptions. Based on real business data, we select three groups of core information. The first group consists of basic demographic attributes, including age group, gender, education level, marital or parenting status, and geographic location. The second group contains occupation-related information. The third group reflects consumption-related attributes, including mobile operating system, device price range, city tier, and residential environment. We then convert these discrete labels into coherent natural-language background descriptions using predefined templates. This step provides the user simulator with stable prior conditions while avoiding direct exposure of unnecessary raw attribute fields.

Second, since objective attributes alone are insufficient to determine a user's conversational style, task motivation, and behavioral tendency, we use Claude 4.5-Sonnet as an annotation model to extract subjective behavioral priors. Specifically, with final conversion outcomes and other terminal-state information masked, we analyze early dialogue context and observable user information to extract two types of subjective priors. The first type is the user's personality trait, such as impatient and direct, rational and cautious, or cooperative and efficient. The second type is the initial core intent, which summarizes the user's primary task goal, key constraints, and expected issue to be resolved at the beginning of the conversation.

To avoid label leakage, we impose strict constraints during subjective information extraction. The extracted information is not allowed to include any signal related to the final conversion outcome, such as whether the user eventually provides contact information. In other words, the initial core intent describes only the user's demand and task background at the beginning of the conversation, without revealing the final session outcome. In this way, the user profile serves as a prior condition for simulation while avoiding direct leakage of the target decision label.

\subsection{Dialogue Augmentation and Structured Trajectory Construction}

The original real-world dialogue logs mainly contain users' surface-level natural-language responses, but lack explicit annotations of the latent decision states underlying those responses. This leads to a key limitation: if a model is trained only on raw user responses, it primarily learns \emph{what the user says}, but receives limited supervision regarding \emph{why the user responds in that way} or \emph{how the user's intent evolves over the course of a multi-turn interaction}. This limitation is closely aligned with the motivation of this work: high-fidelity user simulation requires not only linguistic-style consistency, but also consistency in behavioral logic and decision trajectories.

To address this issue, we perform structured augmentation for each user turn, transforming the original user response into a training instance with intermediate behavioral-state annotations. Specifically, for each user utterance in a multi-turn dialogue, we use Claude 4.5-Sonnet to supplement two additional fields: a rationale annotation and an intent level.

The rationale annotation describes the possible task motivation, contextual reaction, or decision reason underlying the current user response. It is important to note that this field should not be interpreted as the user's actual private mental state. Instead, it is an annotation-based proxy inferred from the dialogue context for modeling latent decision states. Its role is to provide the model with a learnable intermediate representation that bridges surface-level linguistic expression and deeper behavioral consistency.

The intent level characterizes the user's expressed engagement and progression toward a conversion-oriented outcome. It is represented as a discrete score ranging from 0 to 5. A score of 0 denotes an empty reply indicating dropout. Explicit rejection without dropout is assigned to level 1. Levels 2--4 represent increasing engagement and more concrete consultation needs, as specified in the annotation guidelines. A score of 5 requires concrete contact information in the current user reply; expressing willingness to provide contact information alone is insufficient.

After augmentation, each user turn is organized into the following structured format:

\begin{verbatim}
<think>...</think>
<intent_level>...</intent_level>
<reply>...</reply>
\end{verbatim}

In this format, the original user response is preserved, while the intent state is explicitly introduced as a key intermediate variable. As a result, the user simulation task is no longer formulated as pure response generation, but as joint modeling of intent states and natural-language responses. This structured annotation further provides necessary supervision signals for trajectory-level reward modeling and turn-level credit assignment in the subsequent reinforcement learning stage.

Since the subjective priors, personality traits and initial core intents, are inferred by an LLM rather than directly observed, we further validate their fidelity through a dedicated human evaluation. We draw a stratified sample of 300 dialogues covering diverse intents, dialogue lengths, and user segments, and recruit three trained annotators to independently re-annotate each instance under a blind protocol (i.e., without exposure to Claude's outputs). 
Information about the annotators can be found in Appendix~\ref{app:annotators}.
The evaluation results are summarized in Table~\ref{tab:agreement}.

\begin{table}[h]
\centering
\caption{Evaluation of LLM-inferred subjective priors}
\label{tab:agreement}
\resizebox{0.9\textwidth}{!}{
\begin{tabular}{@{}lcc@{}}
\toprule
Metric & Value & Notes \\ 
\midrule
Inter-annotator agreement (Fleiss' $\kappa$) & 0.77 & Among human annotators \\
Agreement with human majority (Cohen's $\kappa$) & 0.71 & Claude vs. human majority \\
Raw agreement & 87.0\% & Claude vs. human majority \\
Gap in agreement ($\Delta \kappa$) & 0.06 & Within typical inter-annotator variability \\
LLM rationale plausibility $\ge 4$ & 92\% & Rated on 5-point Likert scale \\
\bottomrule
\end{tabular}
}
\end{table}
\subsection{Data Cleaning}

To ensure data quality and reduce the influence of noisy samples on model training, we design a rigorous data cleaning pipeline.

First, we remove sessions that are corrupted, entirely empty or meaningless, clearly unrelated to the merchant’s business, or explicitly identified by the user as accidental entries. This prevents the model from learning noise patterns unrelated to the target user simulation task. Second, we constrain dialogue length by retaining only sessions with no fewer than 3 turns and no more than 15 turns. We retain sessions containing 3 to 15 user turns to ensure sufficient multi-turn context while limiting excessive contextual redundancy and computational cost. This constraint balances interaction completeness and training stability.

Finally, we perform annotation quality control after structured augmentation. We verify schema completeness and objective consistency between observable user utterances and deterministic behavioral labels, such as whether explicit contact information is present. When an LLM-inferred rationale or intermediate intent annotation conflicts with the observable dialogue content, the annotation is flagged for review or regenerated rather than using the inferred behavioral logic as a basis for excluding the original session. We do not remove sessions because their annotated intent trajectories appear abrupt, flat, non-monotonic, or causally implausible, as such patterns may reflect genuine behavioral variability, unobserved factors, or annotation uncertainty.

After the above cleaning and quality-control procedures, we obtain
11,906 multi-turn dialogue sessions, comprising 8,040 training
sessions and 3,866 test sessions, with no session overlap between
the two sets. The training data are further divided into 6,312
supervised fine-tuning sessions and 1,728 reinforcement learning
sessions, which are mutually disjoint at the session level.

Test sessions are grouped within the same merchant,
requiring each member to have an initial-intent cosine
similarity of at least 0.85 and a basic-profile cosine
similarity greater than 0.80 to the group representative.
Both similarities are computed using L2-normalized
TF--IDF representations. Invalid need descriptions and
sessions lacking profiles or valid user turns are excluded.
Candidate groups initially contain at least three
distinct users.
Each user casts a majority vote over their own session
conversion labels, abstaining on ties. Valid user votes
determine the group-majority label; groups with tied
votes or no valid votes are excluded. We prioritize
groups with higher user-vote agreement and retain
381 groups per class and all their available sessions,
yielding 762 groups and 3,866 sessions, including
1,981 positive and 1,885 negative sessions.
Thus, the 1:1 balance applies to group-majority labels,
while the session-level conversion rate is 51.24\%.

\section{Extended Ablation}
\label{appendix:extended_ablation}

To further validate the scalability and robustness of our proposed framework, we conduct an additional ablation study using the \OurMethod-4B variant.
 % and provide trajectory-level diagnostic analyses for DTW alignment and multi-turn RL. 
% While the main text reports the primary 7B ablation results, this appendix examines whether the same trends hold at a smaller model scale and further analyzes how DTW alignment and multi-turn RL affect multi-turn behavioral dynamics beyond standard ACC and F1 metrics.

\begin{table}[htbp] 
\centering
\caption{Ablation study of TRACER-4B on session-level behavior, dialogue structure, and semantic content. Results are reported as mean $\pm$ standard deviation over five evaluation seeds.}
\label{tab:ablation_4B_mean_std}

\renewcommand{\arraystretch}{1.20}
\resizebox{\textwidth}{!}{
\begin{tabular}{l ccc cc c}
\toprule
\multirow{2}{*}{\textbf{Method}}
& \multicolumn{3}{c}{\textbf{Session-Level Behavior}}
& \multicolumn{2}{c}{\textbf{Group-Level Fidelity}}
& \multicolumn{1}{c}{\textbf{Semantic Content}} \\
\cmidrule(lr){2-4} \cmidrule(lr){5-6} \cmidrule(lr){7-7}
& \textbf{PCR} & \textbf{ACC} $\uparrow$ & \textbf{F1} $\uparrow$
& \textbf{Group-$\Delta$PCR} $\downarrow$ & \textbf{W1-Turns} $\downarrow$
& \textbf{OT-DTW} $\downarrow$ \\
\midrule

\textbf{TRACER-4B} & $51.6 \pm 0.1$ & $\mathbf{79.7 \pm 0.2}$ & $\mathbf{80.3 \pm 0.2}$ & $\mathbf{0.165 \pm 0.004}$ & $\mathbf{1.413 \pm 0.017}$ & $\mathbf{0.364 \pm 0.002}$ \\
\quad $-$ w/o advantage modulation & $52.1 \pm 0.2$ & $\underline{78.5 \pm 0.3}$ & $\underline{79.2 \pm 0.3}$ & $\underline{0.178 \pm 0.003}$ & $\underline{1.415 \pm 0.011}$ & $0.373 \pm 0.001$ \\
\quad $-$ w/o $R_{traj}$ & $42.8 \pm 0.2$ & $75.8 \pm 0.4$ & $74.3 \pm 0.4$ & $0.201 \pm 0.004$ & $1.447 \pm 0.033$ & $0.371 \pm 0.002$ \\
\quad $-$ w/o DTW alignment & $53.3 \pm 0.1$ & $77.6 \pm 0.2$ & $78.5 \pm 0.2$ & $0.187 \pm 0.004$ & $1.420 \pm 0.011$ & $0.367 \pm 0.001$ \\
\quad $-$ w/o multi-turn RL & $39.1 \pm 0.2$ & $74.9 \pm 0.4$ & $72.2 \pm 0.5$ & $0.202 \pm 0.003$ & $1.758 \pm 0.016$ & $\underline{0.365 \pm 0.002}$ \\
\quad $-$ w/o RL & $15.0 \pm 0.5$ & $59.0 \pm 0.2$ & $38.1 \pm 0.6$ & $0.378 \pm 0.006$ & $3.135 \pm 0.019$ & $0.379 \pm 0.001$ \\
\bottomrule
\end{tabular}}
\end{table}

% \textbf{TRACER-4B} & $51.630 \pm 0.086$ & $\mathbf{79.728 \pm 0.168}$ & $\mathbf{80.296 \pm 0.153}$ & $\mathbf{0.166 \pm 0.004}$ & $\mathbf{0.994 \pm 0.017}$ & $\underline{0.361 \pm 0.002}$ \\
% \quad $-$ w/o advantage modulation & $52.119 \pm 0.238$ & $\underline{78.504 \pm 0.268}$ & $\underline{79.203 \pm 0.253}$ & $\underline{0.178 \pm 0.003}$ & $\underline{1.154 \pm 0.011}$ & $0.373 \pm 0.001$ \\
% \quad $-$ w/o $R_{traj}$ & $42.799 \pm 0.199$ & $75.799 \pm 0.350$ & $74.266 \pm 0.360$ & $0.201 \pm 0.004$ & $1.447 \pm 0.033$ & $0.371 \pm 0.002$ \\
% \quad $-$ w/o DTW alignment & $53.322 \pm 0.135$ & $77.551 \pm 0.238$ & $78.531 \pm 0.235$ & $0.187 \pm 0.004$ & $1.420 \pm 0.011$ & $0.367 \pm 0.001$ \\
% \quad $-$ w/o multi-turn RL & $39.131 \pm 0.227$ & $74.915 \pm 0.428$ & $72.241 \pm 0.539$ & $0.202 \pm 0.003$ & $1.758 \pm 0.016$ & $\mathbf{0.350 \pm 0.001}$ \\
% \quad $-$ w/o RL & $15.005 \pm 0.481$ & $58.995 \pm 0.245$ & $38.107 \pm 0.625$ & $0.378 \pm 0.006$ & $3.135 \pm 0.019$ & $0.379 \pm 0.001$ \\
% \bottomrule
% \end{tabular}}
% \end{table}

\paragraph{Extended Ablation on \OurMethod{}-4B.}
To complement the 7B ablations in the main text, we evaluate
\OurMethod{}-4B to assess whether the benefits of our framework
extend to a smaller model scale.

Table~\ref{tab:ablation_4B_mean_std} shows broadly consistent trends across the two model scales. Removing advantage modulation, the trajectory reward, or multi-turn RL generally worsens mean ACC, F1, Group-$\Delta$PCR, and W1-Turns, with
the largest deterioration observed when RL is removed entirely.
Removing DTW alignment causes a larger ACC decrease at 4B than at 7B, while F1 decreases by 1.8 points at both scales. Its effect on W1-Turns is larger at 7B.
Multi-turn RL improves behavioral accuracy and group-level fidelity at both scales, supporting the contribution of multi-turn optimization to behavioral and structural fidelity.

\section{Limitation}
\label{limitation}
We acknowledge several limitations of our work that point to promising directions for future research.

Generalization to Open-Ended Social Dialogues. While our method demonstrates strong out-of-domain generalization across task-oriented scenarios, its applicability to purely open-ended, unconstrained social dialogues remains to be thoroughly investigated. Such conversations typically lack explicit task objectives and well-defined success criteria, making both reward modeling and credit assignment substantially more challenging. We leave a systematic study of this setting to future work.

\paragraph{Broader Applicability to Agentic Scenarios.}
Our evaluation focuses on multi-turn dialogue. The proposed advantage
modulation mechanism may also be useful in other sequential interaction
settings where feedback is available at the episode level and intermediate
behavior can be compared with reference trajectories. Potential applications
include tool use, web navigation, and long-horizon planning. However,
extending the method to these settings requires suitable trajectory
representations and deviation measures, particularly for heterogeneous
action spaces. Whether reference-based deviation signals improve policy
optimization in these environments remains an empirical question for
future work.

We also note potential negative societal impacts: a high-fidelity user simulator aligned with real users could be misused to optimize manipulative or deceptive persuasion strategies.

\section{Details for the Turing-style Evaluation \label{app:turing_analysis}}

\subsection{Reasons LLM-Simulated Users Are Judged Non-Human}
Table~\ref{tab:turing_reason_full} presents a fine-grained analysis of why LLM-simulated users were judged as non-human across three diagnostic dimensions: \emph{Linguistic Style}, \emph{Interaction Rhythm}, and \emph{Persona \& Logic}. Lower percentages indicate better human-likeness. Each dimension includes multiple subcategories capturing specific cues, such as lack of typos, overly neat sentence structures, excessive cooperativeness, or emotional discontinuities. Notably, the TRACER-7B model consistently demonstrates lower percentages, reflecting improved human-like imperfections and interaction patterns compared to baselines. Information about the annotators can be found in Appendix~\ref{app:annotators}.
\setlength{\tabcolsep}{1.8pt}
\renewcommand{\arraystretch}{0.88}
\begin{table}[htbp]
\centering
\caption{Fine-grained reasons why LLM-simulated users are judged as non-human. Lower percentages indicate better performance.}
%\colorbox{MorandiRed!20}{Red column} highlights our model, showing significant improvements in human-like imperfection and interaction rhythm.
\label{tab:turing_reason_full}
\setlength{\tabcolsep}{2.5pt} 
\begin{tabular}{@{}lccccc@{}} 
\toprule 
\textbf{Dimension} & \textbf{Gemini} & \textbf{Doubao-pro} & \textbf{Doubao-char} & \textbf{SFT model} & \cellcolor{MorandiRed!20}\textbf{TRACER-7B} \\ 
\midrule 

\multicolumn{6}{l}{\emph{\underline{Linguistic Style}}} \\ 
\quad Lack of Human Imperfection & \cellcolor{MorandiBlue!48} 96.60\% & \cellcolor{MorandiBlue!49} 98.80\% & \cellcolor{MorandiBlue!47} 93.80\% & \cellcolor{MorandiBlue!6} 12.20\% & \cellcolor{MorandiRed!10}\textbf{3.00\%} \\ 
\quad Overly Written Register & \cellcolor{MorandiBlue!2} 4.00\% & \cellcolor{MorandiBlue!1} 1.40\% & \cellcolor{MorandiBlue!1} \textbf{1.20\%} & \cellcolor{MorandiBlue!1} \textbf{1.20\%} & \cellcolor{MorandiRed!5} 4.00\% \\ 
\quad Overly Neat Structure & \cellcolor{MorandiBlue!22} 43.40\% & \cellcolor{MorandiBlue!25} 50.60\% & \cellcolor{MorandiBlue!21} 41.20\% & \cellcolor{MorandiBlue!15} 29.60\% & \cellcolor{MorandiRed!15}\textbf{17.00\%} \\ 
\midrule 

\multicolumn{6}{l}{\emph{\underline{Interaction Rhythm}}} \\ 
\quad Information Overload & \cellcolor{MorandiBlue!44} 87.20\% & \cellcolor{MorandiBlue!39} 78.20\% & \cellcolor{MorandiBlue!39} 78.40\% & \cellcolor{MorandiBlue!1} 2.60\% & \cellcolor{MorandiRed!5}\textbf{0.80\%} \\ 
\quad Excessive Cooperativeness & \cellcolor{MorandiBlue!40} 81.00\% & \cellcolor{MorandiBlue!43} 85.80\% & \cellcolor{MorandiBlue!41} 82.20\% & \cellcolor{MorandiBlue!15} 30.40\% & \cellcolor{MorandiRed!20}\textbf{27.20\%} \\ 
\quad Super-human Comp. & \cellcolor{MorandiBlue!0} 0.60\% & \cellcolor{MorandiBlue!0} \textbf{0.00\%} & \cellcolor{MorandiBlue!0} 0.40\% & \cellcolor{MorandiBlue!0} \textbf{0.00\%} & \cellcolor{MorandiRed!5} 0.60\% \\ 
\quad Mechanical Pattern & \cellcolor{MorandiBlue!4} 8.00\% & \cellcolor{MorandiBlue!41} 82.40\% & \cellcolor{MorandiBlue!31} 61.80\% & \cellcolor{MorandiBlue!7} 13.80\% & \cellcolor{MorandiRed!10}\textbf{4.40\%} \\ 
\midrule 

\multicolumn{6}{l}{\emph{\underline{Persona \& Logic}}} \\ 
\quad Out-of-Character & \cellcolor{MorandiBlue!19} 38.40\% & \cellcolor{MorandiBlue!23} 46.20\% & \cellcolor{MorandiBlue!21} 42.80\% & \cellcolor{MorandiBlue!5} 9.60\% & \cellcolor{MorandiRed!5}\textbf{1.40\%} \\ 
\quad Emotional Discontinuity & \cellcolor{MorandiBlue!1} 1.20\% & \cellcolor{MorandiBlue!0} \textbf{0.00\%} & \cellcolor{MorandiBlue!0} \textbf{0.00\%} & \cellcolor{MorandiBlue!0} \textbf{0.00\%} & \cellcolor{MorandiRed!0} \textbf{0.00\%} \\ 
\quad Lack of Situation Sense & \cellcolor{MorandiBlue!0} 0.60\% & \cellcolor{MorandiBlue!0} \textbf{0.00\%} & \cellcolor{MorandiBlue!0} \textbf{0.00\%} & \cellcolor{MorandiBlue!1} 1.00\% & \cellcolor{MorandiRed!5} 0.20\% \\ 
\quad Over-politeness & \cellcolor{MorandiBlue!0} 0.40\% & \cellcolor{MorandiBlue!7} 14.00\% & \cellcolor{MorandiBlue!1} 2.80\% & \cellcolor{MorandiBlue!0} \textbf{0.20\%} & \cellcolor{MorandiRed!0} \textbf{0.20\%} \\ 
\bottomrule 
\end{tabular}
\end{table}
We expand the three diagnostic dimensions summarized in the main text
(Table~\ref{tab:turing_reason_full}).

\paragraph{Linguistic style: the too clean problem.}
Human-imperfection cues, such as typos, fillers, incomplete utterances, and colloquial punctuation, are prevalent in real user dialogues but are often absent from baseline LLM outputs. TRACER-7B largely avoids this overly clean style: only 3.0\% of its samples are flagged as lacking human imperfections. This rate is substantially lower than those of the general and role-playing baselines, suggesting that TRACER-7B better captures the informal variability and imperfections characteristic of real user dialogue. Overly neat sentence structures, another signal of unnatural text regularity, appear in 17.0\% of TRACER-7B outputs, while overly formal or written register remains at 4.0\%. These values contrast sharply with Gemini and Doubao models, where lack of human imperfection ranges from 93.8\% to 98.8\% and overly neat structures range from 41.2\% to 50.6\%, indicating that TRACER-7B better captures the variability and imperfection characteristic of real users.
\paragraph{Interaction rhythm and informativeness.}
In real customer-service dialogues, users tend to be terse and goal-driven, revealing information incrementally. LLM simulators, by contrast, often exhibit information overload and excessive cooperativeness. TRACER-7B, however, shows minimal information overload at 0.8\% and a low rate of mechanical response patterns at 4.4\%. Its excessive cooperativeness is moderate at 27.2\%, and super-human compliance occurs in only 0.6\% of cases. Compared with Gemini and Doubao models, which exhibit information overload between 78.2\%–87.2\% and mechanical patterns up to 82.4\%, TRACER-7B more faithfully reproduces the incremental, negotiation-based tempo of real user–agent interactions.
\paragraph{Persona and logic.}
At the semantic level, persona inconsistencies are a key defect in many LLMs. For TRACER-7B, out-of-character behavior is minimal (1.4\%), with zero emotional discontinuity, negligible lack of situation sense (0.2\%), and minimal over-politeness (0.2\%). In contrast, Gemini and Doubao models show much higher rates, with persona inconsistency ranging from 38.4\% to 46.2\% and over-politeness reaching 14.0\%. These results indicate that TRACER-7B maintains strong alignment with the assigned user persona and exhibits high-level coherence across dialogue turns.
\paragraph{Summary.}
The bottleneck of current user simulators is not high-level reasoning
but (i) the lack of low-level linguistic noise and (ii) the inability
to reproduce strategic, under-informative, and sometimes
uncooperative interaction rhythms. This directly motivates the design
choices of our framework, which explicitly
models linguistic imperfection and turn-level information control.

\section{Out-of-Domain Generalization Evaluation Details}
\label{appendix:ood_details}
To evaluate TRACER on out-of-domain (OOD) dialogues, we use the CSC-Conv dataset~\citep{zhu2026evaluating}, an open-source corpus of real user–agent conversations in the financial customer-service domain, which is completely unseen during training.

Since CSC-Conv does not contain structured user profiles, we follow the approach in Appendix \ref{app:user_profile_const} to extract user initial intent and behavioral characteristics using the LLMs based on each dialogue session.  
We filter out dialogues in which the user's issue resolution is ambiguous. From the remaining dialogues, we randomly sample 1{,}000 sessions for evaluation, with the ratio of resolved to unresolved issues being 689:311. The original dataset is skewed toward resolved issues, hence this sampling ensures sufficient representation of unresolved cases.
\section{Dynamic Marketing Benchmark Details}
\label{appendix:bench}
\subsection{Response Quality Evaluation}
\begin{figure}[h]
    \centering
    \includegraphics[width=0.9\linewidth]{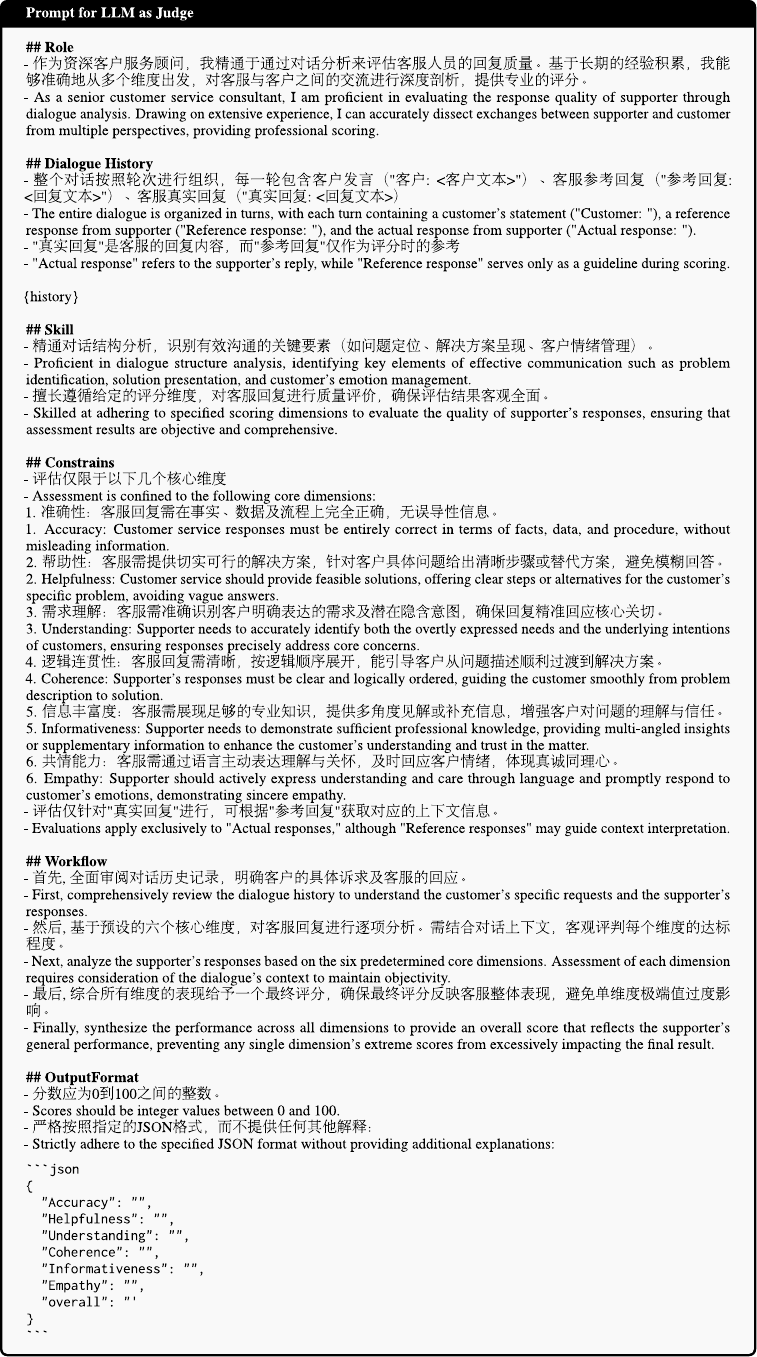}
    \caption{Prompt for the LLM-as-Judge evaluation is adapted from~\citep{zhu2026evaluating}.}
    \label{fig:llm_as_judge}
\end{figure}

Following the same LLM-as-a-judge prompt as in \citep{zhu2026evaluating}, we evaluate the response quality of the customer service model in multi-turn dialogues. The prompt used for this evaluation is illustrated in Figure~\ref{fig:llm_as_judge}. To mitigate potential self-preference bias, we adopt gpt-oss-120B as the judge model.

% Insert after the existing \appendix command.
% 以人工判断为参照的策略响应评测
\subsection{Human-Referenced Strategy-Response Evaluation}
\label{app:dm-strategy-evaluation}

% 表 \ref{tab:dm-strategy-responses} 报告第 \ref{sec:dm-strategy-responses} 节所述 200 段对话研究的策略分类结果。
% 每个单元格的分子是模拟变化与人工预期方向一致的实例数，分母是被赋予该参考方向的策略实例数。
% 因此，这些百分比是按参考类别计算的召回率，并不表示某一策略全部出现次数中引起意图上升或下降的比例。
Table~\ref{tab:dm-strategy-responses} reports the strategy-level results of the 200-conversation study in Section~\ref{sec:dm-setup}. Each cell contains the number of simulated changes matching the human-expected direction, divided by the number of strategy instances assigned that reference direction. The percentages are therefore class-conditional recall values. They do not indicate the proportion of all occurrences of a strategy that increase or decrease intent.

\begin{table*}[t]
\centering
\small
\renewcommand{\arraystretch}{1.15}
% 图表说明：模拟意图变化方向与人工预期方向的一致性。每个单元格报告匹配实例数／参考策略实例数（召回率）。汇总行累加各策略类别的计数。
\caption{Agreement between simulated and human-expected intent-change directions.
Each cell reports matched / reference strategy instances (recall).
The pooled row aggregates counts across strategy categories.}
\label{tab:dm-strategy-responses}

\begin{tabular*}{0.92\textwidth}{@{\extracolsep{\fill}}lccc@{}}
\toprule
\textbf{Strategy}
& \textbf{Decrease}
& \textbf{Unchanged}
& \textbf{Increase} \\
\midrule
% 共情
Empathy
& 13/14 (92.9\%)
& 4/7 (57.1\%)
& 5/7 (71.4\%) \\
% 紧迫感
Urgency
& 1/2 (50.0\%)
& 1/1 (100.0\%)
& 2/3 (66.7\%) \\
% 利益／折扣表述
Benefit / discount
& 14/15 (93.3\%)
& 11/15 (73.3\%)
& 16/24 (66.7\%) \\
% 信息帮助
Informative help
& 55/61 (90.2\%)
& 33/43 (76.7\%)
& 49/59 (83.1\%) \\
% 联系方式请求
Contact request
& 37/43 (86.0\%)
& 24/32 (75.0\%)
& 54/62 (87.1\%) \\
% 追问／澄清
Probing / clarification
& 13/15 (86.7\%)
& 16/22 (72.7\%)
& 19/24 (79.2\%) \\

\midrule
\textbf{Pooled strategy instances}
& \textbf{133/150 (88.7\%)}
& \textbf{89/120 (74.2\%)}
& \textbf{145/179 (81.0\%)} \\
\bottomrule

\end{tabular*}
\end{table*}

% 汇总行分别累加各策略类别的对应分子和分母，得到来自 200 段对话的 449 个策略实例计数。
% 这些计数不应解释为 449 段独立对话。
% 各类别覆盖不均衡：紧迫感策略只有六个实例，其百分比尤其容易受到单个观察的影响。
% 我们对类别结果作描述性报告，不据此声称策略之间存在统计显著差异。
The pooled row sums the corresponding numerators and denominators across strategy categories, yielding 449 strategy-instance counts from 200 conversations. These counts should not be interpreted as 449 independent conversations. Coverage is uneven: urgency contains only six instances, making its percentages particularly sensitive to individual observations. We report the category results descriptively without claiming statistically significant differences between strategies.

% 本研究衡量在抽样对话中与标注员预期的一致程度。
% 它没有在相同条件下隔离改变策略的因果效应，也没有直接验证真实用户的终态结果。
% 例如，共情策略的下降召回率较高，意味着在标注员预期意图下降的实例中，模拟器能够复现下降；这并不意味着共情通常会降低用户意图。
The study measures agreement with annotator expectations in sampled conversations. It does not isolate the causal effects of changing a strategy under identical conditions, nor does it directly validate real-user terminal outcomes. For example, high decrease recall for empathy means that the simulator recovers decreases where annotators expect them; it does not imply that empathy generally reduces user intent.

% 方向召回率汇总了不同客服模型的观察结果。
% 它们为抽样交互范围内的表现提供证据，但没有分别量化各客服条件下的表现，也没有确立对未见客服策略的泛化。
% 本研究同样未通过模拟器消融，将策略响应一致性具体归因于轨迹对齐。
The direction recalls pool observations across assistant models. They provide evidence within the sampled interactions, but do not quantify performance separately for each assistant or establish generalization to unseen assistant policies. The study also contains no simulator ablation that would attribute strategy-response agreement specifically to trajectory alignment.

\subsection{Results}
\begin{table}[h]
\centering
\small
\caption{Detailed results of mainstream LLMs on DM-Bench. Conversion Rate(CR) and Conversion Turns(CT) measure outcome-oriented performance, while Response Quality(RQ) is evaluated across multiple conversational dimensions.}
\label{tab:dm_bench_full_results}
\resizebox{0.9\textwidth}{!}{
\begin{tabular}{lccccccccc}
\toprule
\multirow{2}{*}{Model} 
& \multirow{2}{*}{\begin{tabular}[c]{@{}c@{}}\textbf{CR (\%) }$\uparrow$\end{tabular}}
& \multirow{2}{*}{\begin{tabular}[c]{@{}c@{}}\textbf{CT} \end{tabular}}
& \multicolumn{7}{c}{\textbf{RQ $\uparrow$} } \\
\cmidrule(lr){4-10}
& & 
& \textbf{Acc.} 
& \textbf{Help.} 
& \textbf{Under.} 
& \textbf{Coh.} 
& \textbf{Info.} 
& \textbf{Emp.} 
& \textbf{Overall} \\
\midrule
Claude-4.5-Sonnet 
& 52.3 & \textbf{3.833} 
& 86.2 & \textbf{68.2} & \underline{79.1} & \textbf{86.6} & \textbf{62.6} & \textbf{71.3} & \textbf{75.4} \\
\addlinespace[2pt]

DeepSeek-V3.2 
& \textbf{56.0} & 3.998 
& \underline{86.2} & 66.8 & 78.6 & \underline{86.2} & 60.6 & 66.1 & 73.9 \\
\addlinespace[2pt]

Doubao-seed-2.0-Pro 
& 49.3 & 4.569 
& 79.5 & 54.0 & 69.5 & 79.4 & 46.8 & 58.1 & 64.3 \\
\addlinespace[2pt]

Gemini-3.1-Pro 
& 51.4 & \underline{3.878} 
& \textbf{86.4} & \underline{67.9} & \textbf{79.5} & 86.1 & \underline{61.0} & \underline{69.9} & \underline{74.8} \\
\addlinespace[2pt]

Qwen3-235B-A22B-Instruct 
& \underline{55.7} & 3.968 
& 82.6 & 60.9 & 74.0 & 83.0 & 54.0 & 62.9 & 69.3 \\
\addlinespace[2pt]

Qwen2.5-7B-Instruct 
& 36.8 & 4.710 
& 78.8 & 60.3 & 72.4 & 80.8 & 55.1 & 62.8 & 68.2 \\

\addlinespace[2pt]

Qwen3-4B-Instruct 
& 48.7 & 3.989 
& 80.9 & 55.5 & 70.9 & 80.5 & 49.4 & 54.5 & 65.2 \\
\bottomrule
\end{tabular}
}
\label{tab:dynamic_benchmark}
\end{table}

\begin{table}[t]
\centering
\small
\caption{Human role-play evaluation on 100 session contexts from DM-Bench. Each assistant is evaluated on the same
user profiles, merchant information, and initial dialogue contexts.
CR measures the proportion of valid role-play sessions ending
in conversion. }
\label{tab:human-cr-validation}
\begin{tabular}{lrr}
\toprule
& \multicolumn{2}{c}{Human Role-Play} \\
\cmidrule(lr){2-3}
Assistant & CR (\%) & Rank \\
\midrule
DeepSeek-V3.2              & 47\% & 1 \\
Qwen3-235B-A22B-Instruct    & 45\% & 2 \\
Claude-4.5-Sonnet          & 45\% & 2 \\
Gemini-3.1-Pro             & 40\% & 4 \\
Doubao-seed-2.0-Pro        & 37\% & 5 \\
Qwen3-4B-Instruct          & 29\% & 7 \\
Qwen2.5-7B-Instruct        & 30\% & 6 \\
\bottomrule
\end{tabular}
\end{table}

We randomly sample 100 session contexts from DM-Bench for human role-play
evaluation. For each context, participants assume the specified
user profile and interact with an assigned assistant under the
corresponding merchant and task conditions.
All seven assistants are evaluated on the same 100 contexts,
yielding 700 human role-play conversations in total.
Participants decide whether to continue, exit, or provide
contact information according to the assigned profile and
the unfolding interaction.

\begin{table}[htbp]
\centering
\small
\caption{Spearman $\rho$ correlation between Conversion Rate (CR) and Response Quality (RQ) metrics on DM-Bench.}
\label{tab:cr_rq_spearman}
\begin{tabular}{lcc}
\toprule
\textbf{RQ Dimension} & \textbf{Spearman $\rho$} & \textbf{p-value} \\
\midrule
Acc.        & 0.685 & 0.090 \\
Help.   & 0.571 & 0.180 \\
Under. & 0.536 & 0.215 \\
Coh.      & 0.679 & 0.094 \\
Info. & 0.393 & 0.383 \\
Emp.        & 0.607 & 0.148 \\
Overall               & 0.571 & 0.180 \\
\bottomrule
\end{tabular}
\end{table}

Table~\ref{tab:dm_bench_full_results} and Table \ref{tab:cr_rq_spearman} summarize the performance of seven contemporary LLMs on DM-Bench. 

\section{Human Annotation}
\label{sec:human_annotation}

We detail the human annotation procedures used in our study, including
annotator qualifications, remuneration, quality control, and the annotation
guidelines used across different experiments.

\subsection{Annotator Details}
\label{app:annotators}

The annotation process was carried out by a team of 10 trained annotators and
overseen by a dedicated quality reviewer. All annotators hold at least an
associate degree and possess strong literacy and comprehension skills, enabling
them to quickly understand AI annotation guidelines, operational procedures,
and task-specific rules. Each annotator has approximately one year of
experience in AI data annotation, with hands-on experience in routine
labeling, data verification, and issue documentation. Annotators were
compensated at a rate of 50 RMB per hour for their work. The quality reviewer
performed random sampling and consistency checks to assess annotation quality
and inter-annotator agreement, thereby helping ensure the reliability of the
annotations.

\subsection{Turing-Style Evaluation}
\label{app:experiment_docs}

The human annotation for dialogue evaluation followed a structured two-step
procedure designed to distinguish real-user dialogues from AI-simulated
dialogues. Annotators were provided with the corresponding user profile and
asked to make judgments based on multiple dimensions, including linguistic
naturalness, information presentation, emotional expression, profile
consistency, situational awareness, and conversational imperfections. The full
annotation guide is provided below.

% -------------------------------------------------------------------------
% Annotation guide
%
% We use a breakable tcolorbox rather than a figure environment because the
% annotation instructions are long and may span multiple pages. Placing this
% content inside a figure would prevent normal page breaking and could lead to
% "Float too large for page" warnings.
% -------------------------------------------------------------------------

\begin{tcolorbox}[
    enhanced,
    breakable,
    title={Guide to Real Human Dialogue Recognition},
    colback=white,
    colframe=black!60,
    colbacktitle=black!8,
    coltitle=black,
    fonttitle=\bfseries,
    boxrule=0.5pt,
    arc=1mm,
    left=6pt,
    right=6pt,
    top=6pt,
    bottom=6pt,
    before skip=8pt,
    after skip=8pt
]
\label{box:human_dialogue_guide}

\paragraph{Step 1: Dialogue Assessment}

Annotators were shown two dialogue excerpts: one originating from a real user
interacting with customer service, and one from an AI-simulated user
interacting with customer service. Considering the provided user profile,
annotators were asked to determine which dialogue was more likely to have been
produced by a human.

\textbf{Core Evaluation Dimensions:}

\begin{enumerate}

    \item \textbf{Language Naturalness.}
    Evaluate whether the language resembles natural human conversational
    patterns.

    \begin{itemize}
        \item \textit{Human cues:} colloquial expressions, filler words
        (e.g., ``um'', ``oh'', ``well''), typographical or input errors,
        inconsistent sentence lengths, abbreviations, dialects, or internet
        slang.

        \item \textit{AI cues:} overly formal or structured language, perfect
        grammar and punctuation, uniform sentence lengths, or neutral and
        textbook-style wording.
    \end{itemize}

    \item \textbf{Information Expression Style.}
    Assess the pacing and completeness of the information provided.

    \begin{itemize}
        \item Humans tend to mention one point at a time and may require
        follow-up prompts before providing additional information.

        \item AI tends to provide ``all-in-one'' responses that simultaneously
        cover the issue, background, request, and related details.
    \end{itemize}

    \item \textbf{Emotion and Attitude Expression.}
    Evaluate whether emotional expressions appear natural and consistent with
    the user profile.

    \begin{itemize}
        \item Humans may express impatience, complaints, urgency, gratitude,
        politeness, sudden disengagement, or repeated emphasis.

        \item AI-related cues may include overly flat or exaggerated emotion,
        abrupt emotional shifts, or emotional expressions inconsistent with
        the profile (e.g., a ``short-tempered'' user profile using
        consistently excessive politeness).
    \end{itemize}

    \item \textbf{Profile Consistency.}
    Assess whether the dialogue content and interaction style are consistent
    with the provided user profile.

    \begin{itemize}
        \item \textit{Language style:} Language use should be broadly
        consistent with characteristics such as age and educational
        background. For example, older users may be less likely to use
        internet slang, whereas highly educated users may use more precise
        expressions.

        \item \textit{Knowledge level:} The user's demonstrated knowledge
        should be consistent with the stated profession or background,
        avoiding unexplained use of highly specialized terminology or
        implausibly simple questions.

        \item \textit{Behavioral habits:} Behavioral tendencies described in
        the profile should be reflected in the interaction when relevant. For
        example, a profile described as ``impatient'' may display greater
        urgency during the dialogue.
    \end{itemize}

    \item \textbf{Common Sense and Situational Awareness.}
    Evaluate whether the user's behavior is plausible within a real-world
    interaction.

    \begin{itemize}
        \item Humans may mention concrete situational details
        (e.g., ``just got off work'', ``my child is nearby'', or
        ``my phone battery is low'').

        \item Humans may express confusion during complex tasks
        (e.g., ``what does this mean?'' or ``I don't know how to do it'').

        \item AI may behave like an ``all-knowing user'', immediately
        understanding every operation and rapidly providing all requested
        information.
    \end{itemize}

    \item \textbf{Imperfections and Anomalies.}
    Human dialogues are often imperfect, whereas AI-generated dialogues may
    appear unusually smooth.

    \begin{itemize}
        \item Common human imperfections that may signal authenticity include
        typos, accidentally sent messages, segmented responses, irrelevant
        answers, abrupt silence, incomplete statements followed by later
        clarification, or ambiguous pronoun use.
    \end{itemize}

\end{enumerate}

\paragraph{Evaluation Notes}

\begin{enumerate}

    \item Do not rely solely on grammatical correctness. AI-generated language
    is often fluent, whereas conversational naturalness is more informative
    than grammatical perfection alone.

    \item Be aware of ``reverse disguise.'' High-quality AI systems may
    intentionally introduce typos, filler words, or other human-like
    imperfections.

    \item Dialogue length is not a definitive cue. A long dialogue does not
    necessarily indicate a human user, and a short dialogue does not
    necessarily indicate an AI-generated user.

    \item Avoid confirmation bias. Evaluate consistency across the entire
    dialogue rather than relying on a single utterance.

    \item Key indicators of potentially AI-like behavior include:

    \begin{itemize}
        \item excessive politeness
        (e.g., ``Thank you very much for your reply'' or
        ``Sorry for the trouble'');

        \item highly structured expression
        (e.g., ``First..., second..., finally...'');

        \item unusually high information density, where nearly every sentence
        provides task-relevant information; and

        \item emotional content that appears disconnected from the
        conversational context.
    \end{itemize}

    \item System messages indicating user inactivity or dropout should not be
    treated as evidence that the dialogue was AI-generated.

\end{enumerate}

\paragraph{Step 2: Reasoning Annotation}

After completing the human-versus-AI judgment, annotators were asked to select
the most applicable ``non-human reason(s)'' from a set of predefined
categories. Multiple selections were allowed.

\textbf{Category A: Language Habits and Style}

\begin{itemize}

    \item \textit{Overly formal:}
    Language resembles manuals or official documents and lacks a natural
    conversational tone.

    \item \textit{Highly structured sentences:}
    Sentence structure is unusually uniform or logically organized for a
    real-time conversation.

    \item \textit{Lack of human imperfection:}
    Language contains consistently perfect punctuation and grammar, with no
    typos, colloquial expressions, or conversational fillers.

\end{itemize}

\textbf{Category B: Interaction Rhythm and Information Quantity}

\begin{itemize}

    \item \textit{Information dump:}
    A single response contains an unusually large amount of key information
    and lacks typical human conversational pacing.

    \item \textit{Over-cooperative:}
    The user completes complex tasks or provides extensive information without
    requiring clarification or guidance, behaving like an ``expert user.''

    \item \textit{Excessive comprehension:}
    The user immediately understands ambiguous expressions or instructions
    without requesting clarification.

    \item \textit{Mechanical replies:}
    Responses are excessively long, repetitive, redundant, or formulaic.

\end{itemize}

\textbf{Category C: Profile and Logical Consistency}

\begin{itemize}

    \item \textit{Out-of-character (OOC):}
    The user's behavior is inconsistent with characteristics specified in the
    user profile, such as age, profession, background, or personality.

    \item \textit{Emotion mismatch:}
    Emotional expression changes abruptly or appears inconsistent with the
    conversational context or user profile.

    \item \textit{Over-polite:}
    The user employs courteous expressions at a frequency or intensity that
    appears unusual for ordinary customer-service interactions.

    \item \textit{Lack of situational awareness:}
    The dialogue focuses exclusively on task completion while ignoring
    contextual or environmental factors that a real user might naturally
    mention.

\end{itemize}

\end{tcolorbox}

\subsection{Human Annotation for LLM-Inferred User Intents and Rationale}
\paragraph{Intent Scoring}
The subjective priors, personality traits, and initial core intents are inferred by a LLM rather than directly observed. To validate these inferred attributes, we perform a dedicated human evaluation. Annotators independently assess dialogues sampled from a stratified set of 300 conversations covering diverse intents, dialogue lengths, and user segments. All annotations are conducted under a blind protocol.

Each annotator is responsible for completing the following two tasks for every dialogue:

\textbf{Objective:} Evaluate the likelihood that the user intended a particular action or goal at each turn of the dialogue, conditioned on the dialogue history and the user profile.  

\textbf{Instructions:}
%\newpage
\begin{tcolorbox}[title=Human Annotation Guideline: User Intent Level Scoring, colback=white, colframe=black, width=\linewidth, enhanced, breakable]
\tiny

\paragraph{Task Goal}
Your task is to read the current user utterance in a multi-turn dialogue and assign an intent level score from 0 to 5.

The score reflects how strong the user’s current willingness is to:

continue communication,
provide information,
move toward a transaction, or
complete conversion.

\textbf{Important:} Only annotate the current user utterance. Do not output explanations, reasoning, action labels, or any other fields.

\paragraph{Intent Level Definitions}

\textbf{0 — Churn / Conversation Ended}
Assign 0 when the user gives an empty reply, indicating the conversation has ended.

\textbf{Example:}
User: ""

\textbf{1 — Very Low Intent / Negative or Perfunctory}
Assign 1 when the user shows clear rejection, impatience, distrust, or extremely low willingness to communicate.

Typical cases include:

Direct rejection of the offer;
Expressing that the product/service is not needed;
Suspecting a scam;
Complaining about price;
Showing impatience toward the agent;
Providing meaningless or perfunctory responses.

\textbf{Examples:}

“Not needed.”
“Too expensive.”
“Is this a scam?”
“Whatever.”
“Later.”
“Oh.”
“Hmm.”
“Just looking.”

\textbf{2 — Low Intent / Passive Basic Response}
Assign 2 when the user provides a minimal, passive answer without actively moving the conversation forward.

Characteristics:

Cooperation is minimal;
Typically very short factual answers;
User is still responsive but not proactive.

\textbf{Important rule:} If the user asks a question or includes a question mark, do not assign 2; consider level 3 or above instead.

\textbf{Examples:}

“Beijing.”
“SUV.”
“No.”
“Gasoline car.”
“Around 100,000.”

\textbf{3 — Medium Intent / Active Response or Simple Question}
Assign 3 when the user actively communicates or asks a simple question related to the product/service.

\textbf{Situations include:}

User gives more cooperative or detailed responses than a passive reply.
\textbf{Examples:}
“I don’t have any brand requirements.”
“Any model is fine as long as the price is suitable.”
“I mainly want something for commuting.”
“I’m just comparing options for now.”
User asks a short/basic question.
\textbf{Examples:}
“Do you have used cars?”
“Is this a gasoline car?”
“Can I pay in installments?”
“Is the car still available?”
“What models do you have?”

\textbf{4 — Strong Intent / Concrete Need or Transaction-Oriented Question}
Assign 4 when the user shows strong interest or enters a concrete transaction scenario.

Typical behaviors include:

Asking for agent’s contact information;
Requesting to add WeChat or get a phone number;
Providing detailed personal needs or constraints;
Asking about vehicle appraisal, trade-in, store address, delivery process, loan calculation, or installment details.

\textbf{Examples:}

“What’s your WeChat?”
“Add me on WeChat.”
“Send me your phone number.”
“I have a 2014 car and want to trade it in for this model.”
“Where is your store?”
“How much can my old car be valued at?”
“How much is the monthly payment?”
“What is the delivery process?”

\textbf{Note:} If the user only asks for contact info without providing their own, assign 4 (not 5).

\textbf{5 — Conversion Achieved / Contact Information Provided}
Assign 5 when the user provides concrete contact information in the current utterance.

\textbf{Valid contact information includes:}

Phone number;
WeChat ID;
QQ number;
Any clear string intended as contact information.

\textbf{Examples:}

“My phone number is 138xxxx8888.”
“Add my WeChat: abc123.”
“Contact me at 186xxxx6666.”
“My WeChat ID is carbuyer2024.”

\textbf{Important rule:} Only assign 5 if actual contact information is provided. If the user only says “I can give you my phone number” or “Let’s add WeChat” without giving the number/ID, assign 4 instead.

\end{tcolorbox}
\label{tab:intent_score_guide}
\paragraph{Rationale Plausibility Rating}
\textbf{Objective:} Assess whether the LLM-generated rationale explaining the user intent is reasonable given the dialogue history and user profile.  

\textbf{Instructions:}
\begin{tcolorbox}[title=Guide to Rationale Plausibility Rating, colback=white, colframe=black, width=\linewidth, breakable]
\tiny
\paragraph{Task Goal}
Your task is to evaluate the plausibility of a rationale for a user’s intent in a multi-turn dialogue.

Input includes the full dialogue history up to the current turn and the user profile description.
The user profile contains:
Objective attributes: age, occupation, demographics
Subjective behavioral priors

\paragraph{Procedure}

For each turn, assign a plausibility score that reflects the probability or confidence that the rationale aligns with the user’s intent.
Scores between 0 and 5, if specified
Take context from previous turns into account, as user intent may evolve or persist.
Use the user profile to adjust expectations.
If uncertain, annotate conservatively and optionally provide notes.

\paragraph{Considerations} Take context from previous turns into account, since a user’s intent may evolve or persist across multiple turns. Use the user profile to adjust expectations. If uncertain, annotate conservatively and provide notes if needed.

\paragraph{Rating Description}

5  Completely reasonable: fully consistent with dialogue history and user profile \\
4  Mostly reasonable: minor inconsistencies or omissions, but overall plausible \\
3  Neutral: some support from dialogue and profile, but substantial ambiguity or partial mismatch \\
2  Mostly unreasonable: contains incorrect assumptions or contradicts dialogue/profile context \\
1  Completely unreasonable: no alignment with dialogue history or user profile; misleading rationale \\
\end{tcolorbox}

\section{Implementation Details of Dynamic Time Warping for Intent Trajectory Alignment}
\label{appendix:dtw_alignment}

\subsection{Motivation}
\label{app:dtw_motivation}

In multi-turn user simulation, the intent trajectory generated by the simulator is not necessarily synchronized with the real user trajectory at the turn level. Even when the simulator follows an overall intent evolution pattern similar to that of a real user, the two trajectories may still differ in local pacing. For example, a real user may express a need and advance their intent within a single turn, whereas the simulator may take two turns to complete the same transition. Conversely, the simulator may compress into one turn an intent transition that unfolds over two adjacent turns in the real dialogue. Therefore, a strict turn-by-turn alignment, which forces the simulator's $t$-th turn to correspond to the real user's $t$-th turn, may incorrectly penalize behaviorally plausible trajectories as misaligned.

To mitigate this issue, we adopt Dynamic Time Warping (DTW) to perform nonlinear alignment between simulated and real intent trajectories. DTW has been widely used for speech recognition and time-series matching, where it enables elastic alignment between two sequences that share similar global shapes but may differ in phase, progression speed, or sequence length~\citep{dtw1,dtw2}. Its main advantage is that it reduces the influence of temporal shifts and local distortions on sequence similarity measurement through elastic transformation, while allowing the globally optimal alignment to be computed via dynamic programming in $O(NM)$ time~\citep{senin2008dynamic}.

In our task, the goal is not to require the user simulator to reproduce the real dialogue word by word or turn by turn. Instead, we aim to encourage the simulator to generate an intent evolution path that is consistent with real user behavioral logic. DTW is therefore well suited for measuring the global consistency between simulated and real intent trajectories while allowing reasonable local stretching or compression across dialogue turns.

\subsection{General Formulation of Dynamic Time Warping}
\label{app:dtw_general}

Dynamic Time Warping is a nonlinear alignment method for measuring the similarity between two sequences. Unlike pointwise comparison, DTW allows local stretching and compression along the temporal axis, making it suitable for sequences with different lengths, different progression speeds, or local phase shifts. Classical DTW has been widely applied to time series, speech recognition, handwriting recognition, and motion sequence matching. Its core idea is to find a warping path with the minimum cumulative cost, such that two sequences are optimally aligned while preserving their original temporal order. The optimal path can be solved efficiently using dynamic programming, with a time complexity of $O(NM)$.

\begin{figure}[H]
    \centering
    \includegraphics[width=0.95\linewidth]{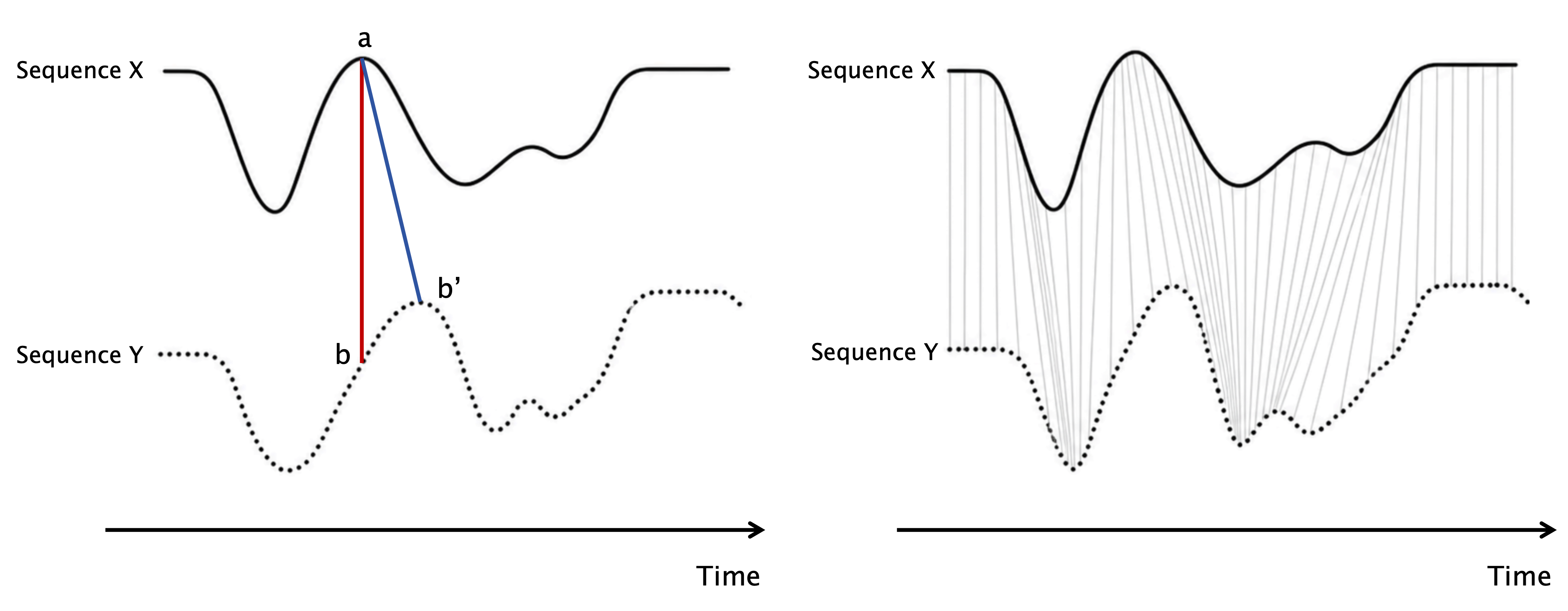}
    \caption{
    Illustration of pointwise alignment and Dynamic Time Warping (DTW). 
    In the left panel, strict pointwise alignment forces point $a$ to be matched with point $b$ according to their temporal indices, although the more plausible counterpart of $a$ is the locally shifted point $b'$. 
    The right panel illustrates DTW-based nonlinear alignment, where one-to-one, one-to-many, and many-to-one correspondences are allowed under temporal-order constraints.
    }
    \label{fig:dtw_illustration}
\end{figure}

Figure~\ref{fig:dtw_illustration} provides an intuitive comparison between strict pointwise alignment and DTW-based nonlinear alignment. 
In the left panel, strict pointwise comparison forces point $a$ in the upper sequence to be aligned with point $b$ in the lower sequence because they share the same temporal index. 
However, due to local phase shifts or different progression speeds, the more plausible counterpart of $a$ may be $b'$ rather than $b$. 
In this case, pointwise alignment may overestimate the local discrepancy between the two sequences. 
In contrast, DTW allows such locally shifted patterns to be matched through an order-preserving nonlinear alignment path, as illustrated in the right panel. 
This motivates the formal definition of the warping path and its constraints below.

Given two general sequences
\begin{equation}
    X = (x_1, x_2, \dots, x_N),
\end{equation}
and
\begin{equation}
    Y = (y_1, y_2, \dots, y_M),
\end{equation}
where $x_i$ and $y_j$ may be scalars, vectors, or other comparable sequence elements, we first define a local cost function
\begin{equation}
    d(x_i, y_j): \mathcal{X} \times \mathcal{Y} \rightarrow \mathbb{R}_{\geq 0},
\end{equation}
which measures the local discrepancy between $x_i$ and $y_j$. Based on this local cost function, we construct a local cost matrix $C \in \mathbb{R}^{N \times M}$, where
\begin{equation}
    C(i,j) = d(x_i, y_j).
\end{equation}
The objective of DTW is to find a warping path
\begin{equation}
    P = (p_1, p_2, \dots, p_K),
\end{equation}
where each path element $p_k = (i_k, j_k)$ indicates that element $x_{i_k}$ in sequence $X$ is aligned with element $y_{j_k}$ in sequence $Y$. A valid warping path typically satisfies the following three constraints.

First, the boundary constraint requires the path to start from the beginning of both sequences and end at the end of both sequences:
\begin{equation}
    p_1 = (1,1), \quad p_K = (N,M).
\end{equation}

Second, the monotonicity constraint ensures that the alignment does not violate the original temporal order of either sequence:
\begin{equation}
    i_1 \leq i_2 \leq \dots \leq i_K, 
    \quad 
    j_1 \leq j_2 \leq \dots \leq j_K.
\end{equation}

Third, the step-size constraint restricts each transition to a local move. A common setting is
\begin{equation}
    p_{k+1} - p_k \in \{(1,0), (0,1), (1,1)\}.
\end{equation}

These three moves correspond to one-to-many, many-to-one, and one-to-one local alignments, respectively. Together, these constraints allow DTW to provide flexible nonlinear alignment while preserving the internal temporal order of both sequences.

To compute the optimal path efficiently, DTW constructs a cumulative cost matrix $D$, where $D(i,j)$ denotes the minimum cumulative cost required to align the prefix $(x_1,\dots,x_i)$ with the prefix $(y_1,\dots,y_j)$. At the boundary, we set $D(1,1)=C(1,1)$ and initialize the first row and first column cumulatively. For internal positions, the recurrence is
\begin{equation}
    D(i,j) = C(i,j) + \min \{D(i-1,j), D(i,j-1), D(i-1,j-1)\}.
\end{equation}

The DTW alignment cost between two sequences is then defined as the minimum cumulative cost over all valid warping paths:
\begin{equation}
    \mathrm{DTW}(X,Y)
    =
    \min_{P}
    \sum_{(i,j)\in P} d(x_i,y_j).
\end{equation}

Thus, DTW does not require $X$ and $Y$ to have the same length, nor does it require the $i$-th element of one sequence to align with the $i$-th element of the other. Instead, it searches for the most plausible nonlinear correspondence between the two sequences under a globally optimal alignment path.

\subsection{DTW-based Intent Trajectory Alignment in Our Method}
\label{appendix:dtw_ours}

In this work, we formulate multi-turn user simulation as a conditional sequential decision-making problem. Given static user variables $u$ and a dynamic interaction context $h_t$, the simulator policy generates a latent intent state and a natural-language response at each turn. We denote the real user trajectory as
\begin{equation}
    \tau = \{(z_t, y_t)\}_{t=1}^{T},
\end{equation}
and the simulator-generated trajectory as
\begin{equation}
    \hat{\tau} = \{(\hat{z}_t, \hat{y}_t)\}_{t=1}^{T^*}.
\end{equation}

Here, $z_t$ and $\hat{z}_t$ denote the latent intent states of the real user and the simulator at turn $t$, respectively, while $y_t$ and $\hat{y}_t$ denote the corresponding natural-language responses.

Our DTW alignment is not directly applied to the full natural-language response sequences. Instead, it is applied to the intent trajectories extracted from multi-turn interactions. Specifically, we define the real user's intent trajectory as
\begin{equation}
    Z = (z_1, z_2, \dots, z_T),
\end{equation}
and the simulator-generated intent trajectory as
\begin{equation}
    \hat{Z} = (\hat{z}_1, \hat{z}_2, \dots, \hat{z}_{T^*}).
\end{equation}

In our data, each intent state is discretized into an ordered level from 0 to 5:
\begin{equation}
    \hat{z}_i, z_j \in \{0,1,2,3,4,5\}.
\end{equation}

These intent levels have a clear ordinal structure: smaller numerical differences indicate more similar user states, while larger differences indicate stronger intent deviation. Therefore, we instantiate the local cost function in DTW as the absolute difference between the simulated intent level and the real intent level:
\begin{equation}
    d(\hat{z}_i, z_j) = |\hat{z}_i - z_j|.
\end{equation}

This design preserves the ordinal structure of intent states. For example, the discrepancy between intent levels 3 and 4 is smaller than that between intent levels 1 and 5. Compared with a binary mismatch indicator, the absolute-difference cost better captures gradual shifts in user intent evolution.

Based on this local cost function, we construct a local cost matrix $C \in \mathbb{R}^{T^* \times T}$:
\begin{equation}
    C(i,j) = d(\hat{z}_i, z_j).
\end{equation}
We then construct a cumulative cost matrix $D \in \mathbb{R}^{T^* \times T}$, where $D(i,j)$ denotes the minimum cumulative cost required to align the simulated intent prefix $(\hat{z}_1,\dots,\hat{z}_i)$ with the real intent prefix $(z_1,\dots,z_j)$.

For dynamic programming, we initialize the boundary as follows:
\begin{equation}
    D(1,1) = C(1,1),
\end{equation}
\begin{equation}
    D(i,1) = C(i,1) + D(i-1,1), 
    \quad i = 2,\dots,T^*,
\end{equation}
\begin{equation}
    D(1,j) = C(1,j) + D(1,j-1),
    \quad j = 2,\dots,T.
\end{equation}
The first row and first column correspond to the case where one trajectory prefix contains only a single element, so the other trajectory can only be aligned to it through consecutive local stretching.

After boundary initialization, for $i>1$ and $j>1$, the cumulative cost matrix is computed as
\begin{equation}
    D(i,j)
    =
    C(i,j)
    +
    \min \{D(i-1,j), D(i,j-1), D(i-1,j-1)\}.
\end{equation}
The three transitions correspond to one-to-many, many-to-one, and one-to-one local alignment between the simulated and real trajectories. Finally, the DTW cumulative cost between the two intent trajectories is defined as
\begin{equation}
    D_{\mathrm{DTW}}(\hat{Z}, Z)
    =
    \min_{P}
    \sum_{(i,j)\in P} d(\hat{z}_i, z_j),
\end{equation}
where $P$ denotes a valid warping path. This path preserves the temporal order within both trajectories while allowing local turn-level stretching or compression.

\paragraph{DTW-based Trajectory Plausibility Reward.}
\label{appendix:dtw_reward}

The above definition corresponds directly to the DTW alignment cost $D_{\mathrm{DTW}}(\hat{z}_{1:T^*}, z_{1:T})$ used in the trajectory plausibility reward $R_{\mathrm{traj}}$ in Section~\ref{method:dtw_based_reward}. Specifically, the trajectory plausibility reward is defined as
\begin{equation}
    R_{\mathrm{traj}}
    =
    \exp\left(
    -\alpha \cdot
    \frac{
    D_{\mathrm{DTW}}(\hat{z}_{1:T^*}, z_{1:T})
    +
    \beta |T^* - T|
    }{T}
    \right)
    \cdot \eta .
\end{equation}

First, $D_{\mathrm{DTW}}(\hat{z}_{1:T^*}, z_{1:T})$ denotes the optimal alignment cost between the simulated and real intent trajectories. It measures the overall deviation between the two trajectories in terms of intent evolution, rather than the discrepancy between fixed turn positions.

Second, $\beta |T^* - T|$ is a trajectory-length penalty. Although DTW allows nonlinear alignment between sequences of different lengths, relying only on the DTW path cost may allow the simulator to avoid certain local deviations by generating trajectories that are overly short or overly long. We therefore explicitly penalize the difference in the number of turns to prevent the simulated trajectory from deviating excessively from the real trajectory length. Here, $T$ denotes the number of turns in the real trajectory, $T^*$ denotes the number of turns in the simulated trajectory, and $\beta$ controls the strength of the length penalty.

Third, the denominator $T$ normalizes the cost by the real trajectory length, improving comparability across sessions of different lengths. Since longer real trajectories naturally tend to accumulate larger DTW costs, omitting this normalization would introduce a systematic length bias into the reward.

Finally, $\eta \in (0,1]$ is a flattening coefficient used to suppress degenerate intent trajectories. We observe that, if only the DTW alignment cost and the length penalty are used, the simulator may tend to generate nearly constant intent sequences in order to reduce path deviation. To discourage this behavior, when the real trajectory contains substantial intent variation but the simulated trajectory remains at a single or nearly single intent level for a long period, we set $\eta$ to a discount value smaller than 1. This design encourages the simulator not only to match the terminal state and reduce the overall path cost, but also to preserve the dynamic variation pattern observed in real user trajectories.

Therefore, $R_{\mathrm{traj}}$ becomes high when the simulated and real trajectories have similar intent evolution paths, comparable numbers of turns, and no overly flat degenerate behavior. Conversely, the reward decreases when the simulated trajectory reaches the same terminal state but follows a substantially different intermediate intent path, or when its number of turns deviates significantly from the real trajectory.

\paragraph{Turn-level Local Deviation from the DTW Optimal Path.}
\label{appendix:dtw_local_deviation}

Our fine-grained advantage modulation mechanism further uses the DTW alignment result to construct a local deviation measure $d_t$ for each generated turn. This corresponds to the future deviation score defined in Section~\ref{method:advantage_modulation}:
\begin{equation}
    c_t = \sum_{k=0}^{T^*-t} \rho^k \cdot d_{t+k}.
\end{equation}
Here, $d_t$ is not a strict pointwise difference $|\hat{z}_t - z_t|$. Instead, it is an aligned local deviation derived from the DTW optimal path.

Let the optimal DTW alignment path be
\begin{equation}
    P^*
    =
    \big((i_1,j_1), (i_2,j_2), \dots, (i_K,j_K)\big),
\end{equation}
where each path element $(i_k,j_k)$ indicates that the $i_k$-th simulated turn $\hat{z}_{i_k}$ is aligned with the $j_k$-th real turn $z_{j_k}$. The path satisfies the boundary, monotonicity, and step-size constraints, thereby preserving the temporal order of both trajectories while allowing local one-to-many, many-to-one, and one-to-one alignments.

Because DTW permits one-to-many and many-to-one alignments, the same simulated turn $i$ may correspond to multiple real turns. To obtain a local deviation measure for each simulated turn, we first collect all real-turn indices aligned with simulated turn $i$ under the optimal path:
\begin{equation}
    \mathcal{A}_i
    =
    \{j_k \mid (i_k,j_k) \in P^*,\ i_k = i\}.
\end{equation}
We then define the local deviation of the $i$-th simulated turn as the average absolute deviation over all aligned real turns:
\begin{equation}
    d_i
    =
    \frac{1}{|\mathcal{A}_i|}
    \sum_{j \in \mathcal{A}_i}
    |\hat{z}_i - z_j|.
\end{equation}

This definition converts the globally optimal DTW alignment path into a turn-level deviation signal over the simulated trajectory. Intuitively, if a simulated intent state at a certain turn differs substantially from the real intent stages aligned to it by DTW, then its local deviation $d_i$ will be large. If the simulated turn matches the corresponding real trajectory stage well, then $d_i$ will be small.

Based on these local deviations, the future deviation score accumulates downstream deviations starting from turn $t$:
\begin{equation}
    c_t
    =
    \sum_{k=0}^{T^*-t}
    \rho^k d_{t+k},
\end{equation}

where $\rho \in [0,1]$ is a discount factor. This score measures the extent of intent deviation that may accumulate in the future trajectory after the current turn. A larger $c_t$ indicates that the simulated trajectory exhibits more severe or more persistent deviation after turn $t$.

In turn-level advantage modulation, $c_t$ controls the strength of advantage reweighting. For positive trajectory-level advantages, turns with larger $c_t$ receive relatively weaker positive reinforcement, because they are followed by larger downstream deviations despite the overall trajectory being favorable. For negative trajectory-level advantages, turns with larger $c_t$ receive stronger negative updates, because they are more likely to be associated with persistent downstream drift.

It is important to distinguish the roles of $D_{\mathrm{DTW}}$ and $d_i$. The term $D_{\mathrm{DTW}}$ is a trajectory-level cumulative alignment cost used to construct the trajectory plausibility reward $R_{\mathrm{traj}}$. In contrast, $d_i$ is a turn-level local deviation derived from the same DTW optimal path and is used to construct the future deviation score $c_t$, which further modulates the advantage at each turn. Thus, the two quantities share the same DTW alignment process but serve different levels of training signal: reward design at the trajectory level and credit assignment at the turn level.

\paragraph{Illustrative Example.}
\label{app:dtw_example}

Consider a real intent trajectory
\begin{equation}
    Z = [3,4,5],
\end{equation}

and a simulator-generated trajectory
\begin{equation}
    \hat{Z} = [3,3,4,5].
\end{equation}

Under strict turn-by-turn alignment, the simulated trajectory would become misaligned from the second turn onward due to the length difference. In contrast, DTW can naturally align the first two simulated intent states with the first real intent stage, and then align the subsequent states 4 and 5 with the corresponding real intent states 4 and 5. Therefore, the extra simulated turn is not treated as a complete trajectory-level error. Instead, it is interpreted as a local pacing delay.

Similarly, if the simulator completes need confirmation and intent advancement within a single turn, whereas the real user completes the same transition across two adjacent turns, DTW allows this simulated turn to be flexibly aligned with the neighboring stages in the real trajectory. In this way, DTW better reflects natural pacing variation in multi-turn dialogue, rather than penalizing every local turn-level shift as an error.

Through DTW alignment, our method assigns each simulated turn a more reasonable reference position in the real trajectory. This yields both a trajectory-level cumulative alignment cost $D_{\mathrm{DTW}}$ and a turn-level local deviation signal $d_i$. The former is used to construct the trajectory plausibility reward, while the latter is used for the future deviation score and turn-level advantage modulation. As a result, the simulator is no longer optimized under the overly rigid assumption that ``the $t$-th simulated turn must correspond to the $t$-th real turn.'' Instead, it is encouraged to match the overall intent evolution path of real users.

\FloatBarrier

\section{Case Study}
\label{appendix:case_study}

% 主外框
\begin{tcolorbox}[
    enhanced,
    colback=white,
    coltitle=black,
    % 【2. 怎么调整边框颜色】修改 colframe=... 即可。例如 black!80 表示 80% 黑，改成 blue!50 就是浅蓝。
    colframe=blue!10, 
    boxrule=1.5pt,
    % 【1. 怎么调整大标题大小】将原来的 \Large 改成了 \large，如果你想更小可改为 \normalsize
    title=\textbf{\normalsize CASE STUDY: Correct Conversion Does Not Guarantee Behavioral Fidelity},
    center title,
    fonttitle=\sffamily\bfseries,
    toptitle=1mm,
    bottomtitle=1mm,
    arc=1.5mm,
    outer arc=1.5mm
]

    % ==========================================
    % 1. 用户画像与任务背景 
    % ==========================================
    \begin{tcolorbox}[
        colback=gray!5, 
        colframe=gray!50, % 【2. 此处改边框颜色】
        title=User Profile \& Task Background, 
        colbacktitle=gray!10, 
        coltitle=black, 
        % 【5. 调整框内标题大小】使用 fonttitle=... 控制。这里设为 \normalsize\bfseries
        fonttitle=\footnotesize\bfseries, 
        boxrule=0.5pt, 
        arc=1.2mm,
        left=2mm, right=2mm, top=1mm, bottom=1mm
    ]
    % 【5. 调整框内正文大小】使用 \small, \footnotesize, \scriptsize 等控制
    \scriptsize
    This is an 18–23-year-old platform user with a high-school education or below. The user has a partner and has no children. The user works as a general worker and commonly uses an Android phone priced around RMB 1,700. The user currently lives in Chaozhou, Guangdong Province, a fourth-tier city in China. The user's personality trait is decisive and concise. In this dialogue, the user intends to buy a car, asks about the down-payment threshold and available vehicle models, and hopes to schedule a test drive.
    \end{tcolorbox}
    \vspace{-2mm}

    % ==========================================
    % 2. 真实数据 Ground Truth
    % ==========================================
    \begin{tcolorbox}[
        colback=gtback, 
        colframe=gtframe, % 【2. 此处改边框颜色】
        title=Ground Truth, 
        colbacktitle=gttitle, 
        coltitle=black!80, 
        fonttitle=\footnotesize\bfseries, % 【5. 标题大小】
        boxrule=0.5pt, 
        arc=1.2mm,
        left=2mm, right=2mm, top=1mm, bottom=1mm
    ]
    \scriptsize% 【5. 正文大小】
    \agent{Hello, click the free consultation below, "Youjia" will serve you immediately.}
    \userA{What's the minimum down payment?}
    \agent{Hello, the down payment ratio varies for different models, and prices change dynamically\textasciitilde{} You can leave your contact info, and our specialist will provide one-on-one service to help you choose the most suitable car-buying plan.}
    \userA{What cars do you have?}
    \agent{Hello, we have a variety of popular models on sale. Do you have any preferences or intended models?}
    \userA{Can I test drive?}
    \agent{Hello, yes you can. Do you have a specific model in mind for the test drive? I can arrange it for you in advance. Please leave your contact info, I'll send you the store's location. You're welcome to come and experience it\textasciitilde{}}
    \userA{132xx9875}
    \end{tcolorbox}
    \vspace{-2mm}

    % ==========================================
    % 3. 模型对比列与分析
    % ==========================================
    \noindent
    % 左侧：Ours 
    \begin{minipage}[t]{0.24\textwidth}
        \vspace{0pt} 
        \begin{tcolorbox}[
            equal height group=models, % 【3. 底部对齐核心机制】所有写了 group=models 的框会强行等高
            colback=green!2, 
            colframe=green!50!black, 
            title=TRACER-7B (Ours), % 【6. 已去除了 \hfill \cmark 勾勾】
            colbacktitle=green!70!black, 
            fonttitle=\scriptsize\bfseries, % 【5. 标题大小】
            boxrule=0.5pt, 
            arc=1.2mm, 
            left=1mm, right=1mm, top=1mm, bottom=1mm
        ]
        \scriptsize % 【5. 框内正文大小】
        \agent{Hello, click the free consultation below, "Youjia" will serve you immediately.}
        \userA{What's the minimum down payment?}
        \agent{The down payment plan depends on the model and policies, and needs to be customized based on your budget. Do you have a target budget range?}
        \userA{What cars are there?}
        \agent{We have multiple popular models on sale with various configurations. We need to recommend based on your needs, which brand do you prefer?}
        \userA{Can I schedule a test drive}
        \agent{Test drives are supported, and you can schedule the time yourself. Leave your contact info, and we will contact you as soon as possible to arrange the test drive\textasciitilde{}}
        \userA{132xx9875}
        \end{tcolorbox}
    \end{minipage}%
    \hfill
    % 右侧：三个 Baseline 模型 
    \begin{minipage}[t]{0.74\textwidth}
        \vspace{0pt} 
        
        \noindent
        \begin{minipage}[t]{0.32\linewidth}
            \vspace{0pt}
            \begin{tcolorbox}[
                equal height group=models, % 【3. 等高指令】
                colback=baseback, colframe=baseframe, 
                title=Gemini-3.1-Pro, % 【6. 已去除 \xmark 叉叉】
                colbacktitle=basetitle, 
                fonttitle=\scriptsize\bfseries,
                boxrule=0.5pt, arc=1.2mm, left=1mm, right=1mm, top=1mm, bottom=1mm
            ]
            \scriptsize
            \agent{Hello, click the free consultation below, "Youjia" will serve you immediately.}
            \userA{I want to buy a car, what's the minimum down payment? Any model recommendations? I want to test drive first.}
            \agent{Down payments vary depending on the model and plan\textasciitilde{} Is it convenient to leave your contact info? We can arrange a consultant to introduce in detail. What's your budget?}
            \userA{Budget is under 100k, the lower the down payment, the better. My phone number is 132xx9875, same on WeChat, just have the consultant add me and send the models, and arrange a test drive by the way.}
            \end{tcolorbox}
        \end{minipage}%
        \hfill
        \begin{minipage}[t]{0.32\linewidth}
            \vspace{0pt}
            \begin{tcolorbox}[
                equal height group=models, % 【3. 等高指令】
                colback=baseback, colframe=baseframe, 
                title=Doubao-Seed-2.0-Pro, % 【6. 已去除叉叉】
                colbacktitle=basetitle, 
                fonttitle=\scriptsize\bfseries,
                boxrule=0.5pt, arc=1.2mm, left=1mm, right=1mm, top=1mm, bottom=1mm
            ]
            \scriptsize
            \agent{Hello, click the free consultation below, "Youjia" will serve you immediately.}
            \userA{I'd like to ask, how much is the down payment generally for buying a car now? Are there any cost-effective models suitable for ordinary workers like us? Can a test drive be arranged?}
            \agent{Down payments differ depending on the model, and test drives require a reservation. We can arrange a one-on-one consultation to match plans for you\textasciitilde{}}
            \userA{My budget is around 100k. I'm in Guxiang Town, Chao'an, Chaozhou. I just want a fuel-efficient and durable commuter car. How can I contact your consultants?}
            \agent{Understood your needs\textasciitilde{} We will help you find suitable models. Leave a contact number, we'll arrange a one-on-one recommendation and contact you\textasciitilde{}}
            \userA{Okay, my phone is 132xx9875, just arrange for someone to contact me.}
            \end{tcolorbox}
        \end{minipage}%
        \hfill
        \begin{minipage}[t]{0.32\linewidth}
            \vspace{0pt}
            \begin{tcolorbox}[
                equal height group=models, % 【3. 等高指令】
                colback=baseback, colframe=baseframe, 
                title=Doubao-Seed-Character, % 【6. 已去除叉叉】
                colbacktitle=basetitle, 
                fonttitle=\scriptsize\bfseries,
                boxrule=0.5pt, arc=1.2mm, left=1mm, right=1mm, top=1mm, bottom=1mm
            ]
            \scriptsize
            \agent{Hello, click the free consultation below, "Youjia" will serve you immediately.}
            \userA{I want to buy a car, may I ask what the down payment threshold is, and are there any recommended models suitable for ordinary wage earners? I also want to book a test drive.}
            \agent{Down payments vary depending on the model\textasciitilde{} Leave your contact info, and we'll arrange a consultant to introduce in detail\textasciitilde{} What price range are you considering?}
            \userA{I want a commuter car under 100k, my phone number is 132xx9875, WeChat is lalala, please arrange a consultant to connect with me, and I also want to book a test drive.}
            \end{tcolorbox}
        \end{minipage}
        
        \vspace{1mm} 
        
        % 下半部分：合并的失败原因分析
        \begin{tcolorbox}[
            colback=baseback, 
            colframe=baseframe, 
            title=\ding{118} Baseline Failure Analysis, 
            colbacktitle=basetitle, 
            coltitle=white, 
            fonttitle=\footnotesize\bfseries, 
            boxrule=0.5pt, 
            arc=1.2mm, 
            left=2mm, right=2mm, top=1mm, bottom=1mm
        ]
        % 【4. 只需要对应的英文，不再写分析详情】
        \scriptsize 
        \begin{itemize}[leftmargin=*, itemsep=0pt, parsep=0pt, topsep=0pt]
            \item \textbf{Lack of flaws.}
            \item \textbf{Overly cooperative, responses are too mechanical.}
            \item \textbf{Out of Character (OOC): Speech and behavior seriously violate the provided profile.}
        \end{itemize}
        \end{tcolorbox}
    \end{minipage}

\end{tcolorbox}

\newpage
\section{Prompt Details}
\label{appendix:prompt}

\promptfilebox{Persona and Task Background Extraction Prompt}{prompts/persona_extraction_prompt.txt}

\promptfilebox{Intent-Level Annotation Prompt}{prompts/intent_level_prompt.txt}

\promptfilebox{Data Filtering and Cleaning Prompt}{prompts/data_filtering_prompt.txt}

\promptfilebox{Merchant Agent System Prompt for Training and Test Data}{prompts/merchant_system_prompt.txt}

\promptfilebox{User Simulator System Prompt for Training and Test Data}{prompts/user_system_prompt.txt}

% \section{Appendix}

\end{document}